%% file: representation.tex
\documentclass[12pt]{article}
\usepackage{amsmath}
\usepackage{amssymb}
\usepackage{graphicx}
\usepackage{color}
\usepackage{multirow}
\usepackage{cancel}
\usepackage[authoryear]{natbib}
\usepackage{subcaption}
\usepackage{placeins}
\usepackage{siunitx}
\usepackage{booktabs}
\usepackage{tikz}
\usepackage{textcomp,gensymb} 
\usepackage{a4wide}
\usepackage{tikz}
\usepackage{rotating}
\usepackage{enumerate}
\usepackage[symbol]{footmisc}
\usepackage{tabularx}
\usepackage{hyperref}
\usepackage{authblk}

\usetikzlibrary{positioning}

\renewcommand{\thefootnote}{\arabic{footnote}}

\input{figures/setup.tex}

\title{A Survey on Self-Supervised Representation Learning}

\author[1]{Tobias Uelwer}
\author[1]{Jan Robine}
\author[2]{Stefan Sylvius Wagner}
\author[1]{Marc Höftmann}
\author[3,4]{\\Eric Upschulte}
\author[1]{Sebastian Konietzny}
\author[2]{Maike Behrendt}
\author[1]{Stefan Harmeling}

\affil[1]{Department of Computer Science, Technical University of Dortmund, Germany}
\affil[2]{Department of Computer Science, Heinrich Heine University D\"usseldorf, Germany}
\affil[3]{Institute of Neuroscience and Medicine (INM-1), Research Center J\"ulich, Germany}
\affil[4]{Helmholtz AI, Research Center J\"ulich, Germany}
\date{}

\newcommand{\mcrot}[4]{\multicolumn{#1}{#2}{\rlap{\rotatebox{#3}{#4}~}}}

\DeclareMathOperator{\sg}{sg}

\DeclareMathOperator*{\argmax}{argmax}

\DeclareMathOperator{\softmax}{softmax}

\DeclareMathOperator{\onehot}{onehot}

\DeclareMathOperator{\infonce}{InfoNCE}

\newcommand{\dmse}{d_\text{se}}
\newcommand{\dnormmse}{d_{\text{nse}}}
\newcommand{\scos}{s_\text{cos}}
\newcommand{\dce}{d_\text{ce}}
\newcommand{\dsoftmax}{d_\text{classification}}
\newcommand{\tminus}{{\scalebox{0.5}[1]{$\displaystyle -$}}}
\newcommand{\tplus}{{\scalebox{0.6}{$\displaystyle +$}}}

\let\svthefootnote\thefootnote

\begin{document}

\maketitle

\let\thefootnote\relax\footnotetext{Correspondence to: \texttt{\href{mailto:tobias.uelwer@tu-dortmund.de}{tobias.uelwer@tu-dortmund.de}}}
\let\thefootnote\svthefootnote

\begin{abstract}
	Learning meaningful representations is at the heart of many tasks in the field of modern machine learning. Recently, a lot of methods were introduced that allow learning of image representations without supervision. These representations can then be used in downstream tasks like classification or object detection. The quality of these representations is close to supervised learning, while no labeled images are needed. This survey paper provides a comprehensive review of these methods in a unified notation, points out similarities and differences of these methods, and proposes a taxonomy which sets these methods in relation to each other. Furthermore, our survey summarizes the most-recent experimental results reported in the literature in form of a meta-study. Our survey is intended as a starting point for researchers and practitioners who want to dive into the field of representation learning. 
\end{abstract}

\tableofcontents

\input{sections/introduction.tex}

\input{sections/pretext.tex}

\input{sections/information-max.tex}

\input{sections/distillation.tex}
\input{sections/contrastive.tex}

\input{sections/cluster.tex}

\input{sections/taxonomy.tex}
\input{sections/comparison.tex}
\input{sections/conclusion.tex}

\bibliographystyle{apalike} 
\bibliography{references}

\end{document}

%% file: figures/setup.tex
\usetikzlibrary{arrows.meta}
\usetikzlibrary{shapes.geometric}
\usetikzlibrary{calc}
\usetikzlibrary{topaths}
\pgfdeclarelayer{back}
\pgfsetlayers{back,main}

\makeatletter
\pgfkeys{%
  /tikz/on layer/.code={
    \def\tikz@path@do@at@end{\endpgfonlayer\endgroup\tikz@path@do@at@end}%
    \pgfonlayer{#1}\begingroup%
  }%
}
\makeatother

\definecolor{niceblue}{HTML}{BBDEFB}
\definecolor{nicedarkblue}{HTML}{42A5F5}
\definecolor{nicegray}{HTML}{BDBDBD}
\definecolor{nicelightgray}{HTML}{E7E7E7}
\definecolor{nicedarkgray}{HTML}{757575}
\definecolor{nicegreen}{HTML}{C8E6C9}
\definecolor{nicedarkgreen}{HTML}{4CAF50}
\definecolor{nicepurple}{HTML}{E8CCEC}
\definecolor{nicedarkpurple}{HTML}{BA68C8}
\definecolor{nicered}{HTML}{F9BABE}
\definecolor{nicedarkred}{HTML}{E57373}
\definecolor{niceyellow}{HTML}{FFF9C4}
\definecolor{nicedarkyellow}{HTML}{FFEE58}

\usetikzlibrary{decorations.pathreplacing,calligraphy} 

\tikzset{var/.style={circle,line width=0.35mm,font=\footnotesize, minimum size=8mm,inner sep=0mm,outer sep=0mm,text height=2mm,text depth=0mm}}
\tikzset{bluevar/.style={var,fill=niceblue,draw=nicedarkblue}}
\tikzset{grayvar/.style={var,fill=nicegray,draw=nicedarkgray}}
\tikzset{greenvar/.style={var,fill=nicegreen,draw=nicedarkgreen}}
\tikzset{purplevar/.style={var,fill=nicepurple,draw=nicedarkpurple}}
\tikzset{redvar/.style={var,fill=nicered,draw=nicedarkred}}
\tikzset{whitevar/.style={var,fill=white,draw=nicegray}}
\tikzset{yellowvar/.style={var,fill=niceyellow,draw=nicedarkyellow}}

\tikzset{smallbox/.style={rectangle, fill=nicegray,draw=nicedarkgray, line width=0.35mm,minimum width=8mm,minimum height=1.5mm,inner sep=0mm,outer sep=0mm}}

\tikzset{bigbox/.style={rectangle, fill=nicegray,draw=nicedarkgray, line width=0.35mm,minimum width=8mm,minimum height=3mm,inner sep=0mm,outer sep=0mm,text height=2mm,text depth=0mm}}

\tikzset{box/.style={rectangle,rounded corners,line width=0.35mm,minimum size=7mm,outer sep=0mm,text height=2mm,text depth=0mm}}
\tikzset{nicearrow/.style={>={Stealth[length=2mm,width=1.5mm]},line width=0.35mm,bend left=90}}
\tikzset{nicelabel/.style={minimum size=0mm,font=\footnotesize,minimum height=6mm,inner ysep=0mm,outer sep=0mm,text height=2mm,text depth=0mm}}

%% file: sections/introduction.tex
\section{Introduction}

Images often contain a lot of information which is irrelevant for
further downstream tasks. A \emph{representation} of an image ideally
extracts the relevant parts of it. The goal of \emph{representation learning} is
to learn an \emph{encoder network} $f_\theta$ with learnable parameters $\theta$ that maps an input image $x$ to a lower-dimensional representation (embedding) $y = f_\theta(x)$.
The main purpose of this paper is to present and discuss the
different approaches and ideas for finding useful encoders. Note that in the
machine learning literature this setup is also called feature learning or
feature extraction.

\subsection{Supervised and Unsupervised Learning}

Before categorizing the various methods of representation learning, we begin by
distinguishing the two basic settings of machine learning: (i) \emph{supervised
  learning} learns a function given labeled data points, interpreted as input and
output examples, (ii) whereas in \emph{unsupervised learning} we learn something
given only data points (unlabeled). It is possible to take the view that also
for unsupervised learning the goal is to learn a function, e.g., in clustering
we learn a classification function and invent labels at the same time, in
dimensionality reduction (such as PCA, ISOMAP, LLE, etc.) we learn a regression
function and invent lower-dimensional embeddings simultaneously. In light of
this, representation learning is an instance of unsupervised learning, since we
are only given a set of unlabeled data points and our goal is to learn an
encoder, that maps the data onto some other representation that has favorable
properties.

\subsection{Self-supervised Learning}

Recently, new machine learning methods emerged, that have been labeled
\emph{self-supervised learning}. In a nutshell, such methods create an encoder
by performing the following two separate steps:
\begin{enumerate}[(i)]
  \item Invent a supervised learning task by creating a target $t$ for each given image $x$.
  \item Apply supervised learning to learn a function from inputs $x$ to targets
        $t$.
\end{enumerate}
The mechanism that generates the targets can be manually designed or can include
learned neural networks. Note that the targets are not necessary stationary, but could change during training.
Even though self-supervised learning applies classical supervised learning as
its second step, it is overall best viewed as an unsupervised method, since it
only takes unlabeled images as its starting point.

\subsection{Outline}

This paper strives to give an overview over recent advances in representation
learning. Starting from the autoencoder, we will discuss different methods
which we group in the following way:
\begin{itemize}
  \item \emph{Pretext task methods} solve an auxiliary task, e.g., by predicting
        the angle by which an input image was rotated. The idea is that the
        representations learned along the way are also useful for solving other
        tasks. We will discuss these methods further in Section
        \ref{sec:pretext}.
  \item \emph{Information maximization methods} learn networks that are
        invariant to various image transformations and at the same time avoid
        trivial solutions by maximizing information content.
        Section~\ref{sec:infomax} presents some of these methods in detail.
  \item \emph{Teacher-student methods} consist of two networks where one
        extracts knowledge from the other. We will take a closer look at these
        methods in Section \ref{sec:distillation}.
  \item \emph{Contrastive learning methods} discriminate between positive and
        negative examples that are defined by the method on-the-fly. In Section
        \ref{sec:contrastive} we review these contrastive methods in detail.
  \item \emph{Clustering-based methods} invent multiple class labels by clustering
        the representations and then train a classifier on those labels. Section
        \ref{sec:clustering} summarizes the most relevant representation
        learning methods that use clustering.
\end{itemize}
In Section \ref{sec:taxonomy} we further put the discussed methods in relation to
each other and in Section \ref{sec:meta} we summarize experimental results that
were reported in the literature in a meta study.

While we focus on approaches for visual representation learning in this paper,
there exist further approaches which are specialized for graphs
\citep{grover2016node2vec,perozzi2014deepwalk,kipf2016variational}, time-series
\citep{eldele2021time} or text
\citep{mikolov2013distributed,mikolov2013efficient,kenton2019bert}, which we
omit.

\subsection{Notation}
\label{sec:notation}

Before describing the specifics of the different representation learning methods, we start by defining the notation used in this paper. Given a dataset of images, we write \begin{equation}
  X = [x_1, \ldots, x_n]
\end{equation} for a randomly sampled batch of images. Every representation learning method trains an encoder network $f_\theta$, where $\theta$ are the learnable
parameters. This encoder network computes a representation
\begin{equation}
  Y =  [y_1, \ldots, y_n] = [f_\theta(x_1), \ldots, f_\theta(x_n)] = f_\theta(X)
\end{equation}
of the images in  $X$. 
Some methods additionally train a projection network $g_\phi$, with parameters
$\phi$, that computes projections
\begin{equation}
  Z =  [z_1, \ldots, z_n] = [g_\phi(y_1), \ldots, g_\phi(y_n)] = g_\phi(Y)
\end{equation}  of the representations in $Y$.  There are methods that also train a prediction network $q_\psi$, with parameters $\psi$, that computes a prediction based on $z$ or $y$. Both, projections and predictions are only used to train the network and after training the projection and prediction networks are discarded, and only the encoder $f_\theta$ is used for downstream tasks.

Most methods apply a transformation to the input image to obtain a \emph{view} of the
original image. We  write $x_i^{(j)} = t(x_i)$ for the $j$-th view of the original image $x_i$, which was obtained by applying the transformation $t$.
Usually, these methods randomly sample transformations from a given set of transformations $\mathcal{T}$ and can differ for each image in the batch. This is why we treat $t$ as a random variable that is sampled for each image of the batch. In contrast to that, other methods use predefined transformations $t^{(1)},\dots, t^{(m)}$ that are fixed and do not change. In Section \ref{sec:infomax} we give more details and some examples of these transformations. 
We write \begin{equation}
  X^{(j)} = [ x_1^{(j)}, \ldots, x_n^{(j)} ] = t([x_1, \ldots, x_n]) = t(X) 
\end{equation} for the batch of the $j$-th view.
We use a similar notation for the resulting representations
$Y^{(j)} = [y_1^{(j)}, \ldots, y_n^{(j)}]$ and projections 	$Z^{(j)} = [z_1^{(j)}, \ldots, z_n^{(j)}]$.
Some methods split the input image into patches, which we also consider as a special case of
transformation, where each patch is a view of the image.
In that case we write $X_i = [x_i^{(1)}, \ldots, x_i^{(m)}],$
which contains all $m$ views of the image $x_i$ and denote the corresponding representations and projections as $Y_i = [y_i^{(1)}, \ldots, y_i^{(m)}]$ and $Z_i = [z_i^{(1)}, \ldots, z_i^{(m)}]$.

In some cases, the calculation of the representations can be decoupled such that each 
view is processed independently, i.e., 
$	Y_i = [f_\theta(x_i^{(1)}), \ldots, f_\theta(x_i^{(m)})] = f_\theta(X_i).$
If this is possible we call the corresponding encoder a \emph{Siamese} encoder. The same distinction can be made for the other networks (projector and
predictor). A Siamese projector computes 
$Z_i  = [g_\phi(y_i^{(1)}), \ldots, g_\phi(y_i^{(m)})]=g_\phi(Y_i)$
 individually. However, as we will see later, this is not always the case as some networks operate on multiple views simultaneously.

We use $\mathcal{L}$ to refer to the loss
function that is used to train the parameters using
stochastic gradient descent (SGD). Sometimes the total loss
consists of multiple components which we denote using the letter $\ell$.

We use square brackets to access elements of vectors and matrices, e.g., the
$j$-th element of vector $v$ is written as $v[j]$ and the entry at column $j$
and row $k$ of matrix $C$ is written as $C[j,k]$. Furthermore, we define the
softmax function that normalizes a vector $v\in\mathbb{R}^d$ to a probability
distribution as
\begin{equation}
  \left(\softmax_\tau(v)\right)\![j]= \dfrac{\exp(v[j]/\tau)}{\sum_{k=1}^{d} \exp(v[k]/\tau)} \text{ for } j = 1, \ldots, d,
\end{equation}
where $\tau > 0$ is a temperature parameter which controls the entropy of that
distribution~\citep{wu2018unsupervised}. The higher the value of $\tau$ the
closer the normalized distribution is to a uniform distribution. We write
$\softmax(v)$ when no temperature is used, i.e., when $\tau = 1$.

\paragraph{Distance metrics and similarity measures.}
Throughout the paper, we use various concepts to compare two vectors. We introduce these concepts in the following. First, we define the squared error between
two vectors $v, w \in \mathbb{R}^d$ as
\begin{equation}
  \dmse(v,w)  = \left\| v - w \right\|_2^2 = (v-w)^\top(v-w)=\sum_{j=1}^d (v[j] - w[j])^2.
\end{equation}
Note that the squared error is the same as the squared Euclidean norm of the residuals.
Sometimes, the vectors $v$ and $w$ are normalized before the squared error is
calculated. We denote the resulting distance metric as normalized squared error
\begin{equation}
  \dnormmse(v,w) = \left\| \frac{v}{\left\| v \right\|_2} - \frac{w}{\left\| w \right\|_2} \right\|_2^2.
\end{equation}
We define the cosine similarity as
\begin{equation}
  \scos(v,w)  = \frac{v^\top w}{\left\| v \right\|_2 \left\| w \right\|_2}.
\end{equation}
Note that $\dnormmse$ and $\scos$ are linearly related, i.e.,
\begin{align}
  \dnormmse(v,w)             & = 2 - 2\, \frac{v^\top w}{\left\| v \right\|_2 \left\| w \right\|_2} = 2-2\, \scos(v,w) \\
  \Leftrightarrow \scos(v,w) & =1- \frac{1}{2}\, \dnormmse(v,w) .
\end{align}
To measure the distance between two discrete probability distributions described
by the entries of two vectors $v$ and $w$, the cross-entropy can be used which
is given as
\begin{equation}
  \dce(v,w)  = -\sum_{j=1}^d v[j] \log w[j].
\end{equation}
Note that the cross-entropy does not fulfill the triangle inequality and is therefore no
distance metric in a mathematical sense. A common loss function for multiclass
classification is the cross-entropy between scores $v \in \mathbb{R}^d$ that are
normalized using softmax and a one-hot distribution of the true class label $c
  \in \mathbb{N}$, which we denote by
\begin{align}
  \dsoftmax(v,c) & = \dce(\onehot(c), \softmax(v))                           \\
                 & = -v[c] + \log \!\left( \sum_{j=1}^d \exp v[j] \right)\!,
\end{align}
where $\onehot(c)$ is a vector with $1$ in the $c$-th component and $0$ everywhere else.

%% file: sections/pretext.tex
\section{Pretext Task Methods}
\label{sec:pretext}
In the introduction we have defined the concept of self-supervised learning
which relies on defining a mechanism that creates targets for a supervised
learning task. There are many possibilities to invent such supervised learning
problems. These learning problems are usually called \emph{pretext} tasks. The
idea is that the features learned by solving the pretext task are also helpful
for solving other problems. In the following we present works, that creatively
came up with such tasks.

\input{sections/autoencoders.tex}
\input{sections/rotnet.tex}
\input{sections/jigsaw.tex}
\input{sections/mae.tex}

%% file: sections/autoencoders.tex
\subsection{Autoencoders}

\begin{figure}
	\centering
	\begin{tikzpicture}
		\node[inner sep=1mm] (x) at (0,0) {\includegraphics[width=1.75cm]{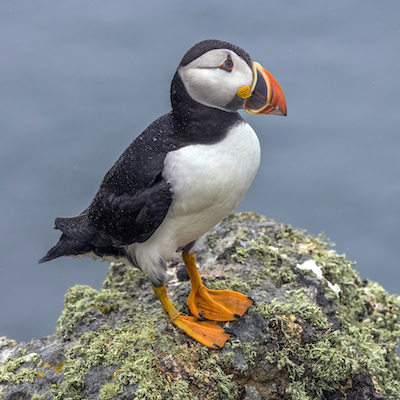}};
		
		\node[bluevar] (y) at ($(x) + (2.5,0)$) {$y$};
		
		\node[inner sep=1mm] (p) at (5,0) {\includegraphics[width=1.75cm]{figures/jigsaw/puffin.jpg}};
		\node[nicelabel] (repr) at ($(y) + (0,1.25)$) {represent};
		\node[nicelabel] (repr) at ($(x) + (0,1.25)$) {view};
		\node[nicelabel] (pred) at ($(p) + (0,1.25)$) {predict};
		
		\begin{pgfonlayer}{back}
			\begin{scope}[transform canvas]  
				\draw[->,nicearrow] (x) -- node[midway,above,nicelabel] {$f_\theta$} (y);
				\draw[->,nicearrow] (y) -- node[midway,above,nicelabel] {$q_\psi$} (p);
			\end{scope}
		\end{pgfonlayer}
	\end{tikzpicture}
	\caption{An autoencoder consists of two networks: an encoder $f_\theta$ that maps the input image to a representation and a predictor $q_\psi$ that is trained to reconstruct the original image from the representation.}
\end{figure}

Autoencoders \citep{le1987modeles} have been part of machine learning for a very
long time and in the light of the previous section, they can be seen as
early instances of self-supervised learning: (i) the invented targets are the
inputs themselves and (ii) the learned function is a bottleneck neural network
consisting of an encoder $f_\theta$ that maps an image $x_i$ to a
low-dimensional representation ${y_i = f_\theta(x_i)}$ and a predictor $q_\psi$
that reconstructs the input image ${\hat{x}_i = q_\psi(y_i)}$ from its
representation.
Given a batch of images $X$, the encoder and predictor networks are jointly trained to minimize the
reconstruction error
\begin{equation}
  \mathcal{L}^\text{AE}_{\theta,\psi} = \frac{1}{n} \sum_{i=1}^n \dmse(\hat{x}_i, x_i).
\end{equation}

There are many variants of the autoencoder: denoising autoencoder
\citep{le1987modeles}, stacked denoising autoencoder \citep{vincent2010stacked},
contractive autoencoder \citep{rifai2011contractive} or variational autoencoder
\citep[VAE,][]{kingma2013auto}.

%% file: sections/rotnet.tex
\subsection{Rotation Network (RotNet)}

The idea of the RotNet which was introduced by
\citet{gidaris2018unsupervised} is to learn a representation that is useful to
predict the angle of a random rotation applied to the input image. The
assumption is that a representation that can predict the rotation is also
valuable for other tasks. The authors show that even a small number of rotations
is sufficient to learn a good representation. Best results are obtained when the
number of rotations is four (0\degree, 90\degree, 180\degree, 270\degree). In
that case the rotation augmentation can be efficiently implemented using flips
and transposes and no interpolation is needed.

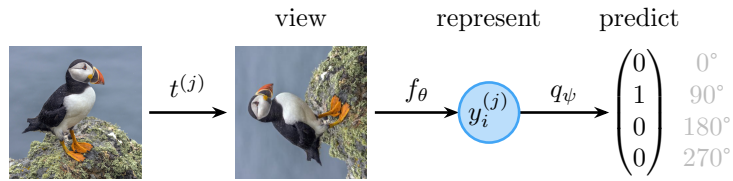
\begin{figure}[h]
	\centering \input{figures/rotnet.tex}
	\caption{RotNet solves the task of predicting 
		image rotations to obtain representations for downstream tasks.}
	\label{fig:rotation}
\end{figure}

Given a batch of images $X$, we consider a single image $x_i$. Four views
${x_i^{(j)} = t^{(j)}(x_i)}$ are created using the rotation transformations
${t^{(1)}, t^{(2)}, t^{(3)}, t^{(4)}}$. The Siamese encoder $f_\theta$ converts
each view into a representation ${y_i^{(j)} = f_\theta(x_i^{(j)})}$. The Siamese
predictor $q_\psi$ is then used to predict the index of the rotation that was applied to the original image.
Both networks are trained by minimizing the classification loss
\begin{equation}
  \label{eq:rotation}
  \mathcal{L}^\text{RotNet}_{\theta,\psi} = \frac{1}{n} \sum_{i=1}^n \sum_{c=1}^4 \dsoftmax(q_\psi(y_i^{(c)}), c)
\end{equation}
for each of the four rotations. After training, the classification head $q_\psi$ is discarded and only $f_\theta$ is used to calculate representations (of  images that were not rotated). The authors use a Network-in-Network architecture~\citep{lin2013network} for their
experiments on CIFAR-10 and the AlexNet architecture~\citep{krizhevsky2017imagenet} for experiments on ImageNet.

%% file: figures/rotnet.tex
\begin{tikzpicture}
  \node[inner sep=1mm] (x) at (0,0) {\includegraphics[width=1.75cm]{figures/jigsaw/puffin.jpg}};

  \node[inner sep=1mm] (xt) at ($(x) + (3,0)$) {\includegraphics[width=1.75cm]{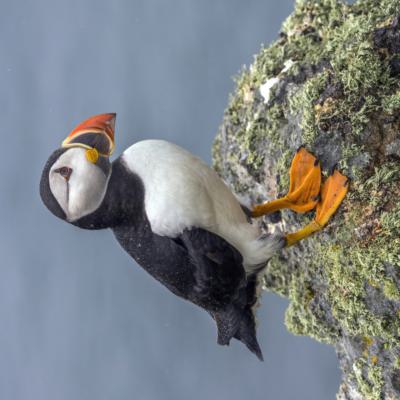}};
  \node[bluevar] (y) at ($(xt) + (2.5,0)$) {$y_i^{(j)}$};

  \node[inner sep=0mm] (p) at ($(y) + (2,0)$) {\footnotesize $\begin{pmatrix} 0 \\ 1 \\ 0 \\ 0 \end{pmatrix}$};
  \node at ($(p) + (0.9,0)$)  {\footnotesize\textcolor{nicegray}{$\begin{matrix} 0\degree \\ 90\degree \\ 180\degree \\ 270\degree \end{matrix}$}};
  
  \node[nicelabel] (view) at ($(xt) + (0,1.25)$) {view};
  \node[nicelabel] (repr) at ($(y) + (0,1.25)$) {represent};
  \node[nicelabel] (pred) at ($(p) + (0,1.25)$) {predict};

  \begin{pgfonlayer}{back}
    \begin{scope}[transform canvas]  
      \draw[->,nicearrow] (x.east) -- node[midway,above,nicelabel] {$t^{(j)}$} (xt.west);
      \draw[->,nicearrow] (xt) -- node[midway,above,nicelabel] {$f_\theta$} (y);
      \draw[->,nicearrow] (y) -- node[midway,above,nicelabel] {$q_\psi$} (p);
    \end{scope}
  \end{pgfonlayer}
\end{tikzpicture}

%% file: sections/jigsaw.tex
\subsection{Solving Jigsaw Puzzles} \label{sec:jigsaw}

\begin{figure}
  \centering \input{figures/jigsaw.tex}
  \caption{Jigsaw extracts patches from an image that are then permuted. The
    pretext task is to find the permutation that was used to permute the images.
    The context-free network $f_\theta$ processes each patch separately and the
    representations are only joined in the latter layers.}
  \label{fig:jigsaw}
\end{figure}
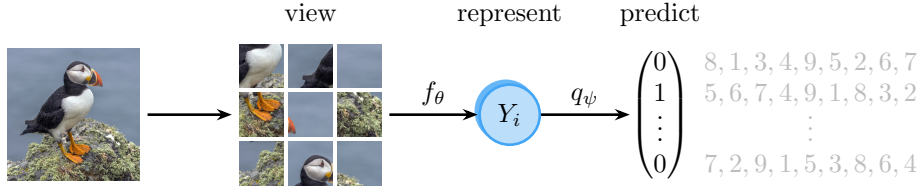

Jigsaw~\citep{noroozi2016unsupervised} is similar to RotNet in the sense that it
also solves a classification task. However, instead of rotating the image,
the transformation consists of randomly permuting several patches of the image
like a jigsaw puzzle. The pretext task of the model is then to predict the class
of the permutation that was used to shuffle the patches. To facilitate the task,
it is necessary to restrict the used permutations to a subset of all
permutations. In their work the authors use \num{1000} predefined permutations (instead
of 9! = \num{362880} for a $3 \times 3$ grid of tiles).

From an input image $x_i$  nine non-overlapping randomly permuted patches
${[x_i^{(1)}, \ldots, x_i^{(9)}]}$ are extracted, where the order of the patches
follows one of the predefined permutations. After that, the Siamese encoder
$f_\theta$ converts each patch into a representation ${y_i^{(j)} =
f(x_i^{(j)})}$. The predictor $q_\psi$ is used to predict the index $c_i$ of the
permutation that was applied to the original image, given all patch
representations $Y_i = [y_i^{(1)}, \ldots, y_i^{(9)}]$ at once. The networks are
trained by minimizing the loss
\begin{align}
  \label{eq:jigsaw}
  \mathcal{L}^\text{Jigsaw}_{\theta,\psi} & = \frac{1}{n} \sum_{i=1}^n \dsoftmax(q_\psi(Y_i), c_i)
\end{align}
between the class scores and the index of the used permutation $c_i$.
The encoder $f_\theta$ is implemented as a truncated AlexNet. The
representations $Y_i$ are concatenated to form the input of the classification
head $q_\psi$, which is implemented as a multi-layer perceptron (MLP). After training, the classification
head is discarded and the encoder is used to obtain image representations for other downstream task.

%% file: figures/jigsaw.tex
\begin{tikzpicture}
  \begin{scope}[x=0.65cm,y=0.65cm]
    \node at (-1,1) {\includegraphics[width=0.58cm]{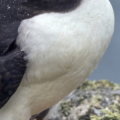}};
    \node[inner sep=1mm] (xtleft) at (-1,0) {\includegraphics[width=0.58cm]{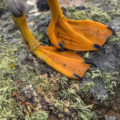}};
    \node at (-1,-1) {\includegraphics[width=0.58cm]{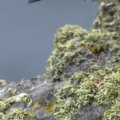}};
    \node at (0,1) {\includegraphics[width=0.58cm]{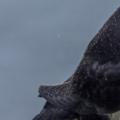}};
    \node (xt) at (0,0) {\includegraphics[width=0.58cm]{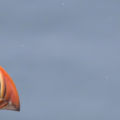}};
    \node at (0,-1) {\includegraphics[width=0.58cm]{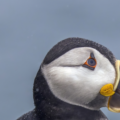}};
    \node at (1,1) {\includegraphics[width=0.58cm]{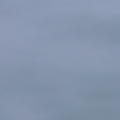}};
    \node[inner sep=1mm] (xtright) at (1,0) {\includegraphics[width=0.58cm]{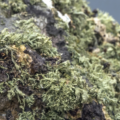}};
    \node at (1,-1) {\includegraphics[width=0.58cm]{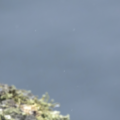}};
  \end{scope}

  \node[inner sep=1mm] (x) at ($(xtleft) - (2.5,0)$) {\includegraphics[width=1.75cm]{figures/jigsaw/puffin.jpg}};
  \node[bluevar] (y) at ($(xtright.east) + (1.6,0)$) {$Y_i$};
  \node[inner sep=0mm] (p) at ($(y) + (2,0)$) {\footnotesize $\begin{pmatrix} 0 \\ 1 \\[1mm] \vphantom{0}\smash{\vdots} \\ 0 \end{pmatrix}$};
  \node at ($(p) + (2,0)$) {\footnotesize\textcolor{nicegray}{$\begin{matrix} 8,1,3,4,9,5,2,6,7 \\ 5,6,7,4,9,1,8,3,2 \\[1mm] \vphantom{0}\smash{\vdots} \\ 7,2,9,1,5,3,8,6,4 \end{matrix}$}};

  \node[nicelabel] (view) at ($(xt) + (0,1.35)$) {view};
  \node[nicelabel] (repr) at ($(y) + (0,1.35)$) {represent};
  \node[nicelabel] (pred) at ($(p) + (0,1.35)$) {predict};

  \begin{pgfonlayer}{back}
    \begin{scope}[transparency group,opacity=0.8]
      \node[bluevar,fill=nicedarkblue] at ($(y) + (-0.05,0.05)$) {};
    \end{scope}

    \begin{scope}[transform canvas]  
      \draw[->,nicearrow] (x) -- (xtleft.west);
      \draw[->,nicearrow] (xtright.east) -- node[midway,above,nicelabel] {$f_\theta$} ($(y.west) - (0.025,0)$);
      \draw[->,nicearrow] (y) -- node[midway,above,nicelabel] {$q_\psi$} (p);
    \end{scope}
  \end{pgfonlayer}
\end{tikzpicture}

%% file: sections/mae.tex
\subsection{Masked Autoencoders (MAE)}

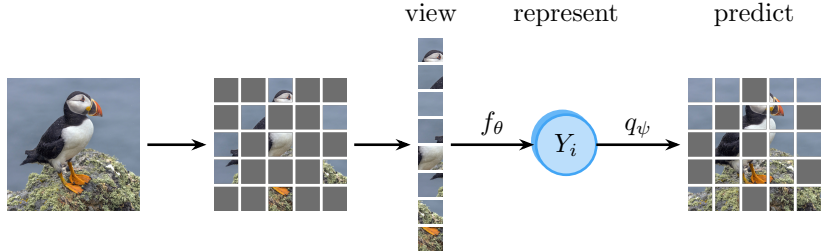
\begin{figure}
  \centering \input{figures/mae.tex}
  \caption{The masked autoencoder predicts masked patches using vision transformers for the encoder and predictor.}
  \label{fig:mae}
\end{figure}

Masked autoencoders \citep{he2022masked} are similar to denoising autoencoders,
where the input images are corrupted and the autoencoder tries to reconstruct
the original image. More specifically, the input image is split into smaller non-overlapping patches,
from which a random subset is masked out.

Given a batch of images $X$, we consider a single image $x_i$. The image is
split up into $m$ patches $X_i = {[x_i^{(1)}, \ldots, x_i^{(m)}]}$, some of
which will be randomly chosen to be masked. We call the set of indices of the
masked patches $M_i^\text{mask}$ and the set of unmasked indices
$M_i^\text{keep}$. The encoder $f_\theta$ converts the \emph{unmasked} patches
into patch representations $Y_i = {[y_i^{(j)} : j \in M_i^\text{keep}]} =
f_\theta({[ x_i^{(j)} : j \in M_i^\text{keep} ]})$, implemented as a vision
transformer \citep{dosovitskiy2021an}. The predictor $q_\psi$ is another vision
transformer that predicts the \emph{masked} patches $\hat{X}_i =
{[\hat{x}_i^{(j)} : j \in M_i^\text{mask}]} = {q_\psi(Y_i, M_i^\text{mask})}$
from $Y_i$ with a mask token for each index in $M_i^\text{mask}$. See
Figure~\ref{fig:mae} for an illustration. The loss is the mean squared error
between the pixels of the predicted patches and the masked patches
\begin{align}
  \mathcal{L}^\text{MAE}_{\theta,\psi} = \frac{1}{n} \sum_{i=1}^n \sum_{j \,\in\, M_i^\text{mask}} \dmse\!\left( \hat{x}_i^{(j)}, x_i^{(j)} \right)\!.
\end{align}

Without precaution the model could ``cheat'' by predicting image patches from
neighboring pixels, since the information in natural images is usually very
spatially redundant. To learn good representations, it is crucial that the
masking ratio is very high, e.g., $75\%$, to encourage the encoder to extract
more high-level features.

%% file: figures/mae.tex
\begin{tikzpicture}
  \node[inner sep=1mm] (x) at (-3,0) {\includegraphics[height=1.75cm]{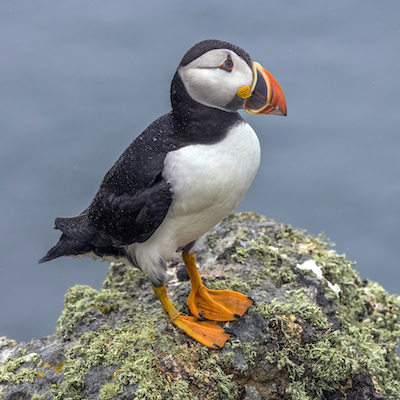}};
  \node[inner sep=1mm] (xt) at ($(x) + (2.75,0)$) {\includegraphics[height=1.75cm]{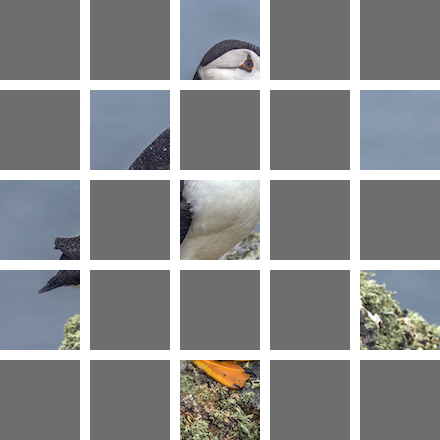}};
  \node[inner sep=1mm] (xt-inputs) at ($(xt) + (2,0)$) {\includegraphics[height=2.824cm]{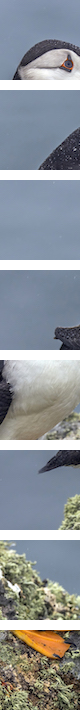}};
  \node[bluevar] (y) at ($(xt-inputs) + (1.8,0)$) {$Y_i$};
  \node[inner sep=1mm] (p) at ($(y) + (2.5,0)$) {\includegraphics[height=1.75cm]{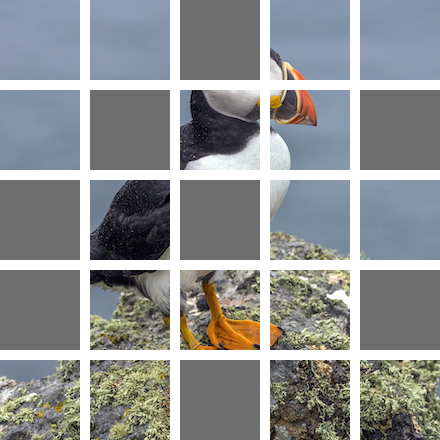}};

  \draw[->,nicearrow] (x) --  (xt);
  \draw[->,nicearrow] (xt) --  (xt-inputs);
  \draw[->,nicearrow] (xt-inputs) -- node[midway,above,nicelabel] {$f_\theta$} ($(y.west) - (0.025,0)$);
  \draw[->,nicearrow] (y) -- node[midway,above,nicelabel] {$q_\psi$} (p);

  \node[nicelabel] (view) at ($(xt-inputs) + (0.0,1.75)$) {\footnotesize view};
  \node[nicelabel] (repr) at ($(y) + (0,1.75)$) {\footnotesize represent};
  \node[nicelabel] (pred) at ($(p) + (0,1.75)$) {\footnotesize predict};

  \begin{pgfonlayer}{back}
    \begin{scope}[transparency group,opacity=0.8]
      \node[bluevar,fill=nicedarkblue] at ($(y) + (-0.05,0.05)$) {};
    \end{scope}
  \end{pgfonlayer}
\end{tikzpicture}

%% file: sections/information-max.tex
\section{Information Maximization Methods}
\label{sec:infomax}

Many self-supervised representation learning methods make use of image
transformations. Jigsaw and Rotation Networks, two approaches from the preceding
section, apply selected transformations to image examples with the aim to
predict the transformation's parametrization. In contrast, the following methods
focus on learning representations that are invariant to certain transformations.
Such a task typically entails a popular failure mode, called
\textit{representation collapse}. It commonly describes trivial solutions, e.g.,
constant representations, that satisfy the invariance objective, but provide
little to no informational value for actual downstream tasks. Another
perspective on the representation collapse is to view it as an information
collapse, concentrating a majority of the probability mass of the embedding in a
single point, which leads to a decrease of information content.

To avoid the representation collapse the so-called \emph{information maximization}
methods have been developed.  They form a class of representation techniques
that focus on the information content of the embeddings
\citep{zbontar2021barlow, bardes2021vicreg, ermolov2021whitening}.  For
instance, some methods explicitly decorrelate all elements of embedding vectors.
This effectively avoids collapse and results in an indirect maximization of
information content. In the following, we present methods that implement this
technique using the normalized cross-correlation matrix of embeddings across
views \citep{zbontar2021barlow}, the covariance matrix for single views
\citep{bardes2021vicreg}, as well as a whitening operation
\citep{ermolov2021whitening}.

\begin{figure}
  \centering
  \begin{subfigure}{0.23\textwidth}
    \includegraphics[width=\linewidth]{figures/transformations/puffin.jpg}
    \caption{Original image}
  \end{subfigure}\hfill
  \begin{subfigure}{0.23\textwidth}
    \includegraphics[width=\linewidth]{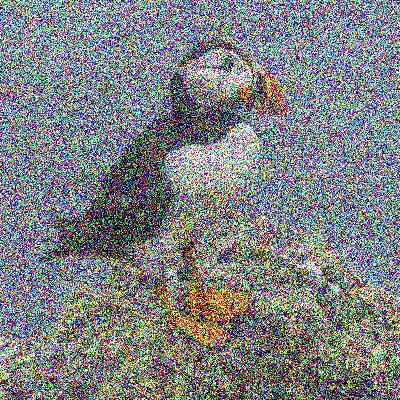}
    \caption{Gaussian noise}
  \end{subfigure}\hfill
  \begin{subfigure}{0.23\textwidth}
    \includegraphics[width=\linewidth]{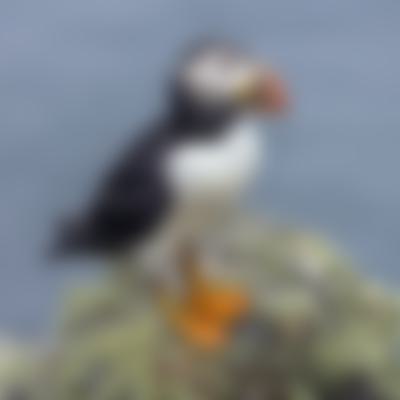}
    \caption{Gaussian blur}
  \end{subfigure}\hfill
  \begin{subfigure}{0.23\textwidth}
    \includegraphics[width=\linewidth]{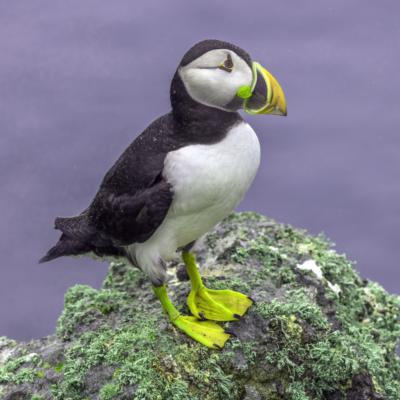}
    \caption{Color jitter}
  \end{subfigure}\vspace{0.2cm}\\
  \begin{subfigure}{0.23\textwidth}
    \includegraphics[width=\linewidth]{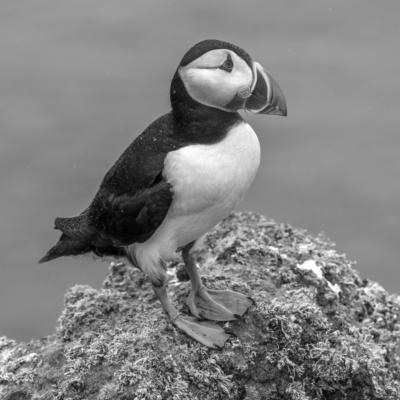}
    \caption{Grayscale}
  \end{subfigure}\hfill
  \begin{subfigure}{0.23\textwidth}
    \includegraphics[width=\linewidth]{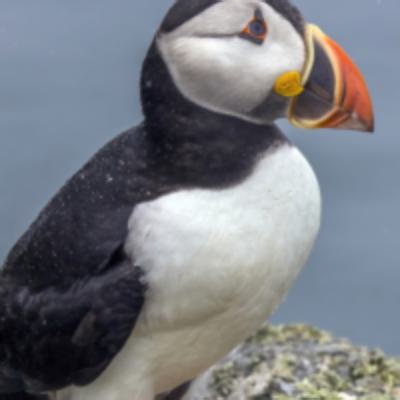}
    \caption{Random Crop}
  \end{subfigure}\hfill
  \begin{subfigure}{0.23\textwidth}
    \includegraphics[width=\linewidth]{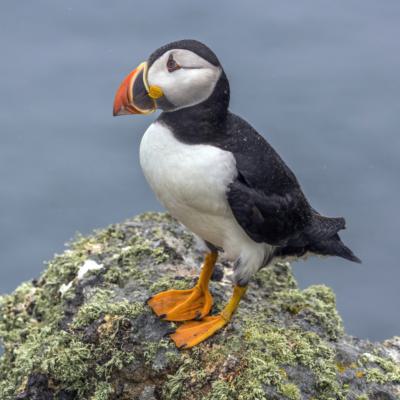}
    \caption{Flip}
  \end{subfigure}\hfill
  \begin{subfigure}{0.23\textwidth}
    \includegraphics[width=\linewidth]{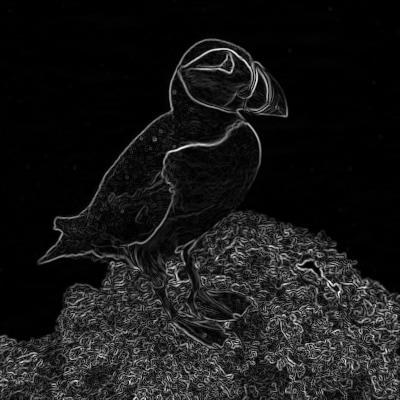}
    \caption{Sobel filter}
  \end{subfigure}
  \caption{Example transformations applied to an image of a puffin.}
  \label{fig:transformations}
\end{figure}

\paragraph{Transformations.}

The main idea of information maximization methods is that the learned representations should be invariant with respect to certain transformations, i.e., the original image and the transformed images should yield the same representations.
We have already encountered two transformations in the previous section, the rotation and jigsaw transformation.  There are many more transformations possible:  the following transformations have been proven useful for the next methods to be described.
\begin{enumerate}
  \item Horizontal flipping: Most natural images can be flipped horizontally
        without changing the semantics, e.g., an image of a car still shows a
        car after being flipped. Vertical flipping can cause problems when for
        example the sky is suddenly at the bottom of the image.
  \item Blurring: Convolving an image with a Gaussian filter is another way to
        transform an image.
  \item Adding Gaussian noise: Learned representations should also be (to some
        extent) invariant to the application of noise.
  \item Sobel filter: Applying a Sobel filter to an image highlights the edges
        of an image. These edges usually still contain a lot of relevant
        information about the image.
  \item Cropping and resizing: Scaling the image to a different size should also
        keep the semantic information.
  \item Color jittering: Changing the contrast, brightness and hue of an image
        yields another instance of an image that shows the same content.
  \item Grayscaling: Converting color-images to grayscale images is closely
        related to color jittering.
\end{enumerate}
Note that these image transformations are closely related to dataset augmentation techniques used in supervised learning \citep{yang2022image,shorten2019survey}.

\input{sections/barlow.tex}
\input{sections/vicreg.tex}
\input{sections/wmse.tex}

%% file: sections/barlow.tex
\subsection{Barlow Twins}

\begin{figure}[ht!]
  \centering \input{figures/barlow.tex}
  \caption{Barlow twins takes two views of the input batch and minimizes the correlation of the projected representations.}
  \label{fig:barlow_twins_arch}
\end{figure}
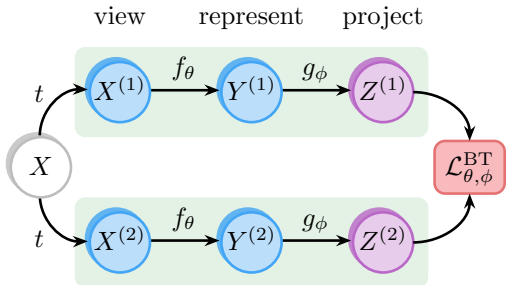

The central idea behind this framework is the principle of redundancy reduction.
This principle was introduced by the neuroscientist \citet{barlow1961possible} and states that the reduction of redundancy is important for the organization
of sensory messages in the brain.

To implement this redundancy reduction principle, the Barlow Twins approach
takes a batch of images $X$ and creates two views ${X^{(1)} = t(X)}$ and
${X^{(2)} = t(X)}$ of these images, where $t\sim\mathcal{T}$ is a transformation
that is randomly sampled from $\mathcal{T}$ for every image and every view. A
Siamese encoder $f_\theta$ computes representations ${Y^{(1)} =
f_\theta(X^{(1)})}$ and ${Y^{(2)} = f_\theta(X^{(2)})}$, which are fed into a
Siamese projector $g_\phi$ to compute projections ${Z^{(1)} = {[ z_1^{(1)},
\ldots, z_n^{(1)} ]} = g_\phi(Y^{(1)})}$ and ${Z^{(2)} = {[ z_1^{(2)}, \ldots,
z_n^{(2)} ]} = g_\phi(Y^{(2)})}$ for both views.

The idea of Barlow Twins is to regularize the cross-correlation matrix between
the projections of both views. The cross-correlation matrix is calculated as
\begin{equation}
  C = \frac{1}{n} \sum_{i=1}^n \left( (z_i^{(1)} - \mu^{(1)}) / \sigma^{(1)} \right) \left( (z_i^{(2)} - \mu^{(2)}) / \sigma^{(2)} \right)^{\!\top}\!,
\end{equation}
where $\mu^{(j)}$ and $\sigma^{(j)}$ are the mean and standard deviation over
the batch of projections of the $j$-th view, calculated as
\begin{align}
  \mu^{(j)}    & = \frac{1}{n} \sum_{i=1}^n z_i^{(j)},                              \\
  \sigma^{(j)} & = \sqrt{\frac{1}{n - 1} \sum_{i=1}^n (z_i^{(j)} - \mu^{(j)})^2}\,.
\end{align}
The loss function is then defined as
\begin{equation}
  \label{eq:barlow}
  \mathcal{L}^\text{BT}_{\theta,\phi}
  = \sum_{k=1}^{d} \left(1 - C[k,k]\right)^2 + \lambda \sum_{k=1}^d \sum_{k' \neq k} C[k,k']^2,
\end{equation}
where $d$ is the number of dimensions of the projection and $\lambda>0$ is a
hyperparameter. The first term promotes invariance with regard to the applied
transformations and the second term decorrelates the learned embeddings, i.e.,
reduces redundancy.  By using this loss, the encoder $f_\theta$ is encouraged to
predict embeddings that are decorrelated and thereby non-redundant. The Barlow
Twins are trained using the LARS optimizer~\citep{you2017large}.  Note that this
loss function is related to the VICReg method, where the first term is called
\emph{variance} term and the second \emph{covariance} term.

%% file: figures/barlow.tex
\begin{tikzpicture}
  \node[whitevar] (x) at (0,0) {$X$};

  \node[bluevar] (xt) at ($(x) + (1.05,1)$) {$X^{(1)}$};
  \node[bluevar] (y) at ($(xt) + (1.75,0)$) {$Y^{(1)}$};
  \node[purplevar] (z) at ($(y) + (1.75,0)$) {$Z^{(1)}$};

  \node[bluevar] (xtp) at ($(xt) - (0,2)$) {$X^{(2)}$};
  \node[bluevar] (yp) at ($(y) - (0,2)$) {$Y^{(2)}$};
  \node[purplevar] (zp) at ($(z) - (0,2)$) {$Z^{(2)}$};

  \node[nicelabel] (view) at ($(xt) + (0,1)$) {view};
  \node[nicelabel] (repr) at ($(y) + (0,1)$) {represent};
  \node[nicelabel] (proj) at ($(z) + (0,1)$) {project};
  
   \node[box, fill=nicered, draw=nicedarkred] (loss) at ($(z) - (-1.15,1)$) {\footnotesize $\mathcal{L}^\text{BT}_{\theta,\phi}$};
   \draw[nicearrow,] (z) edge[->,out=0,in=90,relative=false] (loss);
   \draw[nicearrow,] (zp) edge[->,out=0,in=-90,relative=false] (loss);

  \begin{pgfonlayer}{back}
    \path[box,fill=nicegreen,opacity=0.5] ($(xt) - (0.6,0.6)$) rectangle ($(z) + (0.6,0.6)$);
    \path[box,fill=nicegreen,opacity=0.5] ($(xtp) - (0.6,0.6)$) rectangle ($(zp) + (0.6,0.6)$);

    \begin{scope}[transparency group,opacity=0.8]
      \node[whitevar,fill=nicegray] at ($(x) + (-0.05,0.05)$) {};

      \node[bluevar,fill=nicedarkblue] at ($(xt) + (-0.05,0.05)$) {};
      \node[bluevar,fill=nicedarkblue] at ($(y) + (-0.05,0.05)$) {};
      \node[purplevar,fill=nicedarkpurple] at ($(z) + (-0.05,0.05)$) {};

      \node[bluevar,fill=nicedarkblue] at ($(xtp) + (-0.05,0.05)$) {};
      \node[bluevar,fill=nicedarkblue] at ($(yp) + (-0.05,0.05)$) {};
      \node[purplevar,fill=nicedarkpurple] at ($(zp) + (-0.05,0.05)$) {};
    \end{scope}

    \begin{scope}[transform canvas={xshift=-0.025cm,yshift=0.025cm}]
      \draw[nicearrow] (x) edge[->,bend left=45] node[midway,left,yshift=1mm,nicelabel] {$t$} (xt);
      \draw[->,nicearrow] (xt) -- node[midway,above,nicelabel] {$f_\theta$} (y);
      \draw[->,nicearrow] (y) -- node[midway,above,nicelabel] {$g_\phi$} (z);

      \draw[nicearrow] (x) edge[->,bend right=45] node[midway,left,yshift=-1.5mm,nicelabel] {$t$} (xtp);
      \draw[->,nicearrow] (xtp) -- node[midway,above,nicelabel] {$f_\theta$} (yp);
      \draw[->,nicearrow] (yp) -- node[midway,above,nicelabel] {$g_\phi$} (zp);
    \end{scope}
  \end{pgfonlayer}
\end{tikzpicture}

%% file: sections/vicreg.tex
\subsection{Variance-Invariance-Covariance Regularization (VICReg)}

\begin{figure}[ht!]
  \centering \input{figures/vicreg.tex}
  \caption{VICReg takes two views of the input batch and minimizes the mean squared error between the projected representations, while regularizing the covariance matrices of representations from individual views to avoid representation collapse.}
  \label{fig:vigreg_arch}
\end{figure}
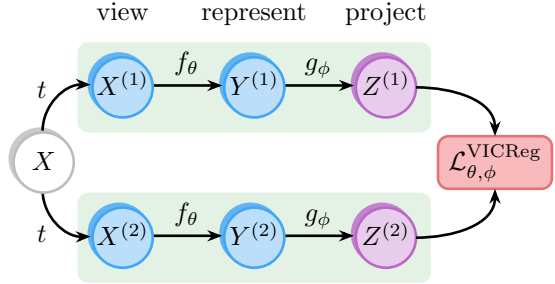

VICReg was introduced by \cite{bardes2021vicreg} and is a joint-embedding
architecture that falls into the category of information maximization methods.
Figure \ref{fig:vigreg_arch} gives an overview of the architecture, which is identical to Barlow twins, but uses a different loss function.  It aims to
maximize agreement between representations of different views of an input, while
preventing \textit{informational collapse} using two additional regularization
terms. Specifically, VICReg defines regularization terms for
variance, invariance and covariance.

Given a batch of images $X$, two views ${X^{(1)} = t(X)}$ and ${X^{(2)} = t(X)}$
are defined, where $t\sim\mathcal{T}$ is, again, randomly sampled from $\mathcal{T}$ for every image and every view. A Siamese encoder $f_\theta$ computes representations ${Y^{(1)} = f_\theta(X^{(1)})}$ and
${Y^{(2)} = f_\theta(X^{(2)})}$, which are fed into a Siamese projector $g_\phi$
to compute projections $Z^{(1)} = {[ z_1^{(1)}, \ldots, z_n^{(1)} ]} =
  g_\phi(Y^{(1)})$ and $Z^{(2)} = {[ z_1^{(2)}, \ldots, z_n^{(2)} ]} =
  g_\phi(Y^{(2)})$. Each projection has $d$ dimensions.
For each view, the covariance matrix of the projections is
computed, which is defined as
\begin{equation}
  C^{(j)} = \frac{1}{n-1} \sum_{i=1}^n \left( z_i^{(j)} - \mu^{(j)} \right) \left( z_i^{(j)} - \mu^{(j)} \right)^{\!\top}\!,
\end{equation}
where $\mu^{(j)}$ is the mean over the batch of projections of the $j$-th view,
i.e.,
\begin{equation}
  \mu^{(j)} = \frac{1}{n} \sum_{i=1}^n z_i^{(j)}.
\end{equation}

The \emph{variance} term aims to keep the standard deviation of each element
of the embedding across the batch dimension above a margin $b$. Practically,
this prevents embedding vectors to be the same across the batch and thus is one
of the two mechanisms that intent to prevent collapse. It can be implemented
using a hinge loss
\begin{equation}
  \ell_\text{V}(Z^{(j)}) = \frac{1}{d} \sum_{k=1}^d \max\!\left(0, b - \sqrt{C^{(j)}[k,k] + \varepsilon}\,\right)\!.
\end{equation}
where ${\varepsilon > 0}$ is a small hyperparameter for numerical stability.
\cite{bardes2021vicreg} used ${b = 1}$. On that note, the variance term is
closely related to the invariance term of Barlow Twins
\citep{zbontar2021barlow}, but applied with a different intention. While Barlow
Twins practically maximizes the squared diagonals of the normalized
cross-correlation matrix to encourage correlation of embedding elements across
views, VICReg maximizes the square root of the diagonals of the covariance
matrix of single views in order to prevent collapse. Note, that the maximization
in Barlow Twins is restricted by the preceding normalization of the embeddings.
As VICReg does not apply a normalization, the margin loss is used to restrict
this optimization.

The \emph{covariance} term decorrelates elements of embedding vectors for single
views in order to reduce redundancy and prevent collapse. This is achieved by
minimizing the squared off-diagonal elements of the covariance matrix $C^{(j)}$
towards $0$, i.e.,
\begin{equation}
  \ell_\text{C}(Z^{(j)}) = \frac{1}{d} \sum_{k=1}^d \sum_{k' \neq k} \left(C^{(j)}[k,k']\right)^2\!.
\end{equation}
Note, that this is similar to the redundancy reduction term used in Barlow Twins
(Equation \ref{eq:barlow}, right summand), the main difference being again that
Barlow Twins applies it across views, but with a similar intention.

Finally, the \emph{invariance} term is used to maximize the agreement between
two projections $z_i^{(1)}$ and $z_i^{(2)}$ of the same image, thus inducing
invariance to the transformations that were applied to $x_i$. For this,
\cite{bardes2021vicreg} apply the mean squared error between the projections
\begin{equation}
  \ell_\text{I}(Z^{(1)},Z^{(2)}) = \frac{1}{n} \sum_{i=1}^n \dmse\!\left( z_i^{(1)}, z_i^{(2)} \right)\!.
\end{equation}
Notably, it is the only loss term in VICReg operating across different views.

Overall, the loss of VICReg can be defined as the weighted sum of all three
regularizations for the given views
\begin{equation}
  \mathcal{L}^\text{VICReg}_{\theta,\,\phi}(X) = \lambda_\text{V} [\ell_\text{V}(Z^{(1)}) + \ell_\text{V}(Z^{(2)})] + \lambda_\text{C}[\ell_\text{C}(Z^{(1)}) + \ell_\text{C}(Z^{(2)})] + \lambda_\text{I}\ell_\text{I}(Z^{(1)}, Z^{(2)}),
\end{equation}
where ${\lambda_\text{V}, \lambda_\text{I},\lambda_\text{C} > 0}$ are
hyperparameters that balance the individual losses.

%% file: figures/vicreg.tex
\begin{tikzpicture}
  \node[whitevar] (x) at (0,0) {$X$};

  \node[bluevar] (xt) at ($(x) + (1.05,1)$) {$X^{(1)}$};
  \node[bluevar] (y) at ($(xt) + (1.75,0)$) {$Y^{(1)}$};
  \node[purplevar] (z) at ($(y) + (1.75,0)$) {$Z^{(1)}$};

  \node[bluevar] (xtp) at ($(xt) - (0,2)$) {$X^{(2)}$};
  \node[bluevar] (yp) at ($(y) - (0,2)$) {$Y^{(2)}$};
  \node[purplevar] (zp) at ($(z) - (0,2)$) {$Z^{(2)}$};

  \node[nicelabel] (view) at ($(xt) + (0,1)$) {view};
  \node[nicelabel] (repr) at ($(y) + (0,1)$) {represent};
  \node[nicelabel] (proj) at ($(z) + (0,1)$) {project};

	\node[box, fill=nicered, draw=nicedarkred] (loss) at ($(z) - (-1.45,1)$) {\footnotesize$\mathcal{L}^\text{VICReg}_{\theta, \phi}$};

	\draw[nicearrow] (z) edge[->,out=0,in=90,relative=false] (loss);
	\draw[nicearrow,] (zp) edge[->,out=0,in=-90,relative=false] (loss);

  \begin{pgfonlayer}{back}
    \path[box,fill=nicegreen,opacity=0.5] ($(xt) - (0.6,0.6)$) rectangle ($(z) + (0.6,0.6)$);
    \path[box,fill=nicegreen,opacity=0.5] ($(xtp) - (0.6,0.6)$) rectangle ($(zp) + (0.6,0.6)$);
  
    \begin{scope}[transparency group,opacity=0.8]
      \node[whitevar,fill=nicegray] at ($(x) + (-0.05,0.05)$) {};

      \node[bluevar,fill=nicedarkblue] at ($(xt) + (-0.05,0.05)$) {};
      \node[bluevar,fill=nicedarkblue] at ($(y) + (-0.05,0.05)$) {};
      \node[purplevar,fill=nicedarkpurple] at ($(z) + (-0.05,0.05)$) {};

      \node[bluevar,fill=nicedarkblue] at ($(xtp) + (-0.05,0.05)$) {};
      \node[bluevar,fill=nicedarkblue] at ($(yp) + (-0.05,0.05)$) {};
      \node[purplevar,fill=nicedarkpurple] at ($(zp) + (-0.05,0.05)$) {};
    \end{scope}

    \begin{scope}[transform canvas={xshift=-0.025cm,yshift=0.025cm}]
      \draw[nicearrow] (x) edge[->,bend left=45] node[midway,left,yshift=1mm,nicelabel] {$t$} (xt);
      \draw[->,nicearrow] (xt) -- node[midway,above,nicelabel] {$f_\theta$} (y);
      \draw[->,nicearrow] (y) -- node[midway,above,nicelabel] {$g_\phi$} (z);

      \draw[nicearrow] (x) edge[->,bend right=45] node[midway,left,yshift=-1.5mm,nicelabel] {$t$} (xtp);
      \draw[->,nicearrow] (xtp) -- node[midway,above,nicelabel] {$f_\theta$} (yp);
      \draw[->,nicearrow] (yp) -- node[midway,above,nicelabel] {$g_\phi$} (zp);
    \end{scope}
  \end{pgfonlayer}
\end{tikzpicture}

%% file: sections/wmse.tex
\subsection{Self-Supervised Representation Learning using Whitening (WMSE)}

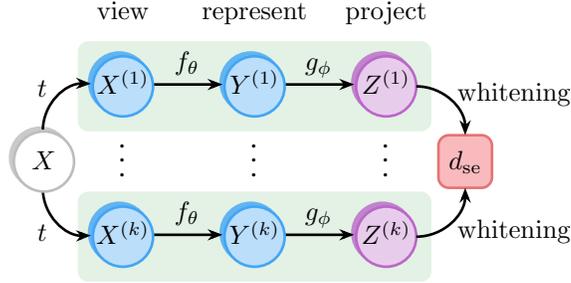
\begin{figure}
  \centering
  \input{figures/wmse.tex}
  \caption{WMSE training: The batch of input images $X$ are randomly transformed and fed into the encoder network $f_\theta$. The representations are then projected using the projection head $g_\phi$. Next, whitening is applied to the projections. The networks are trained by minimizing the normalized MSE between the projections.}
  \label{fig:wsme_pipeline}
\end{figure}

Whitening linearly transforms a set of data points, such that the resulting data
points are decorrelated and have unit variance, i.e., the covariance matrix becomes the identity matrix. The method WMSE
\citep{ermolov2021whitening} applies this idea to the embeddings of images to
prevent the representation collapse.

Given a batch of images $X$, random transformations are applied to obtain $m$
views ${X^{(j)}}$ for all ${j \in \{1, \ldots, m\}}$. A Siamese encoder
$f_\theta$ maps the views onto representations ${Y^{(j)} = f_\theta(X^{(j)})}$,
which are then fed into a Siamese projector $g_\phi$ to compute projections
$Z^{(j)} = {[z_1^{(j)}, \ldots, z_n^{(j)}]} = g_\phi(Y^{(j)})$. All projections
are then concatenated into a single matrix $Z = {[z_1^{(1)}, \ldots, z_n^{(1)},
\ldots, z_1^{(m)}, \ldots, z_n^{(m)}]}$. This matrix is \emph{whitened} to
obtain $\tilde{Z}$ by removing the mean and decorrelating it using the Cholesky
decomposition of the inverse covariance matrix, i.e.,
\begin{equation}
  \tilde{Z} = [\tilde{z}_1^{(1)}, \ldots, \tilde{z}_n^{(1)}, \ldots, \tilde{z}_1^{(m)}, \ldots, \tilde{z}_n^{(m)}] = W_Z \!\left(Z - \frac{1}{nm}\sum_{i=1}^n\sum_{j=1}^m z^{(j)}_i \, \mathbf{1}_{nm}^\top \right)\!,
\end{equation}
where $\mathbf{1}_{nm}$ is an all-ones vector with $nm$ entries and ${W_Z}$ is the Cholesky factor of the inverse covariance matrix $P^{ZZ}$
(also known as the precision matrix), i.e., ${W_Z W_Z^\top = P^{ZZ}}$.
Note that the Cholesky decomposition is differentiable which allows to
backpropagate through it during training.

To train the models, the normalized squared error between all pairs of the
whitened projections is minimized, i.e., the loss function is defined as
\begin{equation}
  \label{unconstraint_wmse}
  \mathcal{L}^\text{WMSE}_{\theta,\phi} = \frac{1}{n} \sum_{i=1}^n \frac{2}{m (m-1)} \sum_{j=1}^m \sum_{k=j + 1}^m \dnormmse\!\left(\tilde{z}^{(j)}_i,\tilde{z}^{(k)}_i\right)\!.
\end{equation}
The constant $2/(m (m-1))$ is the number of comparison per image.
The whitening step is essential to prevent the representations from collapsing.
The objective maximizes the similarity between all augmented pairs while
also preventing representation collapse by enforcing unit covariance on the projections.

\paragraph{Batch slicing.} One problem of the original method is that the loss has a large variance over consecutive training batches.  To counteract this issue, \citet{ermolov2021whitening} employ so-called batch slicing: the idea of batch slicing is that different views of the same image $z_i^{(1)}, z_i^{(2)}, \dots, z_i^{(m)}$ should not be in the same batch when the whitening matrix is calculated. For this, $Z$  is partitioned into $m$ parts. The elements of each part are then permuted using the same permutation for each of the $m$ parts. Finally, each of the parts is further subdivided into $d$ subsets which are then used to calculate the whitening matrix for that specific subset. In that way, the loss minimization is dependent on $m \cdot d$ covariance matrices that need to satisfy the identity, leading empirically to lower variance.

%% file: figures/wmse.tex
\begin{tikzpicture}
  \node[whitevar] (x) at (0,0) {$X$};

  \node[bluevar] (xt) at ($(x) + (1.05,1)$) {$X^{(1)}$};
  \node[bluevar] (y) at ($(xt) + (1.75,0)$) {$Y^{(1)}$};
  \node[purplevar] (z) at ($(y) + (1.75,0)$) {$Z^{(1)}$};

  \node at ($(xt) - (0,0.95)$) {\rotatebox{90}{$\cdots$}};
  \node at ($(y) - (0,0.95)$) {\rotatebox{90}{$\cdots$}};
   \node at ($(z) - (0,0.95)$) {\rotatebox{90}{$\cdots$}};

  \node[bluevar] (xtp) at ($(xt) - (0,2)$) {$X^{(k)}$};
  \node[bluevar] (yp) at ($(y) - (0,2)$) {$Y^{(k)}$};
  \node[purplevar] (zp) at ($(z) - (0,2)$) {$Z^{(k)}$};

  \node[nicelabel] (view) at ($(xt) + (0,1)$) {view};
  \node[nicelabel] (repr) at ($(y) + (0,1)$) {represent};
  \node[nicelabel] (proj) at ($(z) + (0,1)$) {project};
  
  \node[box, fill=nicered, draw=nicedarkred] (loss) at ($(z) - (-1.05,1)$) {\footnotesize $\dmse$};
  
  \draw[nicearrow] (z) edge[->,out=0,in=90,relative=false] node[midway,right,nicelabel,yshift=1mm, xshift=-0.5mm] {whitening} (loss);
  \draw[nicearrow,] (zp) edge[->,out=0,in=-90,relative=false] node[midway,right,nicelabel,yshift=-1mm, xshift=-0.5mm] {whitening}  (loss);

  \begin{pgfonlayer}{back}
    \path[box,fill=nicegreen,opacity=0.5] ($(xt) - (0.6,0.6)$) rectangle ($(z) + (0.6,0.6)$);
    \path[box,fill=nicegreen,opacity=0.5] ($(xtp) - (0.6,0.6)$) rectangle ($(zp) + (0.6,0.6)$);

    \begin{scope}[transparency group,opacity=0.8]
      \node[whitevar,fill=nicegray] at ($(x) + (-0.05,0.05)$) {};

      \node[bluevar,fill=nicedarkblue] at ($(xt) + (-0.05,0.05)$) {};
      \node[bluevar,fill=nicedarkblue] at ($(y) + (-0.05,0.05)$) {};
      \node[purplevar,fill=nicedarkpurple] at ($(z) + (-0.05,0.05)$) {};

      \node[bluevar,fill=nicedarkblue] at ($(xtp) + (-0.05,0.05)$) {};
      \node[bluevar,fill=nicedarkblue] at ($(yp) + (-0.05,0.05)$) {};
      \node[purplevar,fill=nicedarkpurple] at ($(zp) + (-0.05,0.05)$) {};
    \end{scope}

    \begin{scope}[transform canvas={xshift=-0.025cm,yshift=0.025cm}]
      \draw[nicearrow] (x) edge[->,bend left=45] node[midway,left,nicelabel,yshift=1mm] {$t$} (xt);
      \draw[->,nicearrow] (xt) -- node[midway,above,nicelabel] {$f_\theta$} (y);
      \draw[->,nicearrow] (y) -- node[midway,above,nicelabel] {$g_\phi$} (z);

      \draw[nicearrow] (x) edge[->,bend right=45] node[midway,left,nicelabel,yshift=-1.5mm] {$t$} (xtp);
      \draw[->,nicearrow] (xtp) -- node[midway,above,nicelabel] {$f_\theta$} (yp);
      \draw[->,nicearrow] (yp) -- node[midway,above,nicelabel] {$g_\phi$} (zp);
    \end{scope}
  \end{pgfonlayer}
\end{tikzpicture}

%% file: sections/distillation.tex
\section{Teacher-Student Methods}
\label{sec:distillation}
Methods based on teacher-student learning are closely related to information
maximization methods. Similar to information maximization methods, the student
learns to predict the teacher's representations across different image
transformations. This allows the student to learn invariant representations that
are robust to different transformations of the same image. These methods consist
of two branches, where one is considered the student and the other the teacher.
To prevent representational collapse as defined in Section \ref{sec:infomax},
the teacher provides stable target representations for the student to predict.
To provide stable targets, the teacher is not updated using gradient descent and
its parameters are fixed when updating the student. Sometimes a momentum encoder
is used between the teacher and student to update the fixed targets. That is,
the weights from the student are slowly copied to the teacher to provide more
recent targets. The teacher usually has the same architecture as the student,
but does not necessarily have the same parameters. The teacher can be a running
average of the student's representations, where, e.g., a momentum encoder is used
to update the teacher network with the student's weights. For some
teacher-student methods the teacher shares the student's weights and an
additional predictor network has to predict the teacher's representation.

\input{sections/byol.tex} 
\input{sections/dino.tex}

\input{sections/esvit.tex}
\input{sections/simsiam.tex}

%% file: sections/byol.tex
\subsection{Bootstrap Your Own Latent (BYOL)}
\label{sec:byol}

BYOL \citep{grill2020bootstrap} is inspired by the observation that learning
representations by predicting fixed representations from a randomly initialized
target network avoids representational collapse albeit with subpar performance.
This naturally entails a teacher-student architecture where the teacher (target
network) provides stable representations for the student (online network) to
learn on.

\begin{figure}
  \centering \input{figures/byol.tex}
  \caption{BYOL consists of a student and teacher network. The teacher network
    is not updated via gradient descent (stop-gradient) and thus provides stable
    representations for the student network to learn. The teacher network is
    updated iteratively via an exponential moving average of the student. The
    student branch has a predictor which is trained to match the fixed
    representations of the teacher.}
  \label{fig:byol_arch}
\end{figure}
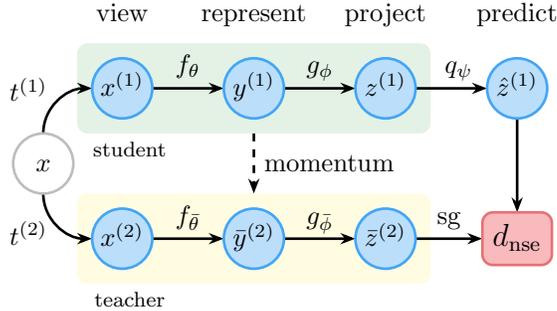
\FloatBarrier

BYOL defines two different networks: a student network and a teacher network.
The architecture is shown in Figure \ref{fig:byol_arch}, the student network and
teacher network consist of the following parts:
\begin{itemize}
\item Student network: encoder $f_\theta$, projector $g_\phi$, predictor
  $q_\psi$
\item Teacher network: encoder $f_{\bar{\theta}}$, projector $g_{\bar{\phi}}$
\end{itemize}
The encoder $f$ and projector $g$ are present in both the student and teacher
networks, whereas the predictor $q$ is only part of the student network.

\paragraph{Learning augmentation-invariant features from the teacher.}
Like information-maximization methods, teacher-student methods learn
representations by applying different transformations to the images (see Section
\ref{sec:infomax}). Given an image $x_i$, BYOL applies randomly-sampled
transformations ${t \sim \mathcal{T}}$ to obtain two different views ${x_i^{(1)}
= t(x_i)}$ and ${x_i^{(2)} = t(x_i)}$. The student network computes
representations ${y_i^{(j)} = f_\theta(x_i^{(j)})}$, projections ${z_i^{(j)} =
g_\phi(y_i^{(j)})}$, and predictions ${\hat{z}_i^{(j)}} = q_\psi(z_i^{(j)})$ for
both views ${j \in \{1, 2\}}$. The views are also fed to the teacher network to
obtain target projections ${\bar{z}_i^{(1)} =
g_{\bar{\phi}}(f_{\bar{\theta}}(x_i^{(1)}))}$ and ${\bar{z}_i^{(2)} =
g_{\bar{\phi}}(f_{\bar{\theta}}(x_i^{(2)}))}$.

BYOL minimizes two normalized squared errors: (i) between the prediction of the
first view and the target projection of the second view, (ii) between the
prediction of the second view and the target projection of the first view. The
final loss function is
\begin{equation}
  \mathcal{L}^{\text{BYOL}}_{\theta,\phi,\psi} = \frac{1}{n} \sum_{i=1}^n \left[ \dnormmse(\hat{z}_i^{(1)}, \bar{z}_i^{(2)}) + \dnormmse(\hat{z}_i^{(2)}, \bar{z}_i^{(1)}) \right]\!.
  \label{eq:loss_byol3}
\end{equation}
Note that the loss is minimal when the cosine similarity between the vectors is
$1$. Thus, representations are learned that are similar for two different
transformations. In other words, the information content in the learned
representations is maximized.

\paragraph{Teacher-student momentum encoder.} At each training step, the loss is
minimized with respect to $\theta$, $\phi$, and $\psi$. That is, only the
weights of the student are updated by the gradient of the loss function using
the LARS optimizer \citep{you2017large}. The weights of the teacher are updated
by the exponential moving average~\citep{lillicrap2019continuous}, i.e.,
\begin{align}
  \bar{\theta} & \leftarrow \tau \bar{\theta} + (1-\tau) \theta, \\
  \bar{\phi} & \leftarrow \tau \bar{\phi} + (1-\tau) \phi,
                               \label{eq:update_byol}
\end{align}
where $\tau \in [0,1]$ controls the rate at which the weights of the teacher
network are updated with the weights of the student network.

The authors show that BYOL's success relies on two key components: (i) keeping
the predictor $q_\psi$ near optimal at all times by predicting the stable target
representations, and (ii) updating the parameters in the direction of
$\nabla_{\theta, \phi, \psi}\mathcal{L}^{\text{BYOL}}$ and not in the direction
of $\nabla_{\bar{\theta}, \bar{\phi}}\mathcal{L}^{\text{BYOL}}$. In other words,
the proposed loss and update do not jointly optimize the loss
over $\theta, \phi$ and $\bar{\theta}, \bar{\phi}$, which
would lead to a representation collapse. \cite{chen2021exploring} provide
further insight on how this is related to the predictor: regarding (i) it is
observed that keeping the learning rate of the predictor fixed instead of
decaying it during training improves performance, which supports the fact that
the predictor should learn the latest representations. Regarding (ii)
\cite{chen2021exploring} use a predictor that maps the projection to the
identity. They argue that the gradient of the symmetrized loss with an identity
predictor, i.e., between the two projections cancels out the stop-gradient
operator. In this case the gradient of the symmetrized loss between the two
projections is in the same direction, hence leading to a collapsed
representation. Note that this analysis was performed for SimSiam
\citep{chen2021exploring}, which does not use a momentum encoder. It is feasible
however that the predictor plays the same role for BYOL.

%% file: figures/byol.tex
\begin{tikzpicture}
  \node[whitevar] (x) at (0,0) {$x$};

  \node[bluevar] (xt) at ($(x) + (1.05,1)$) {$x^{(1)}$};
  \node[bluevar] (y) at ($(xt) + (1.75,0)$) {$y^{(1)}$};
  \node[bluevar] (z) at ($(y) + (1.75,0)$) {$z^{(1)}$};
  \node[bluevar] (p) at ($(z) + (1.75,0)$) {$\hat{z}^{(1)}$};

  \node[bluevar] (xtp) at ($(xt) - (0,2)$) {$x^{(2)}$};
  \node[bluevar] (yp) at ($(y) - (0,2)$) {$\bar{y}^{(2)}$};
  \node[bluevar] (zp) at ($(z) - (0,2)$) {$\bar{z}^{(2)}$};

  \node[box, draw=nicedarkred, fill=nicered] (loss) at ($(zp) + (1.75,0)$) {$\dnormmse$};

  \node[nicelabel] (view) at ($(xt) + (0,1)$) {view};
  \node[nicelabel] (repr) at ($(y) + (0,1)$) {represent};
  \node[nicelabel] (proj) at ($(z) + (0,1)$) {project};
  \node[nicelabel] (pred) at ($(p) + (0,1)$) {predict};

  \node[nicelabel] (student) at ($(xt) + (0.1,-0.8)$) {\scriptsize student};
  \node[nicelabel] (teacher) at ($(xtp) + (0.1,-0.8)$) {\scriptsize teacher};

  \begin{pgfonlayer}{back}
    \path[box,fill=nicegreen,opacity=0.5] ($(xt) - (0.6,0.6)$) rectangle ($(z) + (0.6,0.6)$);
    \path[box,fill=niceyellow,opacity=0.5] ($(xtp) - (0.6,0.6)$) rectangle ($(zp) + (0.6,0.6)$);

    \begin{scope}[transform canvas]  
      \draw[nicearrow] (x) edge[->,out=90,in=180,relative=false] node[midway,left,nicelabel,yshift=1mm] {$t^{(1)}$} (xt);
      \draw[->,nicearrow] (xt) -- node[midway,above,nicelabel] {$f_\theta$} (y);
      \draw[->,nicearrow] (y) -- node[midway,above,nicelabel] {$g_\phi$} (z);
      \draw[->,nicearrow] (z) -- node[midway,above,nicelabel,xshift=1mm] {$q_\psi$} (p);

      \draw[nicearrow] (x) edge[->,out=-90,in=180,relative=false] node[midway,left,nicelabel,yshift=-1.5mm] {$t^{(2)}$} (xtp);
      \draw[->,nicearrow] (xtp) -- node[midway,above,nicelabel] {$f_{\bar{\theta}}$} (yp);
      \draw[->,nicearrow] (yp) -- node[midway,above,nicelabel] {$g_{\bar{\phi}}$} (zp);

      \draw[->,nicearrow] (p) -- (loss);
      \draw[->,nicearrow,] (zp) -- node[midway,above,nicelabel] {sg} (loss);

      \draw[->,nicearrow,dashed] ($(y) - (0,0.6)$) -- node[midway,right,nicelabel] {momentum} ($(yp) + (0,0.6)$);
    \end{scope}
  \end{pgfonlayer}
\end{tikzpicture}

%% file: sections/dino.tex
\subsection{Self-Distillation With No Labels (DINO)} \label{sec:dino}
One of the main contributions of DINO \citep{caron2021emerging} is adapting the
teacher-student architecture closer to the knowledge distillation framework
\citep{gou2021kdsurvey}, where instead of matching the output embeddings
directly, the teacher provides soft labels by applying a softmax operation on
its output. The authors show that this facilitates preventing a representation
collapse.
\begin{figure}
	\centering \input{figures/dino.tex}
	\caption{DINO consists of two ViTs, one acting as a student and the other as
		teacher. The embeddings of the ViT's are transformed into a Softmax
		distribution which produces soft labels for both the student and the
		teacher. The knowledge of the student is then iteratively distilled onto the
		teacher, which provides stable targets for the student.}
	\label{fig:dino_arch}
\end{figure}
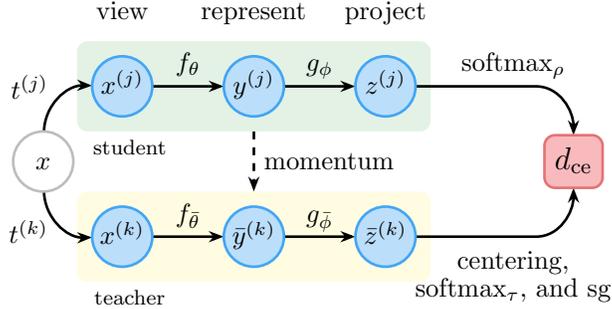

DINO defines a student and a teacher network. The student consists of an encoder
$f_\theta$ and a projector $g_\phi$ with parameters $\theta$ and $\phi$. The
encoder is implemented as a vision transformer \citep[ViT,][]{dosovitskiy2021an}
and the projector as an MLP. The teacher consists of an encoder
$f_{\bar{\theta}}$ and a projector $g_{\bar{\phi}}$ with the same architecture
as the student, but a separate set of parameters $\bar{\theta}$ and
$\bar{\phi}$.

\begin{figure}
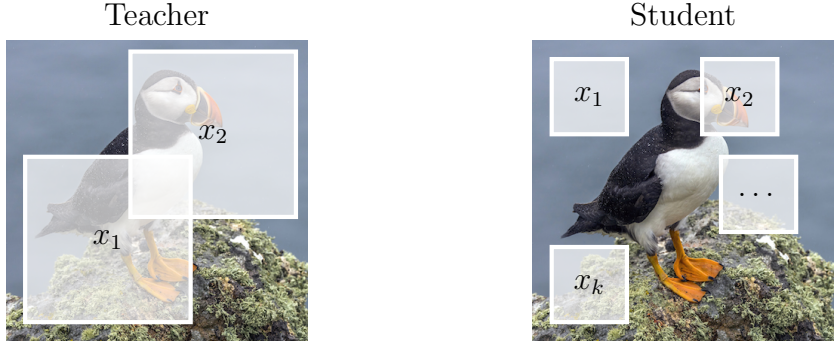

	\centering
	\begin{tikzpicture}
		\node at (0,2.35)  {Teacher};
		\node[inner sep=1mm] (x1) at (0,0){\includegraphics[width=4cm]{figures/jigsaw/puffin.jpg}};
		\draw[white,ultra thick,fill opacity=0.7, fill=white] (-1.75,-1.75) rectangle ++(2.2,2.2);
		\draw[white,ultra thick,fill opacity=0.7, fill=white] (-0.35,-0.35) rectangle ++(2.2,2.2);
		\node at (-0.65,-0.65)  {$x_{1}$};
		\node at (0.75,0.75)  {$x_{2}$};
		\node at (7,2.35)  {Student};
		\node[inner sep=1mm] (x2) at (7,0) {\includegraphics[width=4cm]{figures/jigsaw/puffin.jpg}};
		
		\draw[white,ultra thick,fill opacity=0.7, fill=white] ($(x2) +(-1.75, 0.75)$)  rectangle ++(1,1);
		\node at ($(x2) +(-1.75, 0.75)+(0.5,0.5)$)  {$x_1$};
		\draw[white,ultra thick,fill opacity=0.7, fill=white] ($(x2) +(0.25, 0.75)$)  rectangle ++(1,1);
		\node at ($(x2) +(0.25, 0.75)+(0.5,0.5)$)  {$x_2$};
		\draw[white,ultra thick,fill opacity=0.7, fill=white] ($(x2) +(-1.75, -1.75)$)  rectangle ++(1,1);
		\node at ($(x2) +(-1.75, -1.75)+(0.5,0.5)$)  {$x_k$};
		\draw[white,ultra thick,fill opacity=0.7, fill=white] ($(x2) +(0.5, -0.55)$) rectangle ++(1,1);
		\node at ($(x2) +(0.5, -0.55)+(0.5,0.5)$)  {$\dots$};
	\end{tikzpicture}
	\caption{Two sets of augmented crops
		for the teacher and student as defined by the
		multi-crop augmentation strategy. The teacher set contains two global views which
		cover at least 50\% of the augmented image. The student set is made up of $k$ patches that cover less than 50\% of the other augmented version of the input image.}
	\label{fig:dino_crops-tikz}
\end{figure}

DINO uses a multi-crop strategy first proposed by \cite{swav} to create a batch
of $m$ views $X_i = {[x_i^{(1)}, \ldots, x_i^{(m)}]}$ of an image $x_i$. Each
view is a random crop of $x_i$ followed by more transformations. Most crops
cover a small region of the image, but some crops are of high resolution, which
we refer to as \emph{local} and \emph{global} views, respectively. Let
$M_i$ be the set of indices of the global views. The idea is that
the student has access to all views, while the teacher only has access to the
global views, which creates ``local-to-global'' correspondences
\citep{caron2021emerging}. See Figure~\ref{fig:dino_crops-tikz} for an
illustration.

The student computes representations ${y_i^{(j)} = f_\theta(x_i^{(j)})}$ and
projections ${z_i^{(j)} = g_\phi(y_i^{(j)})}$ for each view. The teacher
computes target projections ${\bar{z}_i^{(j)} =
g_{\bar{\phi}}(f_{\bar{\theta}}(x_i^{(j)}))}$ for the global views ${j \in M_i}$.

\paragraph{Preventing collapse.}
\cite{caron2021emerging} experimentally discover two forms of collapse: Either
the computed probability distribution is uniform, or one dimension dominates,
regardless of the input. This motivates two countermeasures: 
\begin{enumerate}
	\item  To prevent collapse to a uniform distribution, the target distribution of the teacher is
	\emph{sharpened} by setting the temperature parameter~$\tau$ to a small
	value.  
	\item To prevent one dimension from dominating, the output of the teacher
	is centered to make it more uniform. This is accomplished by adding a centering
	vector $c$ as a bias to the teacher, which is computed with an exponentially
	moving average
	\begin{equation}
		c \leftarrow \beta c + (1 - \beta) \bar{z},
	\end{equation}
	where $\beta \in [0, 1]$ is a decay hyperparameter determining to what
	extent the centering vector is updated and 
	\begin{equation}
		\bar{z} = \frac{1}{n} \sum_{i=1}^n \frac{1}{|M_i|} \sum_{j\,\in\,M_i} \bar{z}_i^{(j)}
	\end{equation}
is the mean of all projections of the teacher in the current batch.

\end{enumerate}

\paragraph{Learning invariant features via soft labels.}
\label{sec:mctrain_dino}
DINO formulates the task of predicting the target projections as a knowledge
distillation task. The projections of the teacher and the student are converted
to probability distributions by applying the softmax function over all
components. Hereby, the cross-entropy loss can be applied, where the teacher
computes soft labels for the student. The total loss function matches every view
of the student to every global view of the teacher (except the same global
views), i.e.,
\begin{equation}
  \mathcal{L}^{\text{DINO}}_{\theta, \phi} = \frac{1}{n} \sum_{i=1}^n \sum_{j \,\in\, M_i} \sum_{k \neq j} \dce(\softmax_{\tau}(\bar{z_i}^{(j)} - c), \softmax_{\rho}(z_i^{(k)})),
  \label{eq:cross_ent_dino2}
\end{equation}
where $\tau, \rho > 0$ are hyperparameters controlling the temperature of the
distributions (see Section~\ref{sec:notation}) for the teacher and the student,
respectively. Overall, the parameter updates are very similar to BYOL's, since
the student network is updated by minimizing the loss
$\mathcal{L}^{\text{DINO}}_{\theta, \phi}$ using the AdamW optimizer, and the
teacher network is updated by an exponential moving average of the student,
i.e.,
\begin{align}
  \bar{\theta} & \leftarrow \alpha \bar{\theta}  + (1-\alpha) \theta, \\
  \bar{\phi} & \leftarrow \alpha \bar{\phi}  + (1-\alpha) \phi,
  \label{eq:update_dino}
\end{align}
where $\alpha \in [0,1]$ controls the rate at which the weights of the teacher
network are updated with the weights of the student network.

The authors identify interesting properties when training ViTs in a
self-supervised manner. ViTs trained via self-supervision are able to detect
object boundaries within a scene layout which is information that can be
extracted within the attention layers. Furthermore, the learned attention maps
learn segmentation masks, i.e., the objects are separated from the background in
the attention masks. These attention masks allow DINO to perform well on downstream
tasks simply by using a k-nearest-neighbor classifier on its representations.

%% file: figures/dino.tex
\begin{tikzpicture}
  \node[whitevar] (x) at (0,0) {$x$};

  \node[bluevar] (xt) at ($(x) + (1.05,1)$) {$x^{(j)}$};
  \node[bluevar] (y) at ($(xt) + (1.75,0)$) {$y^{(j)}$};
  \node[bluevar] (z) at ($(y) + (1.75,0)$) {$z^{(j)}$};
  \coordinate (p) at ($(z) + (2,0)$);

  \node[bluevar] (xtp) at ($(xt) - (0,2)$) {$x^{(k)}$};
  \node[bluevar] (yp) at ($(y) - (0,2)$) {$\bar{y}^{(k)}$};
  \node[bluevar] (zp) at ($(z) - (0,2)$) {$\bar{z}^{(k)}$};
  \coordinate (pp) at ($(p) - (0,2)$);

  \node[nicelabel] (view) at ($(xt) + (0,1)$) {view};
  \node[nicelabel] (repr) at ($(y) + (0,1)$) {represent};
  \node[nicelabel] (proj) at ($(z) + (0,1)$) {project};

  \node[nicelabel] (student) at ($(xt) + (0.1,-0.8)$) {\scriptsize student};
  \node[nicelabel] (teacher) at ($(xtp) + (0.1,-0.8)$) {\scriptsize teacher};
  
  \node[box, draw=nicedarkred, fill=nicered] (loss) at ($(p) + (0.5,-1)$) {$\dce$};

  \begin{pgfonlayer}{back}
    \path[box,fill=nicegreen,opacity=0.5] ($(xt) - (0.6,0.6)$) rectangle ($(z) + (0.6,0.6)$);
    \path[box,fill=niceyellow,opacity=0.5] ($(xtp) - (0.6,0.6)$) rectangle ($(zp) + (0.6,0.6)$);

    \begin{scope}[transform canvas]  
      \draw[nicearrow] (x) edge[->,bend left=45] node[midway,left,nicelabel,yshift=1mm] {$t^{(j)}$} (xt);
      \draw[->,nicearrow] (xt) -- node[midway,above,nicelabel] {$f_\theta$} (y);
      \draw[->,nicearrow] (y) -- node[midway,above,nicelabel] {$g_\phi$} (z);
      \draw[nicearrow] (z) -- node[midway,above,nicelabel,xshift=5mm] {$\softmax_{\rho}$} (p);

      \draw[nicearrow] (x) edge[->,bend right=45] node[midway,left,nicelabel,yshift=-1.5mm] {$t^{(k)}$} (xtp);
      \draw[->,nicearrow] (xtp) -- node[midway,above,nicelabel] {$f_{\bar{\theta}}$} (yp);
      \draw[->,nicearrow] (yp) -- node[midway,above,nicelabel] {$g_{\bar{\phi}}$} (zp);
      \draw[nicearrow] (zp) -- node[midway,below,nicelabel,align=center,xshift=5mm,yshift=-4mm] {centering,\\[-0.5mm] {$\softmax_{\tau}$}, and $\sg$} (pp);
     
       \draw[nicearrow] (p) edge[->,out=0,in=90,relative=false] (loss);
        \draw[nicearrow] (pp) edge[->,out=0,in=-90,relative=false] (loss);

      \draw[->,nicearrow,dashed] ($(y) - (0,0.6)$) -- node[midway,right,nicelabel] {momentum} ($(yp) + (0,0.6)$);
    \end{scope}
  \end{pgfonlayer}
\end{tikzpicture}

%% file: sections/esvit.tex
\subsection{Efficient Self-Supervised Vision Transformers (EsVit)}
\cite{li2021efficient} keep the same teacher-student architecture as
\cite{caron2021emerging}, but replace the ViTs in \cite{caron2021emerging} with
multi-stage transformers. As an optimization to the transformer architecture, a
multi-stage transformer subsequently merges image patches together across every
layer to reduce the number of image patches that have to be processed. The
authors show that the merging process destroys important local to global
correspondences which are learnt in common transformers. Therefore, an
additional region matching loss is proposed that mitigates the lost semantic
correspondences during the merging process in multi-stage transformer
architectures.

\begin{figure}[ht!]
  \centering \input{figures/esvit.tex}
  \caption{Multi-stage transformer: An image is decomposed into image patches,
    which are merged and applied self-attention to subsequently. This results in
    a smaller number of patches being processed simultaneously, while also
    learning hierarchical embeddings. At the end, either average pooling is
    performed over the outputs or the output sequence is used as it is.}
  \label{fig:esvit}
\end{figure}
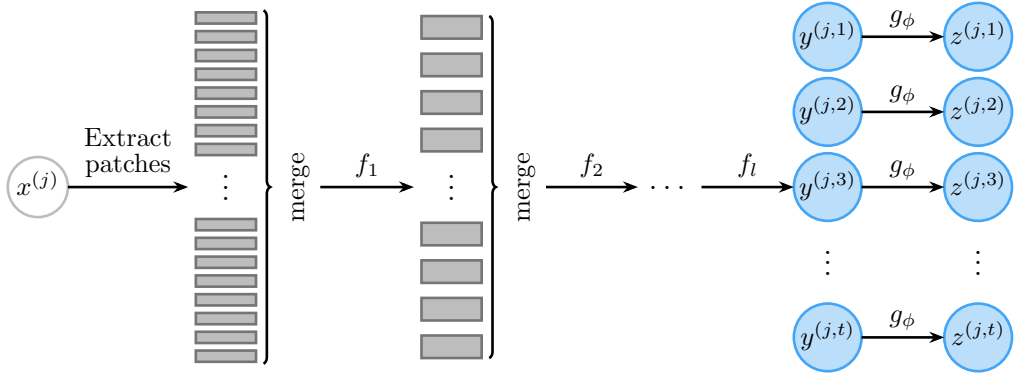

\paragraph{Multi-stage vision transformers.}
\cite{vaswani2021multi} reduce computational complexity of a standard transformer by reducing the number
of patches that go through the transformer at each layer. For this, a special
image patch merging module merges the patches at each layer and attention is
calculated between them via sparse self-attention modules. This process is
repeated multiple times. An illustration of the procedure is provided in Figure
\ref{fig:esvit}. Overall, the number of tokens, i.e., feature maps that have
to be evaluated by one self-attention module is reduced through each subsequent
layer, while allowing for more diverse feature learning due to the different
self-attention heads processing different merged patches, allowing for
hierarchical correspondences to be learned.

\paragraph{Extending the loss function for region level feature matching.}

\cite{li2021efficient} propose to extend DINO's loss for multi-stage
transformers in order to learn the local-to-global correspondences lost during
the merging process. As EsVit is an extension of DINO, the teacher network and
student network are defined as in Section \ref{sec:mctrain_dino}. \cite{li2021efficient} propose a loss function that consists of a view-level
loss $\ell_{\text{view}}$ and a region-level loss $\ell_{\text{region}}$. The
view-level loss is the same loss used to train DINO, i.e.,
\begin{equation}
  \ell_{\text{view}} = \frac{1}{n} \sum_{i=1}^n \sum_{j \,\in\, M_i} \sum_{k \neq j} \dce(\softmax_{\tau}(\bar{z_i}^{(j)} - c_{\text{view}}), \softmax_{\rho}(z_i^{(k)})).
  \label{eq:cross_ent_esvit1}
\end{equation}
where $c_\text{view}$ is the centering vector for the view-level loss.

The region-level loss of EsVit is computed from the encoder outputs for each image patch
$Y_i = [y_i^{(1, 1)}, \ldots, y_i^{(m, T)}]$ directly (see Figure \ref{fig:esvit}), where $T$ is the sequence length, i.e., the number of patches for a given view $j$. Then the region-level loss is defined as
\begin{equation}
  \ell_{\text{region}} = \frac{1}{n} \sum_{i=1}^n \sum_{j \,\in\, M_i} \sum_{k \neq j} \sum_{t=1}^{T} \dce(\softmax_{\tau}(\bar{z}_i^{(j, s^*)} - c_{\text{region}}), \softmax_{\rho}(z_i^{(k, t)})).
  \label{eq:cross_ent_esvit2}
\end{equation}
where $s^* =\argmax_s \scos(\bar{z}^{(j, s)}, z^{(k, t)})$, $T$ is the number of image patches, and $c_{\text{region}}$ is the centering vector for the region-level loss.
The idea is to match every image patch projection of the student $z^{(k, t)}$ to the
best image patch projection of the teacher $\bar{z}^{(j, s^*)}$. That is, for every
projection as defined by the multi-crop strategy in
Figure \ref{fig:dino_crops-tikz}, the region-level loss matches the most
concurring image patches of the student and teacher. The final EsVit loss combines the
view-level and region-level loss, i.e.,
\begin{align}
  \mathcal{L}^{\text{EsVit}}_{\theta, \phi} =  \ell_{\text{view}} +  \ell_{\text{region}}
  \label{eq:esvitce}
\end{align}

The authors show that when only training with the view-level loss on a
multi-stage transformer, the model fails to capture meaningful correspondences,
such as the background being matched for two augmentations of the same image.
Adding the region-level loss mitigates the issue of lost region-level
correspondences in multi-stage transformer architectures and recovers some of
the correspondences learnt inherently by monolithic transformer architectures.

%% file: figures/esvit.tex
\begin{tikzpicture}
  \node[whitevar] (x) at (0.5,0) {$x^{(j)}$};
  \foreach \i in {0.5,0.75,...,2.25}
    {
      \node[smallbox] (a\i) at ($(3,0)+(0,\i)$) {};
      \node[smallbox] (b\i) at ($(3,0)-(0,\i)$) {};
    }
  \node[text depth=2mm] (p1) at (3,0)  {$\vdots$};

  \foreach \i in {0.625,1.125,...,2.125}
    {
      \node[bigbox] (c\i) at ($(6,0)+(0,\i)$) {};
      \node[bigbox] (c\i) at ($(6,0)-(0,\i)$) {};
    }
  \draw [decorate, decoration = {brace}, line width=0.35mm] (3.5,2.35) --  (3.5,-2.35);
  \draw [decorate, decoration = {brace}, line width=0.35mm] (6.5,2.275) --  (6.5,-2.275);
  \node[text depth=2mm] at (6,0)  {$\vdots$};

  \node[rotate=90] (m1) at (4,0)  {\footnotesize merge};
  \node[rotate=90] (m2) at (7,0)  {\footnotesize merge};

  \node[bluevar, minimum size=0.9cm] (y1) at(11,2) {$y^{(j,1)}$};
  \node[bluevar, minimum size=0.9cm] (y2) at(11,1) {$y^{(j,2)}$};
  \node[text depth=2mm] at (11,-1)  {$\vdots$};
  \node[bluevar, minimum size=0.9cm] (y3) at(11,0) {$y^{(j,3)}$};
  \node[bluevar, minimum size=0.9cm] (yk) at(11,-2) {$y^{(j,t)}$};

  \node[bluevar, minimum size=0.9cm] (z1) at (13,2) {$z^{(j,1)}$};
  \node[bluevar, minimum size=0.9cm] (z2) at (13,1) {$z^{(j,2)}$};
  \node[text depth=2mm] at (13,-1)  {$\vdots$};
  \node[bluevar, minimum size=0.9cm] (z3) at (13,0) {$z^{(j,3)}$};
  \node[bluevar, minimum size=0.9cm] (zk) at (13,-2) {$z^{(j,t)}$};

  \draw[->,nicearrow] (y1) -- node[midway,above, black,nicelabel] {$g_\phi$} (z1);
  \draw[->,nicearrow] (y2) -- node[midway,above, black,nicelabel] {$g_\phi$} (z2);
  \draw[->,nicearrow] (y3) -- node[midway,above, black,nicelabel] {$g_\phi$} (z3);
  \draw[->,nicearrow] (yk) -- node[midway,above, black,nicelabel] {$g_\phi$} (zk);

  \draw[->,nicearrow] (x) -- node[above,nicelabel,align=center] {\footnotesize Extract \\[-2pt] \footnotesize patches} (2.5,0);
  \draw[->,nicearrow] (m1) -- node[midway,above,nicelabel] {$f_1$} (5.5,0);
  \node(dots) at (8.9,0)  {$\dots$};
  \draw[->,nicearrow] (m2) -- node[midway,above,nicelabel] {$f_2$} (dots);
  \draw[->,nicearrow] (dots) -- node[midway,above,nicelabel] {$f_l$} (y3);
\end{tikzpicture}

%% file: sections/simsiam.tex
\subsection{Simple Siamese Representation Learning (SimSiam)}
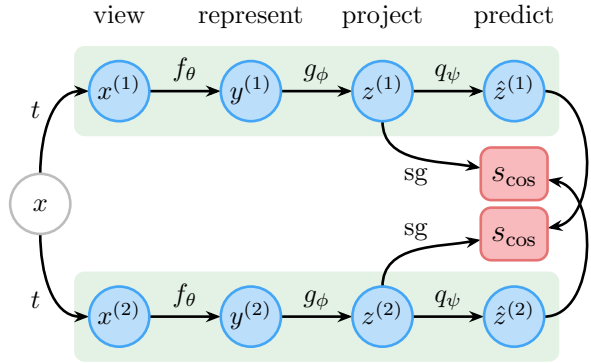
\begin{figure}[ht!]
	\centering \input{figures/simsiam.tex}
	\caption{SimSiam maximizes the cosine similarity between the projected
		representations. Both networks use the same parameters. The stop gradient
		operator interrupts the backward-flow of the gradients and thereby preventing the representational collapse.}
	\label{fig:simsiam_arch}
\end{figure}

\noindent SimSiam was introduced by \citet{chen2021exploring} and uses a similar architecture and loss
function as BYOL. However, teacher and student share the same parameters and
hence a momentum encoder is not used as in previously presented teacher-student
methods.

Given a batch of images $X$, for each image $x_i$ two views
${x_i^{(1)} = t(x_i)}$, ${x_i^{(2)} = t(x_i)}$ are created using
random transformations $t \sim \mathcal{T}$ that are sampled for each image and each view. For each of these views, a Siamese encoder $f_\theta$ computes a representation ${y_i^{(j)} =
f(x_i^{(j)})}$ and a Siamese projector $g_\phi$ computes a projection
${z_i^{(j)} = g_\phi(y_i^{(j)})}$. Finally, the projection is fed through a
predictor $q_\psi$ to obtain a prediction ${\hat{z}_i^{(j)} =
q_\psi(z_i^{(j)})}$.

The goal of the predictor is to predict the projection of the other view.
Therefore, the loss computes the negative cosine similarity between the
prediction of the first view and the projection of the second view, and vice
versa, i.e.,
\begin{equation}
  \mathcal{L}^\text{SimSiam}_{\theta,\phi,\psi} = -\frac{1}{n} \sum_{i=1}^n \frac{1}{2} \left[ \scos(\hat{z}_i^{(1)}, \sg(z_i^{(2)})) + \scos(\hat{z}_i^{(2)}, \sg(z_i^{(1)})) \right]\!\!,
\end{equation}
where $\sg(\cdot)$ is the stop gradient operator that prevents gradients
from being backpropagated through this branch of the computational graph.

The encoder $f_\theta$ is implemented as a ResNet~\citep{he2016deep}. The projector $g_\phi$ and the predictor $q_\psi$ are MLPs.
The authors show empirically that a predictor is crucial to
avoid collapse. \cite{chen2021exploring} argue that the gradient of the
symmetrized loss with a predictor that is the identity is in the same direction
as the gradient of the symmetrized loss between the two projections, such that
the stop-gradient operation is cancelled out, thus leading to representation
collapse. Using a random predictor does not work either and
\cite{chen2021exploring} argue that the predictor should always learn the latest
representations. This argument is similar to \cite{grill2020bootstrap} in
Section \ref{sec:byol}, where the predictor should be kept near optimal at all
times.

Another crucial ingredient to their method is batch
normalization~\citep{ioffe2015batch}, which is used for both $f_\theta$ and
$g_\phi$. Furthermore, the authors experiment with the training objective by
replacing it with the cross-entropy loss. Their experiments show that this also
works, however the performance is worse. The key advantage of SimSiam is that
training does not require large batch sizes allowing the use of SGD instead of
LARS.

%% file: figures/simsiam.tex
\begin{tikzpicture}
  \node[whitevar] (x) at (0,0) {$x$};

  \node[bluevar] (xt) at ($(x) + (1.05,1.5)$) {$x^{(1)}$};
  \node[bluevar] (y) at ($(xt) + (1.75,0)$) {$y^{(1)}$};
  \node[bluevar] (z) at ($(y) + (1.75,0)$) {$z^{(1)}$};
  \node[bluevar] (p) at ($(z) + (1.75,0)$) {$\hat{z}^{(1)}$};

  \node[bluevar] (xtp) at ($(xt) - (0,3)$) {$x^{(2)}$};
  \node[bluevar] (yp) at ($(y) - (0,3)$) {$y^{(2)}$};
  \node[bluevar] (zp) at ($(z) - (0,3)$) {$z^{(2)}$};
  \node[bluevar] (pp) at ($(p) - (0,3)$) {$\hat{z}^{(2)}$};

  \node[box, draw=nicedarkred, fill=nicered] (loss1) at ($(p) + (0,-1.9)$) {$\scos$};
  \node[box, draw=nicedarkred, fill=nicered] (loss2) at ($(pp) + (0,1.9)$) {$\scos$};

  \node[nicelabel] (view) at ($(xt) + (0,1)$) {view};
  \node[nicelabel] (repr) at ($(y) + (0,1)$) {represent};
  \node[nicelabel] (proj) at ($(z) + (0,1)$) {project};
  \node[nicelabel] (pred) at ($(p) + (0,1)$) {predict};

  \begin{pgfonlayer}{back}
      \path[box,fill=nicegreen,opacity=0.5] ($(xt) - (0.6,0.6)$) rectangle ($(p) + (0.6,0.6)$);
      \path[box,fill=nicegreen,opacity=0.5] ($(xtp) - (0.6,0.6)$) rectangle ($(pp) + (0.6,0.6)$);

      \begin{scope}[transform canvas]  
        \draw[nicearrow] (x) edge[->,out=90,in=180,relative=false] node[midway,left,nicelabel,yshift=1mm] {$t$} (xt);
        \draw[->,nicearrow] (xt) -- node[midway,above,nicelabel] {$f_\theta$} (y);
        \draw[->,nicearrow] (y) -- node[midway,above,nicelabel] {$g_\phi$} (z);
        \draw[->,nicearrow] (z) -- node[midway,above,nicelabel] {$q_\psi$} (p);

        \draw[nicearrow] (x) edge[->,out=-90,in=180,relative=false] node[midway,left,nicelabel,yshift=-1.5mm] {$t$} (xtp);
        \draw[->,nicearrow] (xtp) -- node[midway,above,nicelabel] {$f_\theta$} (yp);
        \draw[->,nicearrow] (yp) -- node[midway,above,nicelabel] {$g_\phi$} (zp);
        \draw[->,nicearrow] (zp) -- node[midway,above,nicelabel] {$q_\psi$} (pp);

        \draw[nicearrow] (z) edge[->,out=-90,in=170,relative=false] node[midway,below,nicelabel,yshift=1mm] {sg} (loss2);
        \draw[nicearrow] (zp) edge[->,out=90,in=-170,relative=false] node[midway,above,nicelabel] {sg} (loss1);

        \draw[nicearrow] (p) edge[->,out=0,in=10,relative=false] (loss1);
        \draw[nicearrow] (pp) edge[->,out=0,in=-10,relative=false] (loss2);
      \end{scope}
  \end{pgfonlayer}
\end{tikzpicture}

%% file: sections/contrastive.tex
\section{Contrastive Representation Learning}
\label{sec:contrastive}

Contrastive methods prevent representation collapse by \emph{decreasing} the
similarity between representations of unrelated data points. Given a data point
$x^*$ called the \emph{anchor}, one defines mechanisms to generate
\emph{positive} samples and \emph{negative} samples for the anchor. The
positives should retain the relevant information of the anchor, while the
negatives should contain information different from the anchor. For vision
tasks, the positives could, e.g., be random transformations of the same image,
while the negatives are (transformations of) other images. The goal of
contrastive methods is to move representations of positives closer to the
representation of the anchor while moving representations of negatives away from
the anchor.

More formally, given an anchor $x^*$ we define the conditional distributions of
positives ${p_\text{pos}(x^\tplus | x^*)}$ and negatives
${p_\text{neg}(x^\tminus | x^*)}$. These distributions are induced by the
mechanisms that generate the positives and negatives and are not explicitly
known. Let ${y^* = f_\theta(x^*)}$, ${y^\tplus = f_\theta(x^\tplus)}$, and
${y^\tminus = f_\theta(x^\tminus)}$ be the corresponding representations,
calculated with an encoder $f_\theta$ parameterized by $\theta$. The task of
contrastive representation learning methods is to maximize the likelihood of
positive representations $p_\text{pos}(y^\tplus | y^*)$ while minimizing the
likelihood of negative representations $p_\text{neg}(y^\tminus | y^*)$. Note
that we continue to use representations $y$ in this introduction, but the same
methods can be applied to projections $z$ equivalently.

\paragraph{Noise Contrastive Estimation (NCE).}\label{sec:nce}
The idea of noise contrastive estimation \citep{gutmann2010noise} is to
formulate the task of contrastive representation learning as a supervised
classification problem. One assumption of NCE is that the negatives are
independent from the anchor, i.e., ${p_\text{neg}(x^- | x^*)} =
{p_\text{neg}(x^-)}$. In this context, the negatives are often called
\emph{noise}. There are two widely used approaches, the original NCE and InfoNCE
\citep{oord2018representation}. Roughly speaking, NCE performs binary
classification to decide whether an individual sample is a positive or negative,
whereas InfoNCE performs multiclass classification on a set of samples to decide
which one is the positive.
In the following we will explain InfoNCE in more detail.

\paragraph{InfoNCE.}
For each anchor $x^*$, InfoNCE generates one positive sample from
$p_\text{pos}(x^\tplus | x^*)$ and ${n - 1}$ negative samples from
$p_\text{neg}(x^\tminus)$. Let ${X = [x_1, \ldots, x_n]}$ be the set of those
samples, where $x_c$ is the positive with index ${c \in \{1, \ldots, n\}}$. In
the context of representation learning we further compute representations using
an encoder $f_\theta$ and obtain the set ${Y = [y_1, \ldots, y_n]}$.

InfoNCE now defines a supervised classification task, where the input is ${(y^*,
Y)}$ and the class label is the index of the positive $c$. A classifier
$p_\psi(c | Y, y^*)$ with parameters $\psi$ is trained to match the true data
distribution of the labels $p_\text{data}(c | Y, y^*)$. A common supervised
learning objective is to minimize the cross-entropy between the data
distribution and the model distribution, i.e.,
\begin{align}
  \label{eq:info_nce_ce}
       & \min_{\psi,\theta} \mathbb{E}_{Y, y^*} \left[ H(p_\text{data}(c | Y, y^*), p_\psi(c | Y, y^*)) \right]                   \\
  \label{eq:info_nce_log}
  = \  & \min_{\psi,\theta} \mathbb{E}_{Y, y^*} \left[ \mathbb{E}_{c | Y, y^*} \left[ -\log p_\psi(c | Y, y^*) \right] \right]\!.
\end{align}
Note that this is an anticausal prediction problem, where the underlying cause
(label) is predicted from its effect (input) \citep{scholkopf2012causal}. In
InfoNCE we know the underlying mechanism (since we generate the labels
artificially), so we can derive the optimal classifier using Bayes' theorem.

First, we write down the data distribution of a set $Y$ given a label and an
anchor, i.e.,
\begin{align}
  p_\text{data}(Y | c, y^*)
   & = \prod_{i=1}^n p_\text{data}(y_i | c, y^*)
  = \prod_{i=1}^n \begin{cases}
                    p_\text{pos}(y_i | y^*), & \text{if } i = c,    \\
                    p_\text{neg}(y_i),     & \text{if } i \neq c,
                  \end{cases}   \\
   & = p_\text{pos}(y_c | y^*) \prod_{i \neq c} p_\text{neg}(y_i)
  = \frac{p_\text{pos}(y_c | y^*)}{p_\text{neg}(y_c)} \prod_{i=1}^n p_\text{neg}(y_i),
\end{align}
where we assume conditional independence among the samples in $Y$. InfoNCE
further assumes that the labels are sampled uniformly, i.e., ${p_\text{data}(c)
= \frac{1}{n}}$. Now we apply Bayes' theorem:
\begin{align}
  p_\text{data}(c | Y, y^*)
   & = \frac{p_\text{data}(Y | c, y^*) \: p_\text{data}(c)}{\sum_{c'=1}^n p_\text{data}(Y | c', y^*) \: p_\text{data}(c')}   \\
   & = \dfrac{\frac{p_\text{pos}(y_c | y^*)}{p_\text{neg}(y_c)} \prod_{i=1}^n p_\text{neg}(y_i) \: \frac{1}{n}}
  {\sum_{c'=1}^n \frac{p_\text{pos}(y_{c'} | y^*)}{p_\text{neg}(y_{c'})} \prod_{i=1}^n p_\text{neg}(y_i) \: \frac{1}{n}} \\
   & = \dfrac{\frac{p_\text{pos}(y_c | y^*)}{p_\text{neg}(y_c)}}
  {\sum_{c'=1}^n \frac{p_\text{pos}(y_{c'} | y^*)}{p_\text{neg}(y_{c'})}}.
  \label{eq:info_nce_data}
\end{align}
An optimal classifier with zero cross-entropy would match this distribution. We
can see that the optimal probability of a class is the density ratio
$\frac{p_\text{pos}(y_c | y^*)}{p_\text{neg}(y_c)}$, normalized across all
classes. It describes the likelihood of $y_c$ being a positive sample for $y^*$
versus being a negative sample. This motivates the choice of the classifier of
InfoNCE, which is defined similar to Equation~\ref{eq:info_nce_data}:
\begin{equation}
  \label{eq:info_nce_model}
  p_\psi(c | Y, y^*) = \dfrac{s_\psi(y^*, y_c)}{\sum_{c'=1}^n s_\psi(y^*, y_{c'})},
\end{equation}
where $s_\psi(y^*, y)$ is a predictor that computes a real-valued positive
score. Minimizing the cross-entropy from Equation~\ref{eq:info_nce_ce} brings
the model distribution $p_\psi(c | Y, y^*)$ closer to the data distribution
$p_\text{data}(c | Y, y^*)$, which ensures that $s_\psi$ approaches the density
ratio of the data, i.e., $s_\psi(y^*, y) \approx \frac{p_\text{pos}(y |
y^*)}{p_\text{neg}(y)}$ (in fact, it only needs to be proportional to the
density ratio). The density ratio is high for positive samples and close to zero
for negative samples, which means that $s_\psi(y^*, y)$ learns some similarity
measure between the representations. Since $\psi$ and $\theta$ (i.e., predictor
and encoder) are optimized jointly, the encoder is encouraged to learn similar
embeddings for an anchor and its positive, and to learn dissimilar embeddings
for an anchor and its negative samples (as long as the predictor is not too
expressive). In other words, the encoder is encouraged to extract information
that is ``unique'' to the anchor and the positive sample. Moreover,
\citet{oord2018representation} show that this objective maximizes the mutual
information between $y^*$ and $y^\tplus$, which is a lower bound on the mutual
information between $x^*$ and $x^\tplus$.

By combining the negative logarithm and the classifier
(Equations~\ref{eq:info_nce_log} and \ref{eq:info_nce_model}) the general
InfoNCE loss for $(y^*, Y, c)$ is defined as
\begin{equation}
  \infonce_{s_\psi}(y^*, Y, c) = -\log\!\left( \frac{s_\psi(y^*, y_c)}{\sum_{c'=1}^n s_\psi(y^*, y_{c'})} \right)\!.
\end{equation}
For the sake of notation in the following sections, we slightly adjust this
definition. All considered methods compute the exponential of some score to
obtain positive values for $s_\psi$, which is why we include it directly in the
loss function. In this case the InfoNCE loss computes the commonly used softmax
cross-entropy. Instead of specifying the class label we denote the positive by
$y^\tplus$ and the set of negatives by $\bar{Y}$. Thus, our final definition
of the InfoNCE loss for a score function $s_\psi(y^*, y)$ is
\begin{equation}
  \label{eq:info_nce}
  \infonce_{s_\psi}(y^*, y^\tplus, \bar{Y}) = -\log\!\left( \frac{\exp(s_\psi(y^*, y^\tplus))}{\exp(s_\psi(y^*, y^+)) + \sum_{\bar{y} \,\in\, \bar{Y}} \exp(s_\psi(y^*, \bar{y}))} \right)\!.
\end{equation}

\input{sections/cpc.tex}

\input{sections/cmc.tex}
\input{sections/simclr.tex}
\input{sections/moco.tex}
\input{sections/pirl.tex}

%% file: sections/cpc.tex
\subsection{Contrastive Predictive Coding (CPC)}

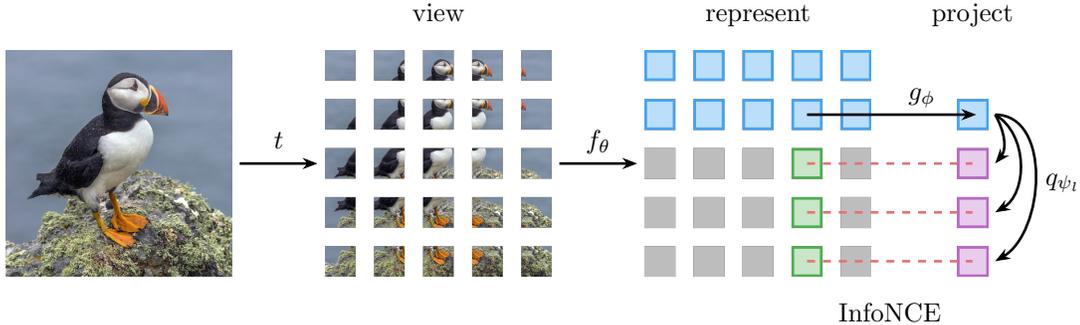
\begin{figure}[ht]
	\centering
	\input{figures/cpc.tex}
	\caption{In CPC training an image is transformed and split into overlapping patch encodings, where the top rows (blue) act as the anchor. The positive samples (purple) lie in a column below the exact anchor position and every other patch of the whole dataset is a negative. InfoNCE is applied to distinguish between both distributions.}
	\label{fig:cpc}
\end{figure}

CPC~\citep{oord2018representation} is an influential self-supervised
representation learning technique which is applicable to a wide variety of input
modalities such as text, speech and images. It is based on the theory of
predictive coding, which originated in the neuroscience literature by observing
the learning behavior of biological neural circuits
\citep{huang2011predictive,bastos2012canonical}. In short, a model tries to
predict future outcomes given the past or \emph{context}. Thus, the learned
representation of the context should incorporate all information necessary for
prediction while removing unimportant noise. This predictive coding task is
solved by formulating it as a contrastive learning problem. CPC operates on
sequential data, which is a natural choice for audio data, but can also be
applied to vision tasks by splitting images into sequences of patches. 

Given a batch of images $X$, we consider a single image $x_i$. The image is
split into $m$ patches ${[ x_i^{(1)}, \ldots, x_i^{(m)} ]}$. Note that the
patches are overlapping and that an additional image augmentation is applied to
each patch. A Siamese encoder $f_\theta$ converts each patch into a
representation ${y_i^{(j)} = f_\theta(x_i^{(j)})}$. The \emph{context} of a
patch are the patches in the same row and all patches in the rows above. Let
${C_i^{(j)} \subset \{1, \ldots, m\}}$ be the set of indices of those patches. A
projector $g_\phi$ computes a context representation ${z_i^{(j)} = g_\phi([
y_i^{(k)} : k \in C_i^{(j)}])}$ for each patch. The network $g_\phi$ is
implemented by a masked CNN \citep[PixelCNN,][]{van2016conditional} that
restricts its receptive field to the patches in the context.

Given a context representation $z_i^{(j)}$, the predictive coding task of CPC is
to predict representations of future patches, which are the patches in the
column below the patch. Let ${F_i^{(j)} = [ k_{i,1}^{(j)}, \ldots, k_{i,K}^{(j)}
] \subset \{1, \ldots, m\}}$ be the set of indices of those patches. The
prediction task is solved by minimizing an InfoNCE loss for each future
representation. For the $l$-th future representation with patch index $k =
k_{i,l}^{(j)}$, a separate predictor $q_{\psi_l}$ computes the anchor
${\hat{y}_i^{(k)} = q_{\psi_l}(z_i^{(j)})}$ from the context representation, and
the positive is the future representation $y_i^{(k)}$. The negatives can be any
unrelated representations, e.g., the representations of all patches outside the
context and the future, as well as all representations from other images in $X$.
We denote the set of negatives by $\bar{Y}_i^{(j)}$. The loss function
accumulates the InfoNCE losses over all contexts and futures across the batch,
i.e.,
\begin{equation}
	\mathcal{L}^\text{CPC}_{\theta, \phi, \psi} = \frac{1}{n} \sum_{i=1}^n \frac{1}{m} \sum_{j=1}^m \sum_{k \,\in\, F_i^{(j)}} \infonce_{s}(\hat{y}_i^{(k)}, y_i^{(k)}, \bar{Y}_i^{(j)}),
\end{equation}
where the similarities are calculated using the dot product ${s(\hat{y}, y) =
\hat{y}^\top y}$, and the parameters of all predictors are combined in ${\psi =
[\psi_1, \ldots, \psi_K]}$. See Figure~\ref{fig:cpc} for an illustration of the
method.

\paragraph{CPC v2.}
The second version of CPC \citep{henaff2020data} targets the problem of sample
efficiency. So far, contrastive methods require a large amount of data,
especially the need to find hard negatives, to perform well on common benchmarks
such as ImageNet \citep{bardes2021vicreg}. This work focuses on the improvement
of training routines rather than modifying the general idea behind CPC.
The most notable changes are improved image augmentations, larger network sizes,
using layer normalization \citep{ba2016layer} instead of batch normalization
\citep{ioffe2015batch}, which creates an unintentional co-dependency between the
patches, and extending the prediction task to all four directions, instead of
predicting bottom from top patches only.

%% file: figures/cpc.tex
\begin{tikzpicture}
  \node[inner sep=1mm] (x) at (0,0) {\includegraphics[height=3cm]{figures/transformations/puffin.jpg}};
  \node[inner sep=1mm] (xt) at ($(x) + (4.25,0)$) {\includegraphics[height=3cm]{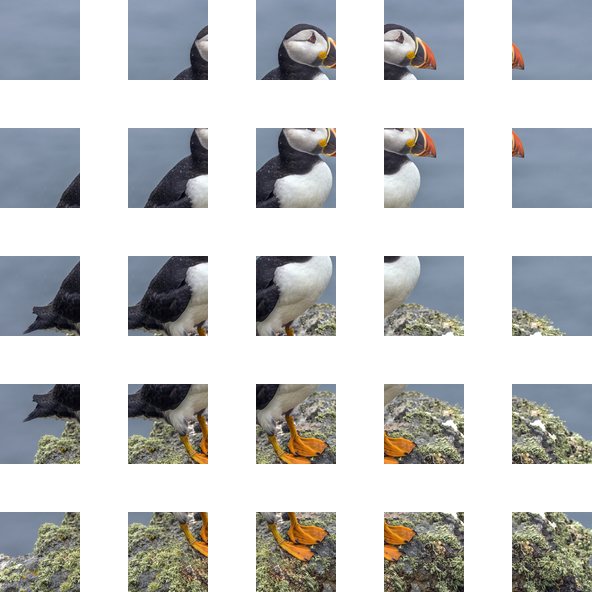}};
  \node[inner sep=1mm] (y) at ($(xt) + (4.25,0)$) {\includegraphics[height=3cm]{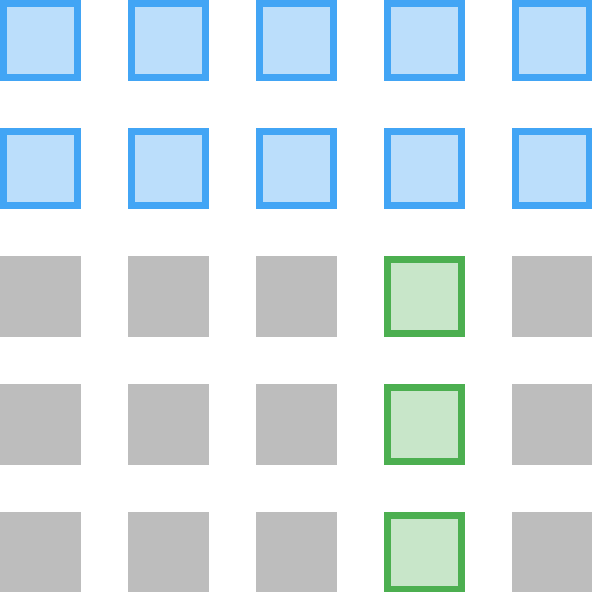}};
  \node[inner sep=1mm] (z) at ($(y) + (2.85,0)$) {\includegraphics[height=3cm]{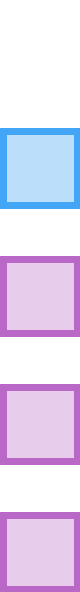}};

  \node[nicelabel] (view) at ($(xt) + (0,2)$) {\footnotesize view};
  \node[nicelabel] (repr) at ($(y) + (0,2)$) {\footnotesize represent};

  \begin{scope}[transform canvas]  
    \node[nicelabel] (proj) at ($(z) + (0,2)$) {\footnotesize project};

    \draw[->,nicearrow] (x.east) -- node[midway,above,nicelabel] {$t$} (xt.west);
    \draw[->,nicearrow] (xt.east) -- node[midway,above,nicelabel] {$f_\theta$} (y.west);
    \draw[->,nicearrow] ($(y) + (0.65,0.65)$) -- node[midway,above,nicelabel,xshift=4mm] {$g_\phi$} ($(z) + (0.05,0.65)$);

    \draw[nicearrow] ($(z.center) + (0.3,0.65)$) edge[->,out=-60,in=50,relative=false] ($(z.center) + (0.3,0)$);
    \draw[nicearrow] ($(z.center) + (0.3,0.65)$) edge[->,out=-30,in=30,relative=false] ($(z.center) + (0.3,-0.65)$);
    \draw[nicearrow] ($(z.center) + (0.3,0.65)$) edge[->,out=0,in=30,relative=false] node[midway,right,nicelabel] {$q_{\psi_l}$} ($(z.center) + (0.3,-1.3)$);
  \end{scope}

  \draw[nicearrow,dashed,nicedarkred] ($(y) + (0.65,0)$) -- (z.center);
  \draw[nicearrow,dashed,nicedarkred] ($(y) + (0.65,-0.65)$) -- ($(z.center) + (0,-0.65)$);
  \draw[nicearrow,dashed,nicedarkred] ($(y) + (0.65,-1.3)$) -- node[midway,below,nicelabel,black,yshift=-4mm] {InfoNCE} ($(z.center) + (0,-1.3)$);
\end{tikzpicture}

%% file: sections/cmc.tex
\subsection{Contrastive Multiview Coding (CMC)}

\begin{figure}
	\centering \input{figures/cmc.tex}
	\caption{CMC assumes there are different interfaces to the world, that collect information about a shared source. The goal is to find a consensus about the perceived information coming from the different modalities. This is done by extracting the view-invariant features in order to learn a shared representation.}
	\label{fig:cmc}
\end{figure}
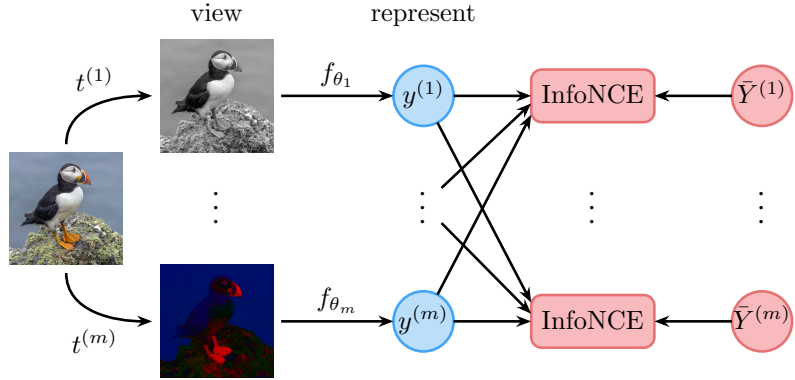

CMC \citep{tian2020contrastive} considers multiple views of the same scene and
tries to maximize the mutual information between those views. For each view, a
view-specific encoder extracts view-invariant features. A contrastive learning
objective forces the encoders to be as informative as possible about the other
views. In general, the views could be different sensory inputs of the same scene
(e.g., color, depth, surface normals). For vision tasks that only have access to
RGB images, the different views could be individual color channels. In this
paper, the authors consider the L and ab channels of an image after converting
it to the Lab color space. Note that the views can be interpreted as image
augmentations, however, each view uses the \emph{same} image augmentation for
all images, which is in contrast to other methods.

To apply CMC to images, we define  $m$ fixed image transformations ${[
t^{(1)}, \ldots, t^{(m)} ]}$. Let $X$ be a batch of images that gets transformed
into ${X^{(j)} = t^{(j)}(X)}$ for each view ${j \in \{ 1, \ldots, m \}}$. The
encoders ${f_{\theta_1}, \ldots, f_{\theta_m}}$ compute view-specific
representations ${y_i^{(j)} = f_{\theta_j}(x_i^{(j)})}$ for each ${j \in \{ 1,
\ldots, m \}}$. The representation used for downstream tasks can either be the
representation of a specific view or the concatenation of representations across
multiple or all views.

The idea of CMC is to apply the InfoNCE loss to pairs of views. Specifically,
the anchor is the representation $y_i^{(j)}$ of the $j$-th view, the positive is
the representation $y_i^{(k)}$ of the same image but from the $k$-th view, where
${k \neq j}$, and the negatives are representations from other images but also
from the $k$-th view. We denote the set of those negatives images by
$\bar{Y}_i^{(k)}$ which are obtained using a memory bank
\citep{wu2018unsupervised}. With memory banks, large batches of negative samples
can be obtained efficiently, at the cost of slightly outdated representations.

The loss for a single image $x_i$ accumulates the InfoNCE losses of all ordered
pairs of views, so the total loss function across the batch is given as
\begin{equation}
	\mathcal{L}_{\theta}^\text{CMC} = \frac{1}{n} \sum_{i=1}^n \sum_{j=1}^m \sum_{k=1;\,k \neq j}^m \infonce_{s_\tau\!}(y_i^{(j)}, y_i^{(k)}, \bar{Y}_i^{(k)}),
\end{equation}
where the similarities are calculated as ${s_\tau(y,y') = \scos(y, y')/\tau}$, i.e.,
as cosine similarity divided by a temperature hyperparameter $\tau > 0$, and
where ${\theta = [\theta_1, \ldots, \theta_m]}$ combines the parameters of all
encoders. See Figure~\ref{fig:cmc} for an illustration of the method.

%% file: figures/cmc.tex
\begin{tikzpicture}
  \node[inner sep=1mm] (x) at (0,0) {\includegraphics[width=1.5cm]{figures/transformations/puffin.jpg}};

  \node[inner sep=1mm] (xt1) at ($(x) + (2,1.5)$) {\includegraphics[width=1.5cm]{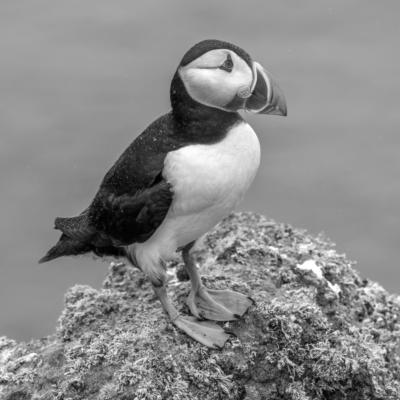}};
  \node[inner sep=1mm] (xtm) at ($(x) + (2,-1.5)$) {\includegraphics[width=1.5cm]{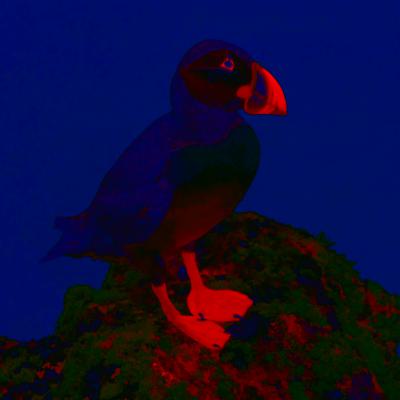}};

  \node[bluevar] (y1) at ($(xt1) + (2.75,0)$) {$y^{(1)}$};
  \node[bluevar] (ym) at ($(xtm) + (2.75,0)$) {$y^{(m)}$};

  \node[yshift=0.5mm] (xdots) at ($(x) + (2,0)$) {\rotatebox{90}{$\cdots$}};
  \node (ydots) at ($(xdots) + (2.75,0)$) {\rotatebox{90}{$\cdots$}};

  \node[nicelabel] (view) at ($(xt1) + (0,1.1)$) {view};
  \node[nicelabel] (repr) at ($(y1) + (0,1.1)$) {represent};

  \node[box, draw=nicedarkred, fill=nicered] (nce1) at ($(y1) + (2.25,0)$) {\footnotesize $\infonce$};
  \node[box, draw=nicedarkred, fill=nicered] (ncem) at ($(ym) + (2.25,0)$) {\footnotesize $\infonce$};

  \node[redvar] (yn1) at ($(nce1) + (2.25,0)$) {$\bar{Y}^{(1)}$};
  \node[redvar] (ynm) at ($(ncem) + (2.25,0)$) {$\bar{Y}^{(m)}$};

  \node (ncedots) at ($(ydots) + (2.25,0)$) {\rotatebox{90}{$\cdots$}};
  \node (yndots) at ($(ncedots) + (2.25,0)$) {\rotatebox{90}{$\cdots$}};

  \begin{scope}[transform canvas]  
    \draw[->,nicearrow] (xt1.east) -- node[midway,above,nicelabel] {$f_{\theta_1}$} (y1);
    \draw[->,nicearrow] (xtm.east) -- node[midway,above,nicelabel] {$f_{\theta_m}$} (ym);

    \draw[nicearrow] (x) edge[->,out=90,in=180,relative=false] node[midway,above,nicelabel] {$t^{(1)}$} (xt1);
    \draw[nicearrow] (x) edge[->,out=-90,in=180,relative=false] node[midway,below,nicelabel,yshift=-1.5mm] {$t^{(m)}$} (xtm);

    \draw[->,nicearrow] (y1) -- (nce1);
    \draw[->,nicearrow] (ym) -- (ncem);

    \draw[->,nicearrow] (yn1) -- (nce1);
    \draw[->,nicearrow] (ynm) -- (ncem);

    \draw[->,nicearrow] (ydots) -- ([yshift=-1mm]nce1.west);
    \draw[->,nicearrow] (ym) -- ([yshift=-2.5mm]nce1.west);

    \draw[->,nicearrow] (ydots) -- ([yshift=1mm]ncem.west);
    \draw[->,nicearrow] (y1) -- ([yshift=2.5mm]ncem.west);
  \end{scope}
\end{tikzpicture}

%% file: sections/simclr.tex
\subsection{A Simple Framework for Contrastive Learning of Visual Representations (SimCLR)}
\label{sec:simclr}

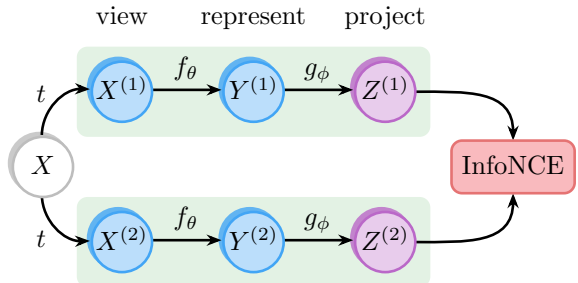
\begin{figure}[ht!]
	\centering \input{figures/simclr.tex}
	\caption{SimCLR defines the views in the batch that were constructed from different images as negative examples.}
	\label{fig:simclr_arch}
\end{figure}

\noindent The architecture used for SimCLR~\citep{chen2020simple} is similar to previous
methods like VICReg or Barlow Twins. Given a batch of images $X$, two views ${X^{(1)} = t(X)}$ and
${X^{(2)} = t(X)}$ are created using random transformations ${t \sim
\mathcal{T}}$. A Siamese encoder $f_\theta$ calculates representations ${Y^{(1)}
= f_\theta(X^{(1)})}$ and ${Y^{(2)} = f_\theta(X^{(2)})}$ which are then fed
into a Siamese projector $g_\phi$ to obtain projections $Z^{(1)} = {[z_1^{(1)},
\ldots, z_n^{(1)}]} = g_\phi(Y^{(1)})$ and $Z^{(2)} = {[z_1^{(2)}, \ldots,
z_n^{(2)}]} = g_\phi(Y^{(2)})$. Figures~\ref{fig:simclr_arch} gives an overview over this process.

SimCLR uses a contrastive loss to maximize the similarity between the two
projections of the same image while minimizing the similarity to projections of
other images. Specifically, for an image $x_i$ two InfoNCE losses are applied.
The first one uses the anchor $z_i^{(1)}$, the positive $z_i^{(2)}$, and the
negatives $\bar{Z}_i = {[z_1^{(1)}, z_1^{(2)}, \ldots, z_n^{(1)}, z_n^{(2)}]}
\setminus {\{ z_i^{(1)}, z_i^{(2)} \}}$, which are all projections from other
images in the batch. The second InfoNCE loss swaps the roles of anchor and
positive but uses the same set of negatives. Therefore, the loss function is
defined as
\begin{equation}
	\mathcal{L}^\text{SimCLR}_{\theta,\phi} = \dfrac{1}{n}\sum_{i=1}^{n} \frac{1}{2} \!\left[ \infonce_{s_\tau\!}(z_i^{(1)}, z_i^{(2)}, \bar{Z}_i) + \infonce_{s_\tau\!}(z_i^{(2)}, z_i^{(1)}, \bar{Z}_i) \right]\!,
\end{equation}
where the similarities are calculated as ${s_\tau(z, z') = \scos(z, z') / \tau}$,
i.e., the cosine similarity  divided by a temperature hyperparameter $\tau > 0$.

The transformations consist of a random cropping followed by a resize back to
the original size, a random color distortion, and a random Gaussian blur. A
ResNet is used as encoder $f_\theta$ and the projector $g_\phi$ is implemented
as an MLP with one hidden layer. To train SimCLR large batch sizes are used in
combination with the LARS optimizer~\citep{you2017large}. The authors note that their method does not
need memory banks \citep{wu2018unsupervised} as it is the case for other contrastive methods and is thus
easier to implement.

%% file: figures/simclr.tex
\begin{tikzpicture}
  \node[whitevar] (x) at (0,0) {$X$};

  \node[bluevar] (xt) at ($(x) + (1.05,1)$) {$X^{(1)}$};
  \node[bluevar] (y) at ($(xt) + (1.75,0)$) {$Y^{(1)}$};
  \node[purplevar] (z) at ($(y) + (1.75,0)$) {$Z^{(1)}$};

  \node[bluevar] (xtp) at ($(xt) - (0,2)$) {$X^{(2)}$};
  \node[bluevar] (yp) at ($(y) - (0,2)$) {$Y^{(2)}$};
  \node[purplevar] (zp) at ($(z) - (0,2)$) {$Z^{(2)}$};

  \node[nicelabel] (view) at ($(xt) + (0,1)$) {view};
  \node[nicelabel] (repr) at ($(y) + (0,1)$) {represent};
  \node[nicelabel] (proj) at ($(z) + (0,1)$) {project};
  
  \node[box,fill=nicered,draw=nicedarkred] (nce) at ($(x) + (6.25,0)$) {\footnotesize $\infonce$};

  \begin{pgfonlayer}{back}
    \path[box,fill=nicegreen,opacity=0.5] ($(xt) - (0.6,0.6)$) rectangle ($(z) + (0.6,0.6)$);
    \path[box,fill=nicegreen,opacity=0.5] ($(xtp) - (0.6,0.6)$) rectangle ($(zp) + (0.6,0.6)$);

    \draw[nicearrow] (z) edge[->,out=0,in=90,looseness=1.2,relative=false] (nce);
    \draw[nicearrow] (zp) edge[->,out=0,in=-90,looseness=1.2,relative=false] (nce);

    \begin{scope}[transparency group,opacity=0.8]
      \node[whitevar,fill=nicegray] at ($(x) + (-0.05,0.05)$) {};

      \node[bluevar,fill=nicedarkblue] at ($(xt) + (-0.05,0.05)$) {};
      \node[bluevar,fill=nicedarkblue] at ($(y) + (-0.05,0.05)$) {};
      \node[purplevar,fill=nicedarkpurple] at ($(z) + (-0.05,0.05)$) {};

      \node[bluevar,fill=nicedarkblue] at ($(xtp) + (-0.05,0.05)$) {};
      \node[bluevar,fill=nicedarkblue] at ($(yp) + (-0.05,0.05)$) {};
      \node[purplevar,fill=nicedarkpurple] at ($(zp) + (-0.05,0.05)$) {};
    \end{scope}

    \begin{scope}[transform canvas={xshift=-0.025cm,yshift=0.025cm}]
      \draw[nicearrow] (x) edge[->,bend left=45] node[midway,left,yshift=1mm,nicelabel] {$t$} (xt);
      \draw[->,nicearrow] (xt) -- node[midway,above,nicelabel] {$f_\theta$} (y);
      \draw[->,nicearrow] (y) -- node[midway,above,nicelabel] {$g_\phi$} (z);

      \draw[nicearrow] (x) edge[->,bend right=45] node[midway,left,yshift=-1.5mm,nicelabel] {$t$} (xtp);
      \draw[->,nicearrow] (xtp) -- node[midway,above,nicelabel] {$f_\theta$} (yp);
      \draw[->,nicearrow] (yp) -- node[midway,above,nicelabel] {$g_\phi$} (zp);

    \end{scope}
  \end{pgfonlayer}

\end{tikzpicture}

%% file: sections/moco.tex
\subsection{Momentum Contrast (MoCo)}
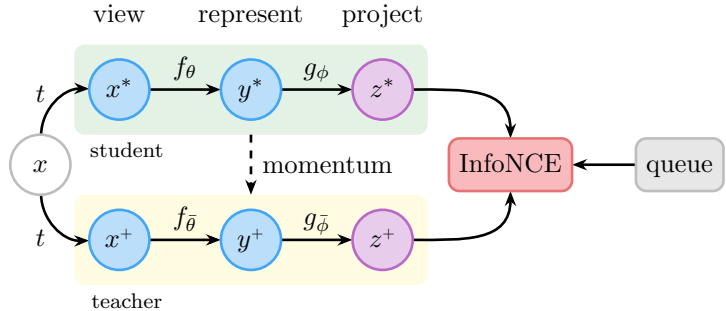
\begin{figure}[ht!]
  \centering
  \input{figures/mocov2.tex}
  \caption{MoCo computes projections of different views with a student and a
    teacher network, and minimizes the contrastive InfoNCE loss. The projections
    $z^*$ and $z^+$ are computed on-the-fly, whereas negative projections are
    cached in a queue of the most recent versions of $z^+$ from previous
    iterations, significantly improving computational efficiency.}
  \label{fig:mocov2_arch}
\end{figure}

\noindent Momentum Contrast \citep{he2020momentum} is a contrastive learning approach that uses a momentum encoder with an encoding queue to bridge the gap between
contrastive \emph{end-to-end} and \emph{memory bank} methods
\citep{wu2018unsupervised}. In essence, it allows the optimization of a
contrastive objective with significantly reduced computational costs, both in
terms of time and GPU memory \citep{chen2020improved}. Figure
\ref{fig:mocov2_arch} gives an overview of the architecture.

Similar to teacher-student methods (Section~\ref{sec:distillation}), MoCo
defines a student network consisting of an encoder $f_\theta$ and a projector
$g_\phi$ with parameters $\theta$ and $\phi$, and a teacher network consisting
of an encoder $f_{\bar{\theta}}$ and a projector $g_{\bar{\theta}}$ with
parameters $\bar{\theta}$ and $\bar{\phi}$. Given an image $x_i$, two views
${x_i^* = t(x_i)}$ and ${x_i^+ = t(x_i)}$ are created using random
transformations ${t \sim \mathcal{T}}$. The student computes the representation
$y_i^* = f_\theta(x_i^*)$ and projection $z_i^* = g_\phi(y_i^*)$, while the
teacher computes the representation ${y_i^+ = f_{\bar{\theta}}(x_i^+)}$ and
projection ${z_i^+ = g_{\bar{\phi}}(y_i^+)}$.

MoCo minimizes the InfoNCE loss to learn projections that are similar for two
views of the same image and dissimilar to projections of views of other images.
In our notation, the student computes the anchor $z_i^*$, the teacher computes
the positive $z_i^+$, and the selection of the negatives $\bar{Z}_i$ is
described below. The loss of MoCo is then defined as
\begin{equation}
	\mathcal{L}^\text{MoCo}_{\theta,\phi} = \dfrac{1}{n}\sum_{i=1}^{n} \infonce_{s_\tau\!}(z_i^*, z_i^+, \bar{Z}_i),
\end{equation}
where the similarities are calculated using the dot product ${s_\tau(z^*, z) =
z^\top z^* / \tau}$ divided by a temperature hyperparameter $\tau > 0$. The
teacher is updated by an exponential moving average of the student, i.e.,
\begin{align}
  \bar{\theta} & \leftarrow \alpha \bar{\theta} + (1 - \alpha) \theta, \\
  \bar{\phi} & \leftarrow \alpha \bar{\phi} + (1 - \alpha) \phi
\end{align}
where $\alpha \in [0, 1]$ controls the rate at which the weights of the teacher
network are updated with the weights of the student network.

In an \emph{end-to-end} setting, the negatives are computed on-the-fly in one
batch (see SimCLR, Section \ref{sec:simclr}), resulting in relatively large
resource consumption. In contrast, memory banks \citep{wu2018unsupervised}
describe the concept of saving projections for all items in the dataset,
drastically reducing resource consumption but introducing potential negative
effects from inconsistent or outdated projections. MoCo aims to combine the
benefits of both end-to-end training and memory banks. Similar to memory banks,
MoCo only computes projections of the positives and saves them for reuse in
later iterations. Instead of saving projections for all images in the dataset,
MoCo uses a queue to cache only the last $K$ computed projections, thus avoiding
outdated projections. Since older projections are removed from the queue, saved
projections no longer require momentum updates. The teacher provides the
projections that are to be cached, while the student is updated via
backpropagation using the contrastive loss.

\paragraph{MoCo v2.} The second version of MoCo  \citep{chen2020improved}
introduces several smaller changes to further improve downstream performance and
outperform SimCLR. The most notable changes include the replacement of the
linear projection layer of MoCo with an MLP, as well as the application of a
cosine learning rate scheduler \citep{loshchilov2017sgdr} and additional
augmentations. The new 2-layer MLP head was adopted following SimCLR. Note, that
the MLP is only used during unsupervised training and is not intended for
downstream tasks. In terms of additional augmentations, MoCo v2 also adopts the
blur operation used in SimCLR.

%% file: figures/mocov2.tex
\begin{tikzpicture}
  \node[whitevar] (x) at (0,0) {$x$};

  \node[bluevar] (xt) at ($(x) + (1.05,1)$) {$x^*$};
  \node[bluevar] (y) at ($(xt) + (1.75,0)$) {$y^*$};
  \node[purplevar] (z) at ($(y) + (1.75,0)$) {$z^*$};

  \node[bluevar] (xtp) at ($(xt) - (0,2)$) {$x^\tplus$};
  \node[bluevar] (yp) at ($(y) - (0,2)$) {$y^\tplus$};
  \node[purplevar] (zp) at ($(z) - (0,2)$) {$z^\tplus$};
  
  \node[nicelabel] (student) at ($(xt) + (0.1,-0.8)$) {\scriptsize student};
  \node[nicelabel] (teacher) at ($(xtp) + (0.1,-0.8)$) {\scriptsize teacher};

  \node[box,fill=nicered,draw=nicedarkred] (nce) at ($(x) + (6.25,0)$) {\footnotesize $\infonce$};
  \node[box,fill=nicelightgray,draw=nicegray] (queue) at ($(nce) + (2.25,0)$) {\footnotesize queue};

  \node[nicelabel] (view) at ($(xt) + (0,1)$) {view};
  \node[nicelabel] (repr) at ($(y) + (0,1)$) {represent};
  \node[nicelabel] (proj) at ($(z) + (0,1)$) {project};

  \begin{pgfonlayer}{back}
    \path[box,fill=nicegreen,opacity=0.5] ($(xt) - (0.6,0.6)$) rectangle ($(z) + (0.6,0.6)$);
    \path[box,fill=niceyellow,opacity=0.5] ($(xtp) - (0.6,0.6)$) rectangle ($(zp) + (0.6,0.6)$);

    \begin{scope}[transform canvas]  
      \draw[nicearrow] (x) edge[->,bend left=45] node[midway,left,nicelabel,yshift=1mm] {$t$} (xt);
      \draw[->,nicearrow] (xt) -- node[midway,above,nicelabel] {$f_\theta$} (y);
      \draw[->,nicearrow] (y) -- node[midway,above,nicelabel] {$g_\phi$} (z);

      \draw[nicearrow] (x) edge[->,bend right=45] node[midway,left,nicelabel,left,yshift=-1.5mm] {$t$} (xtp);
      \draw[->,nicearrow] (xtp) -- node[midway,above,nicelabel] {$f_{\bar{\theta}}$} (yp);
      \draw[->,nicearrow] (yp) -- node[midway,above,nicelabel] {$g_{\bar{\phi}}$} (zp);

      \draw[->,nicearrow,dashed] ($(y) - (0,0.6)$) -- node[midway,right] {\footnotesize momentum} ($(yp) + (0,0.6)$);

      \draw[nicearrow] (z) edge[->,out=0,in=90,looseness=1.2,relative=false] (nce);
      \draw[nicearrow] (zp) edge[->,out=0,in=-90,looseness=1.2,relative=false] (nce);
      \draw[->,nicearrow] (queue) -- (nce);
    \end{scope}
  \end{pgfonlayer}
\end{tikzpicture}

%% file: sections/pirl.tex
\subsection{Pretext-Invariant Representation Learning (PIRL)}

In the previously introduced pretext tasks we compute the representations of transformed images to predict properties from specific transformations, i.e., rotation angles~\citep{noroozi2016unsupervised} or patch permutations~\citep{gidaris2018unsupervised}. 
In this way, representations are encouraged to be covariant to the specific transformation, but are not guaranteed to capture the same underlying semantic information regardless of the transformation used. 
Although such covariance is advantageous in certain cases, we are more interested in representations that are semantically meaningful, so it is desirable to learn representations that are invariant to the transformation. 
In order to achieve this, \cite{misra2020self} refined the pretext task loss formulation and developed an approach called Pretext-Invariant Representation Learning~(PIRL) which also makes use of memory banks \citep{wu2018unsupervised}.

\begin{figure}[ht!]
    \centering
    \input{figures/pirl.tex}
    \caption{Architecture of PIRL. Minimizing a contrastive loss promotes similarity between the representations of the image and its corresponding transformation.}
    \label{fig:pirl_arch}
  \end{figure}
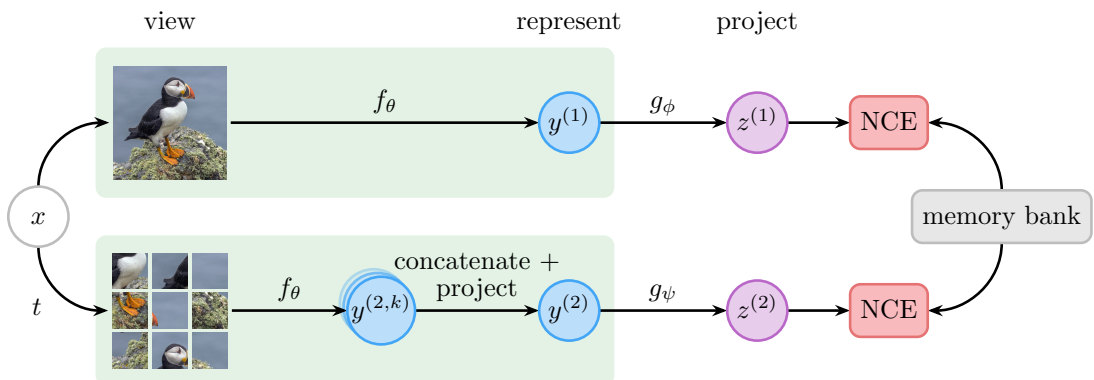

The goal of PIRL is to train an encoder network~$f_\theta$ that maps images~${x_i^{(1)} = x_i}$ and transformed images~${x_i^{(2)} = t_\pi(x_i)}$ to representations $y_i^{(1)}$ and $y_i^{(2)}$, respectively, which are invariant to the transformations used.
Analogous to Section~\ref{sec:jigsaw}, $t_\pi$ denotes a jigsaw transformation consisting of a random permutation of image patches, where $\pi$ is the corresponding permutation.
The loss formulation of pretext tasks as defined in Section~\ref{sec:pretext} emphasizes that the encoder learns representations that contain information about the transformation rather than semantics.
Let $z_i^{(1)} = g_\phi(f_\theta(x^{(1)}_i))$ and $z_i^{(2)} = g_\psi(f_\theta(x^{(2)}_i))$ be the projections obtained by the encoder $f_\theta$ and two separate projectors $g_\phi$ and $g_\psi$. The network is trained by minimizing a convex combination of two noise contrastive estimators~(NCE)~\citep{gutmann2010noise}
\begin{align} \label{eq:pirl-1}
    \mathcal{L}^\text{PIRL}_{\theta, \phi, \psi} = \frac{1}{n} \sum_{i=1}^n \lambda \ell_\text{NCE}\left(m_i, z^{(2)}_i, \bar{M}_i\right) + (1 - \lambda) \ell_\text{NCE}\left(m_i, z^{(1)}_i, \bar{M}_i\right),
\end{align}
where $m_i$ is a projection from a memory bank corresponding to the original image $x_i$, each positive sample is assigned a randomly drawn set of negative projections $\bar{M}_i$ of images other than $x_i$ obtained from the memory bank, and ${\lambda \in [0,1]}$ is a hyperparameter.
In contrast to the previously introduced pretext tasks, the loss formulation of PIRL does not explicitly aim to predict particular properties of the applied transformations, such as rotation or patch indices.
Instead, it is solely defined on images and their corresponding transformed counterparts.
NCE applies binary classification to each data point to distinguish positive and negative samples. 
Here, the NCE loss is formulated as
\begin{align}\label{eq:pirl-nce}
    \ell_{\text{NCE}}(m,z,\bar{M}) = - \log [h(m, z, \bar{M})] - \sum_{\bar{m} \,\in\, \bar{M}} \log [1 - h(z, \bar{m}, \bar{M})],
\end{align}
where $h$ models the probability that $(x_i, x'_i)$ is derived from $X$ as 
\begin{align}
    h(u,v,\bar{M}) = \frac{\exp(\scos(u,v)/\tau)}{\exp(\scos(u,v)/\tau) + \sum_{\bar{m} \,\in\, \bar{M}} \exp(\scos( \bar{m}, v)/\tau)}
\end{align}
for a temperature $\tau > 0$. 
Considering that the projections depend on the intermediate representations, the individual terms in Equation~\ref{eq:pirl-nce} encourage $y_i^{(1)}$ to be similar to $y_i^{(2)}$ and also $y_i^{(2)}$ to be dissimilar to the representations of other images. 
Since this formulation alone does not compare features between different untransformed images, the authors propose to use the convex combination of two NCE losses as defined in Equation~\ref{eq:pirl-1}.  An overview of this approach is illustrated in Figure~\ref{fig:pirl_arch}.
The encoder network~$f_\theta$ consists of the final layer of ResNet50~\citep{he2016deep}, average pooling and a 128-dimensional fully connected layer.
As for the image transformation in the lower branch of Figure~\ref{fig:pirl_arch}, we first extract nine image sections and apply them individually to $f_\theta$ to obtain patch representations $y_i^{(2,k)}$.
These are then randomly concatenated and sent through another fully connected layer to obtain the 128-dimensional representation~$y_i^{(2)}$.
Although the authors focus their work on the Jigsaw pretext task, their approach can be generalized to any other pretext task. For demonstration purposes, the authors also conduct experiments with the rotation pretext task and its combination with the Jigsaw task. 
In this way we have to adapt the lower branch of Figure~\ref{fig:pirl_arch} by transforming the image at the beginning and feeding forward the transformed image to achieve the representation~$y_i^{(2)}$ directly. Thus, using a secondary fully connected layer is not necessary anymore.

Note that PIRL can also be classified as a pretext task approach as defined in
Section~\ref{sec:pretext}. However, it also uses ideas of contrastive representation learning which is why we decided to discuss it at this point.

%% file: figures/pirl.tex
\begin{tikzpicture}
  \node[whitevar] (x) at (0,0) {$x$};

  \node[inner sep=0.5mm] (xt) at (1.75,1.25) {\includegraphics[width=1.5cm]{figures/jigsaw/puffin.jpg}};
  \node[bluevar] (y) at ($(xt.east) + (4.5,0)$) {$y^{(1)}$};
  \node[purplevar] (z) at ($(y) + (2.5,0)$) {$z^{(1)}$};

  \begin{scope}[xshift=1.75cm,yshift=-1.25cm,x=0.53cm,y=0.53cm]
    \node at (-1,1) {\includegraphics[width=0.47cm]{figures/jigsaw/crop4.png}};
    \node[inner sep=0.5mm] (xtpleft) at (-1,0) {\includegraphics[width=0.47cm]{figures/jigsaw/crop5.png}};
    \node at (-1,-1) {\includegraphics[width=0.47cm]{figures/jigsaw/crop2.png}};
    \node at (0,1) {\includegraphics[width=0.47cm]{figures/jigsaw/crop1.png}};
    \node (xtp) at (0,0) {\includegraphics[width=0.47cm]{figures/jigsaw/crop6.png}};
    \node at (0,-1) {\includegraphics[width=0.47cm]{figures/jigsaw/crop3.png}};
    \node at (1,1) {\includegraphics[width=0.47cm]{figures/jigsaw/crop0.png}};
    \node[inner sep=0.5mm] (xtpright) at (1,0) {\includegraphics[width=0.47cm]{figures/jigsaw/crop8.png}};
    \node at (1,-1) {\includegraphics[width=0.47cm]{figures/jigsaw/crop7.png}};
  \end{scope}

  \node[bluevar, minimum size=0.9cm] (yp) at ($(y) - (2.5,2.5)$) {$y^{(2,k)}$};
  \node[bluevar] (ypp) at ($(yp) + (2.5,0)$) {$y^{(2)}$};
  \node[purplevar] (zp) at ($(z) - (0,2.5)$) {$z^{(2)}$};

  \node[box,fill=nicered,draw=nicedarkred] (nce1) at ($(z) + (1.75,0)$) {\footnotesize NCE};
  \node[box,fill=nicered,draw=nicedarkred] (nce2) at ($(zp) + (1.75,0)$) {\footnotesize NCE};

  \node[box,fill=nicelightgray,draw=nicegray] (bank) at ($(nce1)!0.5!(nce2) + (1.5,0)$) {\footnotesize memory bank};

  \node[nicelabel] (view) at ($(xt) + (0,1.35)$) {\footnotesize view};
  \node[nicelabel] (repr) at ($(y) + (0,1.35)$) {\footnotesize represent};
  \node[nicelabel] (proj) at ($(z) + (0,1.35)$) {\footnotesize project};

  \begin{pgfonlayer}{back}
    \path[box,fill=nicegreen,opacity=0.5] let \p1=(xtpleft),\p2=(xt),\p3=(y) in ($(\x1,\y2) - (0.46,1)$) rectangle ($(\x3,\y2) + (0.6,1)$);
    \path[box,fill=nicegreen,opacity=0.5] let \p1=(xtpleft),\p2=(ypp) in ($(\x1,\y1) - (0.46,1)$) rectangle ($(\x2,\y1) + (0.6,1)$);

    \node[bluevar,opacity=0.33,minimum size=0.9cm] at ($(yp) + (-0.1,0.1)$) {};
    \node[bluevar,opacity=0.66,minimum size=0.9cm] at ($(yp) + (-0.05,0.05)$) {};

    \begin{scope}[transform canvas]  
      \draw[nicearrow] (x) edge[->,out=90,in=180,relative=false] (xt.west);
      \draw[->,nicearrow] (xt.east) -- node[midway,above,nicelabel] {$f_\theta$} (y);
      \draw[->,nicearrow] (y) -- node[midway,above,nicelabel] {$g_\phi$} (z);

      \draw[nicearrow] (x) edge[->,out=-90,in=180,relative=false] node[midway,below,nicelabel,left,xshift=-1mm,yshift=-2mm] {$t$} (xtpleft.west);
      \draw[->,nicearrow] (xtpright.east) -- node[midway,above,nicelabel] {$f_\theta$} (yp);
      \draw[->,nicearrow] (ypp) -- node[midway,above,nicelabel] {$g_\psi$} (zp);
      \draw[->,nicearrow] (yp) -- node[above,nicelabel,align=center] {\footnotesize concatenate +\\[-2pt] \footnotesize project} (ypp);

       \draw[->,nicearrow] (bank) edge[->,out=90,in=0,relative=false] (nce1.east);
       \draw[->,nicearrow] (bank) edge[->,out=-90,in=0,relative=false] (nce2.east);
       \draw[->,nicearrow] (z) -- (nce1);
       \draw[->,nicearrow] (zp) -- (nce2);
    \end{scope}
  \end{pgfonlayer}
\end{tikzpicture}

%% file: sections/cluster.tex
\section{Clustering-based Methods}
\label{sec:clustering}
So far, some of the presented representation learning methods define a classification problem with hand-crafted labels to solve an auxiliary task (see Section~\ref{sec:pretext}).  Instead of specifying these class labels by hand, clustering algorithms, e.g.,~k-means~\cite{lloyd1982least}, can be used to create the labels in an unsupervised fashion.

The objective of clustering-based representation learning is to group images with similar representations into clusters.
In contrastive learning, for example, this would allow us to discriminate between cluster assignments rather than individual images or representations, which significantly increases efficiency.
Over the time, numerous clustering-based methods have been developed, each with its own strengths and weaknesses.
In the subsequent sections, we present the most significant approaches.

\subsection{DeepCluster}

The first approach that implements the idea of clustering for representation learning is DeepCluster~\citep{deep-cluster}, which alternates between inventing pseudo-labels via cluster assignments and adjusting the representation to classify images according to their invented labels. 
The motivation behind this is to increase the performance of convolutional architectures that already exhibit a strong inductive bias, as these already perform reasonably well with randomly initialized weights~\citep{noroozi2016unsupervised}. 
Overall, the authors propose to repeatedly alternate between the following two steps to further improve the encoder network: 
\begin{enumerate}
	\item Group the representations $y_i = f_\theta(x_i)$ produced by the current state of the encoder~$f_\theta$ to $k$ clusters (e.g., by using $k$-means clustering).
	\item Use the cluster assignments from step 1 as pseudo-labels~$\beta_i$ for supervision to update the weights, i.e.,
    \begin{equation}
      \mathcal{L}^\text{DeepCluster}_{\theta, \psi} \frac{1}{n} \sum_{i=1}^n \dsoftmax(q_\psi(y_i), \beta_i),
	  \end{equation}
	where a predictor network $q_\psi$ tries to predict the cluster assignments of the representations ${y_i = f_\theta(x_i)}$.
\end{enumerate}
In their experiments, the authors utilize a standard AlexNet~\citep{krizhevsky2017imagenet} with $k$-means and argue that the choice of the clustering algorithm is not crucial.

\subsection{Self Labelling (SeLa)}
A common weakness of the naive combination of clustering and representation learning is its proneness to degenerate solutions where, for example, all representations are assigned to the same cluster. 
To circumvent this issue, \cite{sela} have developed an refined alternating update scheme, called Self Labelling~(SeLa), 
which imposes constraints on the labels such that each cluster is assigned the same amount of data points.
The pseudo-labels corresponding to images ${x_1, \ldots, x_n}$ are encoded as one-hot vectors ${\beta_1, \ldots, \beta_n \in \{0,1\}^k}$. To assign the pseudo-labels ${\beta_1, \ldots, \beta_n}$ and fit encoder and predictor networks $f_\theta$ and $q_\psi$, respectively, the authors consider the optimization problem
\begin{align}
	&\! \min_{\beta, \theta, \psi}\;\; \frac{1}{n} \sum_{i=1}^n \dce(q_\psi(f_\theta(x_i)), \beta_i), \\
	&  \text{s.t.}\;\;\; \sum_{i=1}^n \beta_i[j] = \frac{n}{k}, \quad \beta_i[j] \in \{0,1\} \quad \text{for } j \in \{1, \ldots, k\},
\end{align} 
which they solve by alternating between the following two steps:
\begin{enumerate}
  \item The problem of assigning the pseudo-label to the images is formulated as an optimal transport problem which is solved using a fast variant of the Sinkhorn-Knopp algorithm~\citep{sinkhorn-knopp}.
  \item Fix the pseudo-labels from step 1 and update the parameters $\theta$ and $\psi$ by minimizing the cross-entropy loss.
\end{enumerate}
Note that step 2 is the same as used in DeepCluster.
However, in DeepCluster it is possible for all data points to be grouped into a single cluster, which results in a constant representation being learned and consequently the minimum being achieved in both optimization steps. 

\subsection{Swapping Assignments Between Multiple Views of the Same Image (SwAV)}

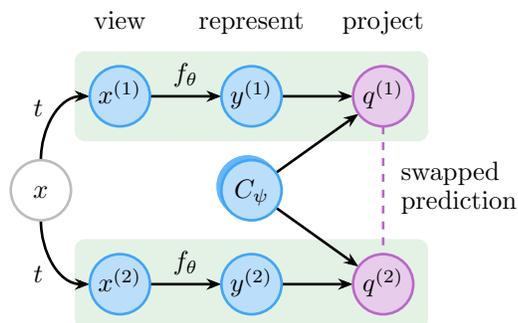
\begin{figure}[ht!]
  \centering
  \input{figures/swav.tex}
  \caption{SwAV does not actually measure the similarity between image representations of different views, but instead compares the representations with codes obtained by assigning the features to parameterized prototypes.}
  \label{fig:swav_arch}
\end{figure}

In general, contrastive methods are computationally challenging due to the need for numerous explicit pairwise feature comparisons.
However, \cite{swav} propose an alternative algorithm, called SwAV, that circumvents this problem by data clustering while promoting consistency among cluster assignments across different views.
In contrast to DeepCluster and SeLa, SwAV is an online clustering-based approach, i.e., it does not alternate between a cluster assignment and training step.
An encoder network $f_\theta$ is used to compute image representations $y^{(1)}$ and $y^{(2)}$ of two views of the same image $x$.
These representations are then mapped to a set of $k$ parameterized prototypes~$C_\psi = [c_1, \ldots, c_k]$, resulting in corresponding codes $q^{(1)}$ and $q^{(2)}$.
Next, a swapped prediction problem is addressed, where the codes derived from one view are predicted using the encoding from the second view.
To achieve this, we minimize 
\begin{align}
  \mathcal{L}^\text{SwAV}_{\theta, \psi} = \frac{1}{n} \sum_{i=1}^n \ell(q_i^{(1)},y_i^{(2)}) + \ell(q_i^{(2)},y_i^{(1)}),
\end{align}
where $\ell(q, y) = \dce(q, \softmax_\tau(C^\top y))$ quantifies the correspondence between the representation~$y$ and the code~$q$ for a temperature $\tau > 0$.
For an overview of the architecture we refer to Figure~\ref{fig:swav_arch}.
Note that, although SwAV takes advantage of contrastive learning, it does not require the use of a large memory bank or a momentum network.

In addition to this method, the authors also propose the augmentation technique called \textit{multi-crop}, which was also used for DINO (see Section~\ref{sec:dino}).
Instead of using two views with full resolution, a mixture of views with different resolutions is used. In this approach, multiple transformations are compared by using considerably smaller ones, which leads to a further improvement of previous methods such as SimCLR, DeepCluster and SeLa.

%% file: figures/swav.tex
\begin{tikzpicture}
  \node[whitevar] (x) at (0,0) {$x$};

  \node[bluevar] (xt) at ($(x) + (1.05,1.25)$) {$x^{(1)}$};
  \node[bluevar] (y) at ($(xt) + (1.75,0)$) {$y^{(1)}$};
  \node[purplevar] (z) at ($(y) + (1.75,0)$) {$q^{(1)}$};

  \node[bluevar] (xtp) at ($(xt) - (0,2.5)$) {$x^{(2)}$};
  \node[bluevar] (yp) at ($(y) - (0,2.5)$) {$y^{(2)}$};
  \node[purplevar] (zp) at ($(z) - (0,2.5)$) {$q^{(2)}$};

  \node[bluevar] (c) at ($(y) - (0,1.25)$) {$C_\psi$};

  \node[nicelabel] (view) at ($(xt) + (0,1)$) {view};
  \node[nicelabel] (repr) at ($(y) + (0,1)$) {represent};
  \node[nicelabel] (proj) at ($(z) + (0,1)$) {project};

  \begin{pgfonlayer}{back}
    \path[box,fill=nicegreen,opacity=0.5] ($(xt) - (0.6,0.6)$) rectangle ($(z) + (0.6,0.6)$);
    \path[box,fill=nicegreen,opacity=0.5] ($(xtp) - (0.6,0.6)$) rectangle ($(zp) + (0.6,0.6)$);

    \begin{scope}[transparency group,opacity=0.8]
      \node[bluevar,fill=nicedarkblue] at ($(c) + (-0.05,0.05)$) {};
    \end{scope}

    \begin{scope}[transform canvas]  
      \draw[nicearrow] (x) edge[->,out=90,in=180,nicelabel,relative=false] node[midway,left,nicelabel,yshift=1mm] {$t$} (xt);
      \draw[->,nicearrow] (xt) -- node[midway,above,nicelabel] {$f_\theta$} (y);
      \draw[->,nicearrow] (y) -- (z);

      \draw[nicearrow] (x) edge[->,out=-90,in=180,relative=false] node[midway,left,nicelabel,yshift=-1.5mm] {$t$} (xtp);
      \draw[->,nicearrow] (xtp) -- node[midway,above,nicelabel] {$f_\theta$} (yp);
      \draw[->,nicearrow] (yp) -- (zp);

      \draw[->,nicearrow] (c) -- (z);
      \draw[->,nicearrow] (c) -- (zp);

      \draw[nicearrow,dashed,nicedarkpurple] (z) -- node[midway,right,nicelabel,black,align=left,xshift=1mm,yshift=-1.5mm] {swapped \\[-1pt] prediction} (zp);
    \end{scope}
  \end{pgfonlayer}
\end{tikzpicture}

%% file: sections/taxonomy.tex
\section{Taxonomy of Representation Learning Methods}
\label{sec:taxonomy}

As we have seen in this survey, there are several ways to learn meaningful representations of images. These include solving specific tasks for pre-training such as predicting the rotation angle of an image, maximizing the mutual information between different views of the same image, using a contrastive loss in order to separate positive and negative samples in latent space, learning from a teacher network and clustering and subsequently self-labeling images. Based on these distinctions, we adapt and expand the taxonomy, proposed by \citet{bardes2021vicreg} in the following section, which includes the following five categories:
\begin{enumerate}
	\item Pretext Tasks Methods
	\item Information Maximization
	\item Teacher-Student Methods
	\item Contrastive Representation Learning
	\item Clustering-based Methods
\end{enumerate}
Note that some methods fit into multiple categories, as they combine different approaches. CPC (v2) and CMC, e.g., both use a contrastive loss, as well as information maximization. PIRL includes solving a pre-text  and a contrastive loss to learn representations.

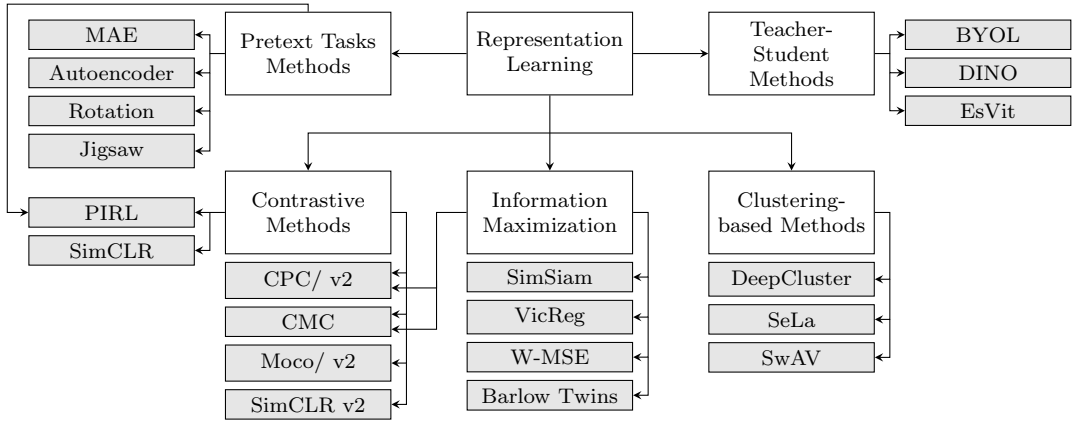
\begin{figure}
\scriptsize
\centering
\begin{tikzpicture}[>=stealth, 
category/.style={shape=rectangle,draw, align=center,text width=2cm, minimum height=1.1cm},
method/.style={shape=rectangle,draw, align=center,text width=2cm, fill=gray!20, node distance=0.1cm and 0.4cm}]
    \node (ptm) [category]{Pretext Tasks Methods};
    \node (rl) [category, right=of ptm]{Representation Learning};
    \node (ts) [category,right=of rl]{Teacher-Student Methods};
    \node (cm) [category,below =of ts]{Clustering-based Methods};
    \node (crl) [category,below =of ptm]{Contrastive Methods};
    \node (im) [category,right=of crl]{Information Maximization};
    \node (mae) [method, left=of ptm, yshift=0.25cm]{MAE};
    \node (ae) [method, below=of mae]{Autoencoder};
    \node (rot) [method, below=of ae]{Rotation};
    \node (jig) [method, below=of rot]{Jigsaw};
    \node (pirl) [method, left=of crl]{PIRL};
    \node (sclr) [method, below=of pirl]{SimCLR};
    \node (cpc) [method, below=of crl]{CPC/ v2};
	\node (cmc) [method, below=of cpc]{CMC};
	\node (moco) [method, below=of cmc]{Moco/ v2};
    \node (sclr2) [method, below=of moco]{SimCLR v2};
    \node (sim) [method, below=of im]{SimSiam};
    \node (vic) [method, below=of sim]{VicReg};
    \node (wmse) [method, below=of vic]{W-MSE};
    \node (barlow) [method, below=of wmse]{Barlow Twins};
    \node (byol) [method, right=of ts, yshift=0.25cm]{BYOL};
    \node (dino) [method, below=of byol]{DINO};
    \node (esvit) [method, below=of dino]{EsVit};
    \node (dc) [method, below=of cm]{DeepCluster};
    \node (sela)[method, below=of dc]{SeLa};
    \node (swav)[method, below=of sela]{SwAV};

    \draw[->] (ptm.north) |- ++(0,0.1) |- ++(-4,0) |-(pirl.west) ; 
    \draw[->] (rl) -- (ptm);
    \draw[->] (rl) -- (ts);
    \draw[->] (rl.south) ++(0,-0.5) -|(cm.north);
    \draw[->] (rl.south) ++(0,-0.5) -|(crl.north);
    \draw[->] (rl) -- (im);
    
    \draw[->] (ptm.west) -- ++(-0.2,0) |- (mae);
    \draw[->] (ptm.west) ++(-0.2,0) |-(ae.east);
    \draw[->] (ae.east) ++(0.2,0) |-(rot.east);
    \draw[->] (rot.east) ++(0.2,0) |-(jig.east);
    
    \draw[->] (ts.east) -- ++(0.2,0) |- (byol);
    \draw[->] (crl) -- (pirl);
    \draw[->] (ts.east) ++(0.2,0) |-(dino.west);
    \draw[->] (dino.west) ++(-0.2,0) |-(esvit.west);
    \draw[->] (crl.west) ++(-0.2,0) |-(sclr.east);
    \draw[->] (crl.east) -| ++(0.2,0) |-([yshift=1mm]cpc.east);
    \draw[->] ([yshift=1mm]cpc.east) ++(0.2,0) |-([yshift=1mm]cmc.east);
    \draw[->] ([yshift=1mm]cmc.east) ++(0.2,0) |-(moco.east);
    \draw[->] (moco.east) ++(0.2,0) |-(sclr2.east);

    \draw[->] (im.east) -| ++(0.2,0) |-(sim.east);
	\draw[->] (sim.east) ++(0.2,0) |-(vic.east);
	\draw[->] (vic.east) ++(0.2,0) |-(wmse.east);
	\draw[->] (wmse.east) ++(0.2,0) |-(barlow.east);

    \draw[->] (cm.east) -| ++(0.2,0) |-(dc.east);
    \draw[->] (dc.east) ++(0.2,0) |-(sela.east);
    \draw[->] (sela.east) ++(0.2,0) |-(swav.east);
    
    \draw[->] (im.west) -| ++(-0.4,0) |-([yshift=-1mm]cpc.east);
    \draw[->] (cpc.east) ++(0.6,0) |-([yshift=-1mm]cmc.east);

\end{tikzpicture}
	\caption{Graphical overview of the taxonomy of image representation learning methods.}
	\label{fig:taxonomy}
\end{figure}

\begin{sidewaystable}
	\centering
	\footnotesize
	\caption{Overview over the representation learning approaches discussed in
		this paper.}
	\label{tab:overview}
	\begin{tabular}{lcl}
		\toprule
		Method & Class & Code \\
		\midrule
		Autoencoder \citep{le1987modeles} & Generative & \\
		Rotation \citep{gidaris2018unsupervised}& Pretext-Task& \scriptsize{\url{https://github.com/gidariss/FeatureLearningRotNet}}\\
		Jigsaw \citep{noroozi2016unsupervised}& Pretext-Task& \scriptsize{\url{https://github.com/MehdiNoroozi/JigsawPuzzleSolver}}\\
		MAE \citep{he2022masked}& Pretext-Task &\scriptsize{\url{https://github.com/facebookresearch/mae}}\\
		DINO \citep{caron2021emerging} & Teacher-Student & \scriptsize{\url{https://github.com/facebookresearch/dino}}\\
		EsViT \citep{li2021efficient} & Teacher-Student & \scriptsize{\url{https://github.com/microsoft/esvit}}\\
		BYOL \citep{grill2020bootstrap} & Teacher-Student & \scriptsize{\url{https://github.com/deepmind/deepmind-research/tree/master/byol}} \\
		VicReg \citep{bardes2021vicreg} & Info-Max & \scriptsize{\url{https://github.com/facebookresearch/vicreg}}\\
		Barlow Twins \citep{zbontar2021barlow}& Info-Max & \scriptsize{\url{https://github.com/facebookresearch/barlowtwins}}\\
		SimSiam \citep{chen2021exploring} & Info-Max   & \scriptsize{\url{https://github.com/facebookresearch/simsiam}}\\
		W-MSE \citep{ermolov2021whitening} & Info-Max & \scriptsize{\url{https://github.com/htdt/self-supervised}}\\
		SimCLR \citep{chen2020simple} &   Contrastive& \scriptsize{\url{https://github.com/google-research/simclr}}\\
		MoCo \citep{he2020momentum} &  Contrastive & \scriptsize{\url{https://github.com/facebookresearch/moco}}\\
		MoCo v2 \citep{chen2020improved} &  Contrastive & \scriptsize{\url{https://github.com/facebookresearch/moco}}\\
		CPC \citep{oord2018representation} &  Contrastive/Info-Max\\
		CPC v2 \citep{henaff2020data} &  Contrastive/Info-Max\\
		CMC \citep{tian2020contrastive} & Contrastive/Info-Max& \scriptsize{\url{https://github.com/HobbitLong/CMC}}\\
		PIRL \citep{misra2020self} & Contrastive/Pretext-Task & \scriptsize{\url{https://github.com/facebookresearch/vissl/tree/main/projects/PIRL}}\\
		DeepCluster \citep{deep-cluster} & Clustering & \scriptsize{\url{https://github.com/facebookresearch/deepcluster}}\\
		SeLa \citep{sela} & Clustering & \scriptsize{\url{https://github.com/yukimasano/self-label}}\\
		SwAV \citep{swav} & Clustering & \scriptsize{\url{https://github.com/facebookresearch/swav}}\\
		\bottomrule
	\end{tabular}
\end{sidewaystable}

Figure \ref{fig:taxonomy} gives a visual overview on all methods within the proposed taxonomy. The inner nodes show the five categories that are connected to each reviewed method. Methods that can be assigned to several categories have multiple ingoing edges. As an additional overview, Table \ref{tab:overview} lists all methods including their primary class assignment and the URL to the original implementation on Github\footnote{\url{https://github.com}}, if available.

%% file: sections/comparison.tex
\section{Meta-Study of Quantitative Results}
\label{sec:meta}
Accessing the quality of the learned representations can be tricky. One approach that has become established in literature is to evaluate the obtained representations on downstream computer vision tasks, such as image classification, object detection, or instance segmentation. 
In this section we explain the evaluation process for representation learning methods and take a closer look at the most common evaluation tasks in literature. We compare all reviewed methods with regard to their performance and report the results on three different datasets.
We perform a quantitative comparison and provide some insights on potential future directions to further evaluate and compare the reviewed methods. 

\subsection{Evaluation of Representation Learning Methods}

The performance of representation learning models is often measured and compared by letting  pre-trained models solve downstream tasks such as image classification. For pre-training,  a base architecture is chosen as encoder and trained in a self-supervised manner without the use of labels. Many authors experiment with multiple architectures. One default architecture for image classification is the ResNet-50~\citep{he2016deep}, newer methods often use Vision Transformers (ViT) \citep{dosovitskiy2020image} as encoder.
The learned representations are then evaluated on different downstream tasks. We give more details on the evaluation protocols later in this section.

We identified five datasets that were most commonly used to evaluate representation learning methods: ImageNet~\citep{russakovsky2015imagenet}, the Pascal visual object classes (VOC)~\citep{everingham2009pascal}, Microsoft common objects in context (COCO)~\citep{lin2014microsoft}, CIFAR-10, CIFAR-100~\citep{krizhevsky2009learning} and Places205~\citep{zhou2014learning}. All listed datasets include one or multiple of the following tasks: image classification (IC), object detection (OD) and instance segmentation (Seg).
Table~\ref{tab:datasets} shows which of the most common datasets are used for evaluating each method. 
\begin{table}
\centering
\footnotesize
\caption{Overview on datasets (ImageNet, Pascal VOC, Microsoft COCO, CIFAR and Places-205) and tasks (image classification, object detection and instance segmentation) each representation learning method has been evaluated on. The numbers underneath indicate how many of the shown 20 methods used the corresponding dataset for evaluation.}
\label{tab:datasets}
	\begin{tabular}{lccccccccc}
		 & \mcrot{1}{l}{60}{ImageNet} & \mcrot{1}{l}{60}{VOC IC} & \mcrot{1}{l}{60}{VOC OD} & \mcrot{1}{l}{60}{VOC Seg} & \mcrot{1}{l}{60}{COCO OD} & \mcrot{1}{l}{60}{COCO Seg} & \mcrot{1}{l}{60}{CIFAR 10} & \mcrot{1}{l}{60}{CIFAR 100} & \mcrot{1}{l}{60}{Places 205}\\
		\midrule
		Rotation & \checkmark &\checkmark &\checkmark &\checkmark & & & \checkmark & & \checkmark \\
		Jigsaw & \checkmark & \checkmark &\checkmark &\checkmark &\\
		DINO & \checkmark & & & & & & \checkmark & \checkmark &\\
		MAE &  \checkmark & & & & \checkmark & \checkmark & & & \checkmark\\
		EsViT &\checkmark &\checkmark & & & \checkmark &\checkmark &\checkmark &\checkmark &\\
		BYOL  &\checkmark &\checkmark &\checkmark &\checkmark & & & \checkmark &\checkmark\\
		VicReg  &\checkmark &\checkmark &\checkmark & & \checkmark &\checkmark & & & \checkmark \\
		Barlow Twins & \checkmark &\checkmark &\checkmark & & \checkmark &  \checkmark \\
		SimSiam &\checkmark & & \checkmark & & \checkmark &\checkmark &\\
		W-MSE &\checkmark & & & & & & \checkmark & \checkmark \\
		SimCLR &\checkmark & \checkmark & & & & & \checkmark &\checkmark &\checkmark \\
		MoCo &\checkmark & & \checkmark & &\checkmark &\checkmark &\\
		MoCo v2 &\checkmark & & \checkmark &\\
		CPC  &\checkmark &\\
		CPC v2 & \checkmark & & \checkmark \\
		CMC &\checkmark &\\
		PIRL & \checkmark & \checkmark &\checkmark & & & & & & \checkmark \\
		DeepCluster & \checkmark &\checkmark &\checkmark &\checkmark & & & & & \checkmark \\
		SeLa & \checkmark &\checkmark &\checkmark &\checkmark & & & \checkmark &\checkmark &\checkmark \\
		SwAV & \checkmark & \checkmark & \checkmark & & \checkmark & \checkmark & & & \checkmark \\
		\midrule
		\# Total & 20 & 11 & 13 & 5 & 7 & 7 & 8 & 7 & 8\\
		\bottomrule
	\end{tabular}
\end{table}

To get a sense of the performance of all reviewed methods at a glance, we conduct a quantitative comparison in the following. We report the evaluation results for ImageNet, Pascal VOC and Microsoft COCO and identify gaps where further evaluation would be interesting. All methods have been pre-trained on the ImageNet training set before further evaluation.

\paragraph{ImageNet.} Image classification on ImageNet, which includes \num{1000} different classes and has established as an evaluation standard for representation learning methods, is mostly evaluated in two ways. A linear classifier is either trained on top of the frozen pre-trained representations, 
or the model weights are initialized with the pre-trained weights and fine-tuned on 1\% and 10\% of the labeled ImageNet training data, respectively. Table \ref{tab:imagenet_accuracy} shows the accuracy for each method either using a ResNet-50 encoder for better comparability or a different architecture. We report both the top-1, as well as the top-5 accuracies. Among the methods evaluated with a ResNet-50, DINO and SwAV perform best, nearly reaching the performance of a supervised trained ResNet50. Considering the reported top-5 accuracy, BYOL performs best, right before VicReg and Barlow Twins. For other, bigger architectures, EsViT and DINO both perform best, while utilizing architectures with a comparatively little number of parameters.  

\begin{table}
    \begin{minipage}{\linewidth}
	\centering
	\footnotesize
	\caption{Top-1 and top-5 accuracy on ImageNet classification for a linear evaluation (on the left) and semi-supervised learning, where the classifier is fine-tuned on 1\% and 10\% respectively of the labeled ImageNet data (on the right). The upper part shows performance for a ResNet-50 encoder and the lower part shows more results where other architectures have been used. We also report the number of parameters for each network.}	\label{tab:imagenet_accuracy}
	\begin{tabular}{lc|cc|cccc}
	\toprule
	  & \#Params & Top-1 & Top-5 & \multicolumn{2}{c}{Top-1} & \multicolumn{2}{c}{Top-5}\\
	 Method & & & & 1\% & 10\% & 1\% & 10\% \\
	 \midrule
	 \multicolumn{8}{c}{\textit{ResNet-50}}\\
	 \midrule
	 \textcolor{gray}{Supervised (\scriptsize{\citeauthor{zbontar2021barlow})}} & \textcolor{gray}{24M} & \textcolor{gray}{76.5} &\textcolor{gray}{-}& \textcolor{gray}{25.4} & \textcolor{gray}{56.4} & \textcolor{gray}{48.4} & \textcolor{gray}{80.4} \\
	 DINO & 24M & \textbf{75.3} &-&-&-&-&-\\
	 BYOL & 24M & 74.3 & \textbf{91.6} & 53.2 & 68.8 & 78.4 & 89.0\\
	 VicReg & 24M & 73.2 & 91.1 & 54.8 & 69.5 & 79.4 & 89.5 \\
	 Barlow Twins & 24M & 73.2 & 91.0 & \textbf{55.0} & 69.7 & \textbf{79.2} & 89.3\\
	 SimSiam & 24M & 71.3&-&-&-&-&-\\
	 SimCLR & 24M & 69.3 & 89.0 & 48.3 & 65.6 & 75.5 & 87.8\\
	 MoCo & 24M & 60.6 & - &-&-&-&-\\
	 MoCo v2 & 24M & 71.1 & 90.1 & - & - & - & - \\
	 CPC v2 & 24M & 63.8 & 85.3 & - & - & - & - \\
	 CMC\footnotemark[3] & 24M & 66.2 & 87.0\\
	 PIRL & 24M & 63.6 & - & 30.7 & 60.4 & 57.2 & 83.8\\
	 SeLa & 24M & 61.5 & 84.0\\
	 SwAV w multi-crop & 24M & \textbf{75.3} & - & 53.9 & \textbf{70.2} & 78.5 & \textbf{89.9} \\
	 \midrule
	 \multicolumn{8}{c}{\textit{other architectures}}\\
	 \midrule
	 Rotation\footnotemark[4]  & & 55.4 & 77.9 &-&-&-&- \\
	 Jigsaw\footnotemark[4] & & 44.6 & 68.0 &-&-&-&-\\
	 MAE \scriptsize{(ViT-H)} & 643M & 76.6 & - & - &-&-&-\\
	 DINO \scriptsize{(ViT-B/8)} & 85M & 80.1 &-&-&-&-\\
	 EsViT \scriptsize{(Swin-B)} & 87M & \textbf{80.4} & - &-&-&-&-\\
	 BYOL \scriptsize{(ResNet-200x2)} & 250M & 79.6 &  \textbf{94.8} & \textbf{71.2} & \textbf{77.7} & \textbf{89.5} & \textbf{93.7} \\
	 W-SME 4 \scriptsize{(ResNet-18)} & & 79.0 & 94.5 & - &-&-&- \\
	 SimCLR \scriptsize{(ResNet-50x4)} & 375M & 76.5 & 93.2 & - & - & 85.8 & 92.6\\
	 MoCo \scriptsize{(ResNet-50x4)} & 375M & 68.6  & - &-&-&-&-\\
	 CPC \scriptsize{(ResNetv2-101)} & 28M & 48.7 & 73.6 &-&-&-&- \\
	 CPC v2 \scriptsize{(ResNet-161)} & 305M & 71.5 & 90.1 &-&-&-&- \\
	 CPC v2 \scriptsize{(ResNet-33)} & & - &-&52.7&73.1&78.3&91.2\\
	 CMC\footnotemark[3] \scriptsize{(ResNet-50x2)} & 188M & 70.6 & 89.7 &-&-&-&- \\
	 DeepCluster \scriptsize{(AlexNet)} & 62M & 41.0 & - &-&-&-&- \\
	 \bottomrule
	\end{tabular}
	\setcounter{footnote}{3}
	\footnotetext{trained with RandAugment~\citep{cubuk2020randaugment}.}
	\setcounter{footnote}{4}
	\footnotetext{reported numbers are from \citet{kolesnikov2019revisiting}.}
    \end{minipage}
\end{table}

\begin{figure}
	\centering \includegraphics[width=\textwidth]{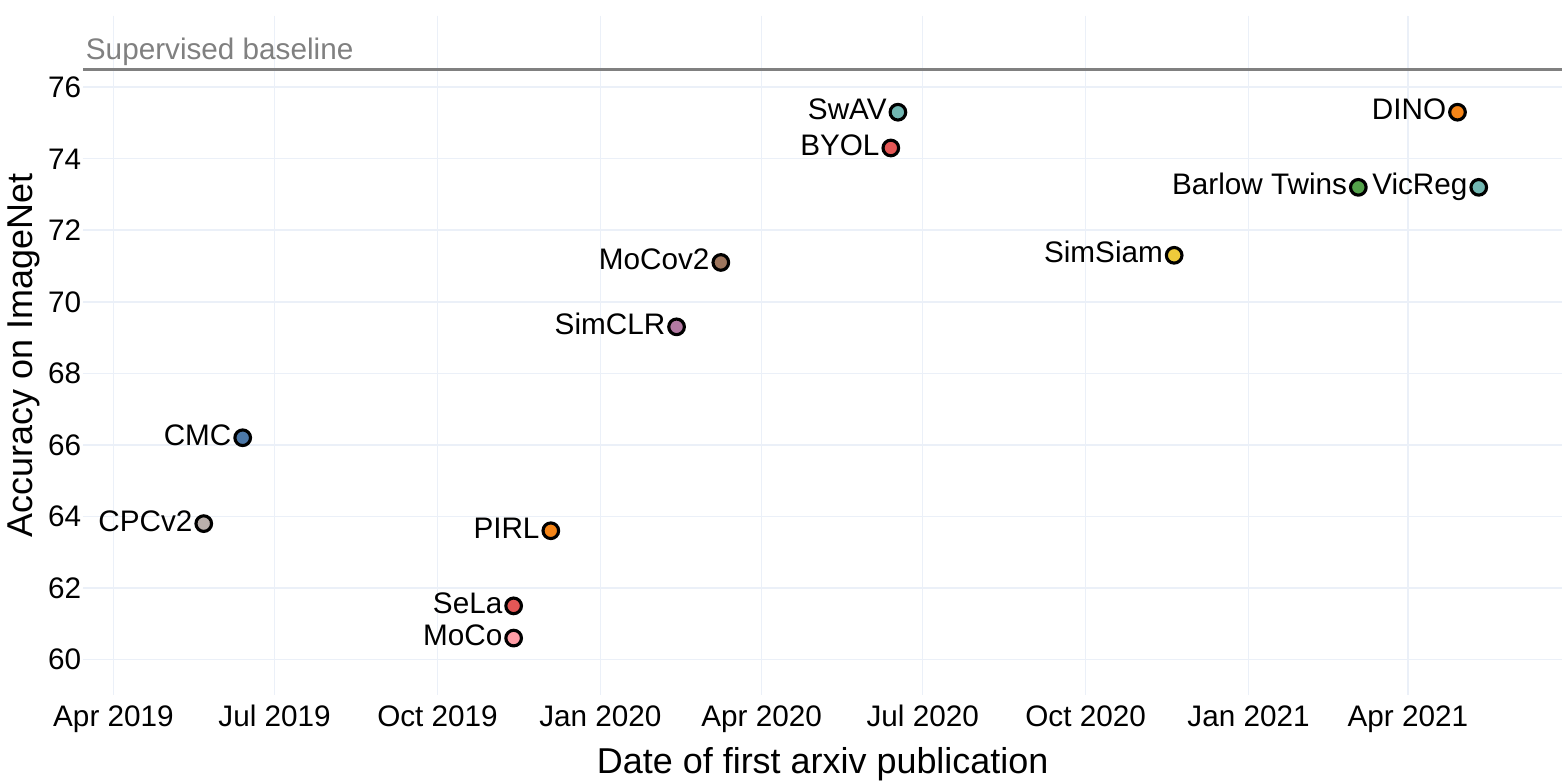}
	\caption[Reviewed methods Top-1 accuracy on ImageNet.]{Reviewed methods Top-1 accuracy on ImageNet using a ResNet-50 encoder, sorted by their first publication on arXiv\footnotemark[2]. The gray line marks the supervised benchmark.}
	\label{fig:accuracy_by_date}
\end{figure}

\setcounter{footnote}{2}

Figure \ref{fig:accuracy_by_date} shows the accuracy of all methods on the ImageNet training set using a ResNet-50 encoder in chronological order of their first publication on arXiv\footnotemark[2]\footnotetext{\url{https://arxiv.org/}}. Note that not every reviewed method appears in the Figure, because some methods have not been evaluated with a ResNet-50 architecture on ImageNet. It would be interesting to measure the performance of every method under the same conditions to make them comparable. 
Nevertheless, the plot reveals some interesting points, e.g., that CPC v2 and CMC perform better than most other early published self-supervised methods and SwAV has not been beaten by any of the compared methods.

\begin{sidewaystable}
	\centering
	\footnotesize
	\caption{Evaluation of a linear classifier on the PASCAL VOC image classification (IC), object detection (OD) and instance segmentation (Seg) and COCO OD and Seg tasks. For IC we report the mean average precision (mAP), the AP\textsubscript{all}, AP\textsubscript{50} and AP\textsubscript{75} for the VOC OD, COCO OD and COCO Seg and for PASCAL VOC Seg the mean intersection over union (mIoU).
		We report the best values reported in the original papers using different architectures.}
	\label{tab:pascal_accuracy}
	\begin{tabular}{l|c|ccc|c|ccc|ccc}
		\toprule
		& \multicolumn{5}{c}{Pascal VOC} & \multicolumn{6}{c}{COCO}\\
		& IC & \multicolumn{3}{c}{OD} & Seg & \multicolumn{3}{c}{OD} & \multicolumn{3}{c}{Seg} \\
		Method & AP & AP\textsubscript{all} & AP\textsubscript{50} & AP\textsubscript{75} & mIoU & $\text{AP}_{\text{all}}^{\text{BB}}$ & $\text{AP}_{50}^{\text{BB}}$ & $\text{AP}_{75}^{\text{BB}}$ & $\text{AP}_{\text{all}}^{\text{MK}}$ & $\text{AP}_{50}^{\text{MK}}$ & $\text{AP}_{75}^{\text{MK}}$\\
		\midrule
		Rotation\footnotemark[5] 
		& 72.9 & 54.4 & - & - & 39.1&-&-&-&-&-&-\\
		Jigsaw\footnotemark[5] 
		& 67.6 & 53.2 & - & - & 37.6&-&-&-&-&-&-\\
		MAE &-&-&-&-&-& \textbf{53.3} &-&-& \textbf{47.2} &-& \\
		EsViT & 85.5 &-&-&-&- & 46.2 & \textbf{68.0} & \textbf{50.6} & 41.6 & \textbf{64.9} & \textbf{44.8}\\
		BYOL & 85.4 & - & 77.5 & - & \textbf{76.3}&-&-&-&-&-&-\\
		VicReg & 86.6 & - & 82.4 & - & - & 39.4 & -& -& 36.4&-&-\\
		Barlow Twins & 86.2 & 56.8 & \textbf{82.6} & 63.4& - & 39.2 & 59.0 & 42.5 & 34.3 & 56.0 & 36.5\\
		SimSiam & - & 57.0 & 82.4 & 63.7 & - & 39.2 & 59.3 & 42.1 & 34.4 & 56.0 & 36.7 \\
		SimCLR & 80.5 & - &-&-&-&-&-&-&-&-&- \\
		MoCo & - & 55.9 & 81.5 & 62.6 & - & 40.8 & 61.6 & 44.7 & 36.9 & 58.4 & 39.7 \\
		MoCo v2 & - & 57.4 & 82.5 & \textbf{64.0} & - &-&-&-&-&-&-\\
		CPC v2 & - & 76.6 & - & - & - &-&-&-&-&-&-\\
		PIRL & 81.1 & 54.0 & 80.7 & 59.7 & - &-&-&-&-&-&-\\
		DeepCluster & 73.7 & 65.9 & - & - & 45.1 &-&-&-&-&-&-\\
		SeLa & 75.9 & \textbf{57.8} & - & - & 44.7 &-&-&-&-&-&-\\
		SwAV with multi-crop & \textbf{88.9} & 56.1 & \textbf{82.6} & 62.7 & - & 41.6 & 62.3 & 45.5 & 37.8 & 59.6 & 40.5\\
		\bottomrule
	\end{tabular}
	\setcounter{footnote}{5}
	\footnotetext{uses a data dependent initialization as proposed by \citet{krahenbuhl2015data}.}
\end{sidewaystable}

The goal of every representation learning method is to extract meaningful features from images that are useful for various tasks. The quality of extracted features learned on the ImageNet data can therefore be further assessed by transferring them to solve tasks on other datasets like the Pascal VOC and COCO object detection and instance segmentation. To evaluate the features on other tasks, the learned weights serve as initialization for a network and a linear classifier is trained on top, while the network layers are fine-tuned. Usually, the features are extracted from different layers of the network, while freezing the weights of the others. The best values are reported for each method. In the following we take a closer look at the results on the Pascal VOC and Microsoft COCO datasets. The goal when performing object detection on these two datasets is to predict bounding boxes for every object shown on the image. The task of instance segmentation is conducted pixel-wise, where every pixel is classified separately.

\paragraph{Pascal VOC.} With only \num{20} classes, the Pascal VOC~\citep{everingham2009pascal} dataset is rather small and designed to resemble a real world setting. The data includes annotations for image classification, object detection, as well as instance segmentation. The standard metrics for the classification and object detection task are mean average precision (AP) with different intersection over union (IoU) threshold values. For the task of segmentation the mean IoU is reported. For a detailed overview on the aforementioned metrics we refer to the work of \citet{9145130}.

The first five columns of Table \ref{tab:pascal_accuracy} show the results for the Pascal VOC tasks side by side for different architectures. For the object detection task Fast R-CNN \citep{girshick2015fast} and Faster R-CNN \citep{ren2015faster} are most widely used. 
SwAV with multicrop performs best on the image classification task, just before VicReg and Barlow Twins. Barlow Twins and SwAV also perform best on the object detection task. From all methods that have been evaluated on the segmentation task, BYOL, by far, outperforms every other method. However, the results cannot be deemed as representative due to a lack of comparative values for other methods.

\paragraph{COCO.} The Microsoft COCO dataset is a large dataset for object detection and segmentation including objects from \num{91} different classes, depicted in their natural setting. 
In the second half of Table \ref{tab:pascal_accuracy} we present the  average precision for object detection and instance segmentation. In the case of object detection it is the bounding box (BB) AP and in case of the segmentation it is the AP of the segmentation mask (MK).  Note that we again report values for different encoder architectures. The most commonly used architecture for both object detection and segmentation is a Mask R-CNN \citep{he2017mask} model with either the C4 or feature pyramid network \citep{lin2017feature} encoder. For both tasks the overall AP is best for MAE, other values are missing and EsViT outperforms every other method for the AP50 and AP75. Again, for a profound insight a more detailed experimental evaluation is necessary.

\subsection{Future Directions}
To conclude our quantitative meta-study we want to point out some interesting insights and suggest some experiments to conduct in the future. 

As the comparison of all representation learning methods has shown, the performance strongly depends on the used network architecture. Some architectures, datasets and tasks have been utilized throughout various works to evaluate the quality of self-supervised trained image representations. Nevertheless, there is no standardized benchmark for a consistent comparison of methods, yet. \citet{goyal2019scaling} suggest a range of tasks as benchmark to cover many aspects and obtain a detailed comparison of methods in the future.

One main contribution of this meta-study is the categorization and comprehensive presentation of different approaches to the overall aim to learn meaningful representations of images. In Section \ref{sec:taxonomy} five main categories have been described into which different approaches can be grouped. There already are some methods that combine multiple approaches that work well, which shows that the combination of different representation learning methods holds potential for future research.

%% file: sections/conclusion.tex
\section{Conclusion}

The goal of self-supervised representation learning is to extract meaningful features from unlabeled image data and use them to solve all kinds of downstream tasks.
In this work we saw different strategies for representation learning and how they are related. We gave an extensive overview of methods that have been developed over the last years and adapted a framework to categorize them.

%% file: representation.bbl
\begin{thebibliography}{}

\bibitem[Asano et~al., 2019]{sela}
Asano, Y.~M., Rupprecht, C., and Vedaldi, A. (2019).
\newblock Self-labelling via simultaneous clustering and representation
  learning.
\newblock {\em CoRR}, abs/1911.05371.

\bibitem[Ba et~al., 2016]{ba2016layer}
Ba, J.~L., Kiros, J.~R., and Hinton, G.~E. (2016).
\newblock Layer normalization.
\newblock {\em arXiv preprint arXiv:1607.06450}.

\bibitem[Bardes et~al., 2021]{bardes2021vicreg}
Bardes, A., Ponce, J., and LeCun, Y. (2021).
\newblock Vicreg: Variance-invariance-covariance regularization for
  self-supervised learning.
\newblock In {\em International Conference on Learning Representations}.

\bibitem[Barlow, 1961]{barlow1961possible}
Barlow, H.~B. (1961).
\newblock Possible principles underlying the transformation of sensory
  messages.
\newblock {\em Sensory communication}, 1(01).

\bibitem[Bastos et~al., 2012]{bastos2012canonical}
Bastos, A.~M., Usrey, W.~M., Adams, R.~A., Mangun, G.~R., Fries, P., and
  Friston, K.~J. (2012).
\newblock Canonical microcircuits for predictive coding.
\newblock {\em Neuron}, 76(4):695--711.

\bibitem[Caron et~al., 2018]{deep-cluster}
Caron, M., Bojanowski, P., Joulin, A., and Douze, M. (2018).
\newblock Deep clustering for unsupervised learning of visual features.
\newblock {\em CoRR}, abs/1807.05520.

\bibitem[Caron et~al., 2020]{swav}
Caron, M., Misra, I., Mairal, J., Goyal, P., Bojanowski, P., and Joulin, A.
  (2020).
\newblock Unsupervised learning of visual features by contrasting cluster
  assignments.
\newblock {\em CoRR}, abs/2006.09882.

\bibitem[Caron et~al., 2021]{caron2021emerging}
Caron, M., Touvron, H., Misra, I., J{\'e}gou, H., Mairal, J., Bojanowski, P.,
  and Joulin, A. (2021).
\newblock Emerging properties in self-supervised vision transformers.
\newblock In {\em Proceedings of the IEEE/CVF International Conference on
  Computer Vision}, pages 9650--9660.

\bibitem[Chen et~al., 2020a]{chen2020simple}
Chen, T., Kornblith, S., Norouzi, M., and Hinton, G. (2020a).
\newblock A simple framework for contrastive learning of visual
  representations.
\newblock In {\em International conference on machine learning}, pages
  1597--1607. PMLR.

\bibitem[Chen et~al., 2020b]{chen2020improved}
Chen, X., Fan, H., Girshick, R., and He, K. (2020b).
\newblock Improved baselines with momentum contrastive learning.
\newblock {\em arXiv preprint arXiv:2003.04297}.

\bibitem[Chen and He, 2021]{chen2021exploring}
Chen, X. and He, K. (2021).
\newblock Exploring simple siamese representation learning.
\newblock In {\em Proceedings of the IEEE/CVF Conference on Computer Vision and
  Pattern Recognition}, pages 15750--15758.

\bibitem[Cubuk et~al., 2020]{cubuk2020randaugment}
Cubuk, E.~D., Zoph, B., Shlens, J., and Le, Q.~V. (2020).
\newblock Randaugment: Practical automated data augmentation with a reduced
  search space.
\newblock In {\em Proceedings of the IEEE/CVF conference on computer vision and
  pattern recognition workshops}, pages 702--703.

\bibitem[Cuturi, 2013]{sinkhorn-knopp}
Cuturi, M. (2013).
\newblock Sinkhorn distances: Lightspeed computation of optimal transport.
\newblock In Burges, C., Bottou, L., Welling, M., Ghahramani, Z., and
  Weinberger, K., editors, {\em Advances in Neural Information Processing
  Systems}, volume~26. Curran Associates, Inc.

\bibitem[Dosovitskiy et~al., 2020]{dosovitskiy2020image}
Dosovitskiy, A., Beyer, L., Kolesnikov, A., Weissenborn, D., Zhai, X.,
  Unterthiner, T., Dehghani, M., Minderer, M., Heigold, G., Gelly, S., et~al.
  (2020).
\newblock An image is worth 16x16 words: Transformers for image recognition at
  scale.
\newblock {\em arXiv preprint arXiv:2010.11929}.

\bibitem[Dosovitskiy et~al., 2021]{dosovitskiy2021an}
Dosovitskiy, A., Beyer, L., Kolesnikov, A., Weissenborn, D., Zhai, X.,
  Unterthiner, T., Dehghani, M., Minderer, M., Heigold, G., Gelly, S.,
  Uszkoreit, J., and Houlsby, N. (2021).
\newblock An image is worth 16x16 words: Transformers for image recognition at
  scale.
\newblock In {\em International Conference on Learning Representations}.

\bibitem[Eldele et~al., 2021]{eldele2021time}
Eldele, E., Ragab, M., Chen, Z., Wu, M., Kwoh, C.~K., Li, X., and Guan, C.
  (2021).
\newblock Time-series representation learning via temporal and contextual
  contrasting.
\newblock {\em arXiv preprint arXiv:2106.14112}.

\bibitem[Ermolov et~al., 2021]{ermolov2021whitening}
Ermolov, A., Siarohin, A., Sangineto, E., and Sebe, N. (2021).
\newblock Whitening for self-supervised representation learning.
\newblock In {\em International Conference on Machine Learning}, pages
  3015--3024. PMLR.

\bibitem[Everingham et~al., 2009]{everingham2009pascal}
Everingham, M., Van~Gool, L., Williams, C.~K., Winn, J., and Zisserman, A.
  (2009).
\newblock The pascal visual object classes (voc) challenge.
\newblock {\em International journal of computer vision}, 88:303--308.

\bibitem[Gidaris et~al., 2018]{gidaris2018unsupervised}
Gidaris, S., Singh, P., and Komodakis, N. (2018).
\newblock Unsupervised representation learning by predicting image rotations.
\newblock In {\em International Conference on Learning Representations}.

\bibitem[Girshick, 2015]{girshick2015fast}
Girshick, R. (2015).
\newblock Fast r-cnn.
\newblock In {\em Proceedings of the IEEE international conference on computer
  vision}, pages 1440--1448.

\bibitem[Gou et~al., 2021]{gou2021kdsurvey}
Gou, J., Yu, B., Maybank, S.~J., and Tao, D. (2021).
\newblock Knowledge distillation: A survey.
\newblock {\em Int. J. Comput. Vision}, 129(6):1789–1819.

\bibitem[Goyal et~al., 2019]{goyal2019scaling}
Goyal, P., Mahajan, D., Gupta, A., and Misra, I. (2019).
\newblock Scaling and benchmarking self-supervised visual representation
  learning.
\newblock In {\em Proceedings of the ieee/cvf International Conference on
  computer vision}, pages 6391--6400.

\bibitem[Grill et~al., 2020]{grill2020bootstrap}
Grill, J.-B., Strub, F., Altch{\'e}, F., Tallec, C., Richemond, P.,
  Buchatskaya, E., Doersch, C., Avila~Pires, B., Guo, Z., Gheshlaghi~Azar, M.,
  et~al. (2020).
\newblock Bootstrap your own latent-a new approach to self-supervised learning.
\newblock {\em Advances in neural information processing systems},
  33:21271--21284.

\bibitem[Grover and Leskovec, 2016]{grover2016node2vec}
Grover, A. and Leskovec, J. (2016).
\newblock node2vec: Scalable feature learning for networks.
\newblock In {\em Proceedings of the 22nd ACM SIGKDD international conference
  on Knowledge discovery and data mining}, pages 855--864.

\bibitem[Gutmann and Hyv{\"a}rinen, 2010]{gutmann2010noise}
Gutmann, M. and Hyv{\"a}rinen, A. (2010).
\newblock Noise-contrastive estimation: A new estimation principle for
  unnormalized statistical models.
\newblock In {\em Proceedings of the thirteenth international conference on
  artificial intelligence and statistics}, pages 297--304. JMLR Workshop and
  Conference Proceedings.

\bibitem[He et~al., 2022]{he2022masked}
He, K., Chen, X., Xie, S., Li, Y., Doll{\'a}r, P., and Girshick, R. (2022).
\newblock Masked autoencoders are scalable vision learners.
\newblock In {\em Proceedings of the IEEE/CVF Conference on Computer Vision and
  Pattern Recognition}, pages 16000--16009.

\bibitem[He et~al., 2020]{he2020momentum}
He, K., Fan, H., Wu, Y., Xie, S., and Girshick, R. (2020).
\newblock Momentum contrast for unsupervised visual representation learning.
\newblock In {\em Proceedings of the IEEE/CVF conference on computer vision and
  pattern recognition}, pages 9729--9738.

\bibitem[He et~al., 2017]{he2017mask}
He, K., Gkioxari, G., Doll{\'a}r, P., and Girshick, R. (2017).
\newblock Mask r-cnn.
\newblock In {\em Proceedings of the IEEE international conference on computer
  vision}, pages 2961--2969.

\bibitem[He et~al., 2016]{he2016deep}
He, K., Zhang, X., Ren, S., and Sun, J. (2016).
\newblock Deep residual learning for image recognition.
\newblock In {\em Proceedings of the IEEE conference on computer vision and
  pattern recognition}, pages 770--778.

\bibitem[Henaff, 2020]{henaff2020data}
Henaff, O. (2020).
\newblock Data-efficient image recognition with contrastive predictive coding.
\newblock In {\em International conference on machine learning}, pages
  4182--4192. PMLR.

\bibitem[Huang and Rao, 2011]{huang2011predictive}
Huang, Y. and Rao, R.~P. (2011).
\newblock Predictive coding.
\newblock {\em Wiley Interdisciplinary Reviews: Cognitive Science},
  2(5):580--593.

\bibitem[Ioffe and Szegedy, 2015]{ioffe2015batch}
Ioffe, S. and Szegedy, C. (2015).
\newblock Batch normalization: Accelerating deep network training by reducing
  internal covariate shift.
\newblock In {\em International conference on machine learning}, pages
  448--456. PMLR.

\bibitem[Kenton and Toutanova, 2019]{kenton2019bert}
Kenton, J. D. M.-W.~C. and Toutanova, L.~K. (2019).
\newblock Bert: Pre-training of deep bidirectional transformers for language
  understanding.
\newblock In {\em Proceedings of NAACL-HLT}, pages 4171--4186.

\bibitem[Kingma and Welling, 2013]{kingma2013auto}
Kingma, D.~P. and Welling, M. (2013).
\newblock Auto-encoding variational bayes.
\newblock {\em arXiv preprint arXiv:1312.6114}.

\bibitem[Kipf and Welling, 2016]{kipf2016variational}
Kipf, T.~N. and Welling, M. (2016).
\newblock Variational graph auto-encoders.
\newblock {\em arXiv preprint arXiv:1611.07308}.

\bibitem[Kolesnikov et~al., 2019]{kolesnikov2019revisiting}
Kolesnikov, A., Zhai, X., and Beyer, L. (2019).
\newblock Revisiting self-supervised visual representation learning.
\newblock In {\em Proceedings of the IEEE/CVF conference on computer vision and
  pattern recognition}, pages 1920--1929.

\bibitem[Kr{\"a}henb{\"u}hl et~al., 2015]{krahenbuhl2015data}
Kr{\"a}henb{\"u}hl, P., Doersch, C., Donahue, J., and Darrell, T. (2015).
\newblock Data-dependent initializations of convolutional neural networks.
\newblock {\em arXiv preprint arXiv:1511.06856}.

\bibitem[Krizhevsky et~al., 2009]{krizhevsky2009learning}
Krizhevsky, A., Hinton, G., et~al. (2009).
\newblock Learning multiple layers of features from tiny images.

\bibitem[Krizhevsky et~al., 2017]{krizhevsky2017imagenet}
Krizhevsky, A., Sutskever, I., and Hinton, G.~E. (2017).
\newblock Imagenet classification with deep convolutional neural networks.
\newblock {\em Communications of the ACM}, 60(6):84--90.

\bibitem[Le~Cun, 1987]{le1987modeles}
Le~Cun, Y. (1987).
\newblock {\em Mod{\`e}les connexionnistes de l'apprentissage}.
\newblock PhD thesis, Paris 6.

\bibitem[Li et~al., 2021]{li2021efficient}
Li, C., Yang, J., Zhang, P., Gao, M., Xiao, B., Dai, X., Yuan, L., and Gao, J.
  (2021).
\newblock Efficient self-supervised vision transformers for representation
  learning.
\newblock In {\em International Conference on Learning Representations}.

\bibitem[Lillicrap et~al., 2019]{lillicrap2019continuous}
Lillicrap, T.~P., Hunt, J.~J., Pritzel, A., Heess, N., Erez, T., Tassa, Y.,
  Silver, D., and Wierstra, D. (2019).
\newblock Continuous control with deep reinforcement learning.

\bibitem[Lin et~al., 2013]{lin2013network}
Lin, M., Chen, Q., and Yan, S. (2013).
\newblock Network in network.
\newblock {\em arXiv preprint arXiv:1312.4400}.

\bibitem[Lin et~al., 2017]{lin2017feature}
Lin, T.-Y., Doll{\'a}r, P., Girshick, R., He, K., Hariharan, B., and Belongie,
  S. (2017).
\newblock Feature pyramid networks for object detection.
\newblock In {\em Proceedings of the IEEE conference on computer vision and
  pattern recognition}, pages 2117--2125.

\bibitem[Lin et~al., 2014]{lin2014microsoft}
Lin, T.-Y., Maire, M., Belongie, S., Hays, J., Perona, P., Ramanan, D.,
  Doll{\'a}r, P., and Zitnick, C.~L. (2014).
\newblock Microsoft coco: Common objects in context.
\newblock In {\em Computer Vision--ECCV 2014: 13th European Conference, Zurich,
  Switzerland, September 6-12, 2014, Proceedings, Part V 13}, pages 740--755.
  Springer.

\bibitem[Lloyd, 1982]{lloyd1982least}
Lloyd, S. (1982).
\newblock Least squares quantization in pcm.
\newblock {\em IEEE transactions on information theory}, 28(2):129--137.

\bibitem[Loshchilov and Hutter, 2017]{loshchilov2017sgdr}
Loshchilov, I. and Hutter, F. (2017).
\newblock {SGDR}: Stochastic gradient descent with warm restarts.
\newblock In {\em International Conference on Learning Representations}.

\bibitem[Mikolov et~al., 2013a]{mikolov2013efficient}
Mikolov, T., Chen, K., Corrado, G., and Dean, J. (2013a).
\newblock Efficient estimation of word representations in vector space.
\newblock {\em arXiv preprint arXiv:1301.3781}.

\bibitem[Mikolov et~al., 2013b]{mikolov2013distributed}
Mikolov, T., Sutskever, I., Chen, K., Corrado, G.~S., and Dean, J. (2013b).
\newblock Distributed representations of words and phrases and their
  compositionality.
\newblock {\em Advances in neural information processing systems}, 26.

\bibitem[Misra and Maaten, 2020]{misra2020self}
Misra, I. and Maaten, L. v.~d. (2020).
\newblock Self-supervised learning of pretext-invariant representations.
\newblock In {\em Proceedings of the IEEE/CVF Conference on Computer Vision and
  Pattern Recognition}, pages 6707--6717.

\bibitem[Noroozi and Favaro, 2016]{noroozi2016unsupervised}
Noroozi, M. and Favaro, P. (2016).
\newblock Unsupervised learning of visual representations by solving jigsaw
  puzzles.
\newblock In {\em European conference on computer vision}, pages 69--84.
  Springer.

\bibitem[Padilla et~al., 2020]{9145130}
Padilla, R., Netto, S.~L., and da~Silva, E. A.~B. (2020).
\newblock A survey on performance metrics for object-detection algorithms.
\newblock In {\em 2020 International Conference on Systems, Signals and Image
  Processing (IWSSIP)}, pages 237--242.

\bibitem[Perozzi et~al., 2014]{perozzi2014deepwalk}
Perozzi, B., Al-Rfou, R., and Skiena, S. (2014).
\newblock Deepwalk: Online learning of social representations.
\newblock In {\em Proceedings of the 20th ACM SIGKDD international conference
  on Knowledge discovery and data mining}, pages 701--710.

\bibitem[Ren et~al., 2015]{ren2015faster}
Ren, S., He, K., Girshick, R., and Sun, J. (2015).
\newblock Faster r-cnn: Towards real-time object detection with region proposal
  networks.
\newblock {\em Advances in neural information processing systems}, 28.

\bibitem[Rifai et~al., 2011]{rifai2011contractive}
Rifai, S., Vincent, P., Muller, X., Glorot, X., and Bengio, Y. (2011).
\newblock Contractive auto-encoders: Explicit invariance during feature
  extraction.
\newblock In {\em Proceedings of the 28th International Conference on
  International Conference on Machine Learning}, ICML'11, page 833–840.
  Omnipress.

\bibitem[Russakovsky et~al., 2015]{russakovsky2015imagenet}
Russakovsky, O., Deng, J., Su, H., Krause, J., Satheesh, S., Ma, S., Huang, Z.,
  Karpathy, A., Khosla, A., Bernstein, M., et~al. (2015).
\newblock Imagenet large scale visual recognition challenge.
\newblock {\em International journal of computer vision}, 115:211--252.

\bibitem[Sch{\"o}lkopf et~al., 2012]{scholkopf2012causal}
Sch{\"o}lkopf, B., Janzing, D., Peters, J., Sgouritsa, E., Zhang, K., and
  Mooij, J. (2012).
\newblock On causal and anticausal learning.
\newblock {\em arXiv preprint arXiv:1206.6471}.

\bibitem[Shorten and Khoshgoftaar, 2019]{shorten2019survey}
Shorten, C. and Khoshgoftaar, T.~M. (2019).
\newblock A survey on image data augmentation for deep learning.
\newblock {\em Journal of big data}, 6(1):1--48.

\bibitem[Tian et~al., 2020]{tian2020contrastive}
Tian, Y., Krishnan, D., and Isola, P. (2020).
\newblock Contrastive multiview coding.
\newblock In {\em European conference on computer vision}, pages 776--794.
  Springer.

\bibitem[van~den Oord et~al., 2016]{van2016conditional}
van~den Oord, A., Kalchbrenner, N., Espeholt, L., Vinyals, O., Graves, A.,
  et~al. (2016).
\newblock Conditional image generation with pixelcnn decoders.
\newblock {\em Advances in neural information processing systems}, 29.

\bibitem[van~den Oord et~al., 2018]{oord2018representation}
van~den Oord, A., Li, Y., and Vinyals, O. (2018).
\newblock Representation learning with contrastive predictive coding.
\newblock {\em arXiv preprint arXiv:1807.03748}.

\bibitem[{Vaswani} et~al., 2021]{vaswani2021multi}
{Vaswani}, A., {Ramachandran}, P., {Srinivas}, A., {Parmar}, N., {Hechtman},
  B., and {Shlens}, J. (2021).
\newblock {Scaling Local Self-Attention for Parameter Efficient Visual
  Backbones}.
\newblock {\em arXiv e-prints}, page arXiv:2103.12731.

\bibitem[Vincent et~al., 2010]{vincent2010stacked}
Vincent, P., Larochelle, H., Lajoie, I., Bengio, Y., Manzagol, P.-A., and
  Bottou, L. (2010).
\newblock Stacked denoising autoencoders: Learning useful representations in a
  deep network with a local denoising criterion.
\newblock {\em Journal of machine learning research}, 11(12).

\bibitem[Wu et~al., 2018]{wu2018unsupervised}
Wu, Z., Xiong, Y., Yu, S.~X., and Lin, D. (2018).
\newblock Unsupervised feature learning via non-parametric instance
  discrimination.
\newblock In {\em Proceedings of the IEEE conference on computer vision and
  pattern recognition}, pages 3733--3742.

\bibitem[Yang et~al., 2022]{yang2022image}
Yang, S., Xiao, W., Zhang, M., Guo, S., Zhao, J., and Shen, F. (2022).
\newblock Image data augmentation for deep learning: A survey.
\newblock {\em arXiv preprint arXiv:2204.08610}.

\bibitem[You et~al., 2017]{you2017large}
You, Y., Gitman, I., and Ginsburg, B. (2017).
\newblock Large batch training of convolutional networks.
\newblock {\em arXiv preprint arXiv:1708.03888}.

\bibitem[Zbontar et~al., 2021]{zbontar2021barlow}
Zbontar, J., Jing, L., Misra, I., LeCun, Y., and Deny, S. (2021).
\newblock Barlow twins: Self-supervised learning via redundancy reduction.
\newblock In {\em International Conference on Machine Learning}, pages
  12310--12320. PMLR.

\bibitem[Zhou et~al., 2014]{zhou2014learning}
Zhou, B., Lapedriza, A., Xiao, J., Torralba, A., and Oliva, A. (2014).
\newblock Learning deep features for scene recognition using places database.
\newblock {\em Advances in neural information processing systems}, 27.

\end{thebibliography}
